\documentclass{article}

\usepackage{microtype}
\usepackage{graphicx}
\usepackage{subfigure}
\usepackage{booktabs} 

\PassOptionsToPackage{hyphens}{url}\usepackage{hyperref}
\usepackage{url}            
\usepackage{amsfonts}       
\usepackage{amsmath}       
\usepackage{nicefrac}       
\usepackage{pifont}         %
\usepackage{wrapfig}
\usepackage{lipsum}

\usepackage{caption}

\usepackage{natbib}
\usepackage{verbatim}

\usepackage[inline]{enumitem}

\newif\iftrvar
\trvartrue
\iftrvar

\newcommand{\eg}{\textit{e.g.}}
\newcommand{\ie}{\textit{i.e.}}

\def\pdvf{PD-VF}
\def\maml{MAML}
\def\rl2{$RL^2$}
\def\ppoall{PPOall}
\def\ppoenv{PPOenv}
\def\kmeans{Kmeans}
\def\nn{NN}
\def\condpi{PPOdyn}
\def\ndp{NoAggPolicy}
\def\ndv{NoAggValue}

\usepackage{soul}

\usepackage[accepted]{include/icml2020}

\icmltitlerunning{Fast Adaptation to New Environments via Policy-Dynamics Value Functions}

\begin{document}

\twocolumn[
\icmltitle{Fast Adaptation to New Environments via Policy-Dynamics Value Functions}



\icmlsetsymbol{equal}{*}

\begin{icmlauthorlist}
\icmlauthor{Roberta Raileanu}{nyu}
\icmlauthor{Max Goldstein}{nyu}
\icmlauthor{Arthur Szlam}{fair}
\icmlauthor{Rob Fergus}{nyu}
\end{icmlauthorlist}

\icmlaffiliation{nyu}{Department of Computer Science, New York University, New York, USA}
\icmlaffiliation{fair}{Facebook AI Research, New York, USA}

\icmlcorrespondingauthor{Roberta Raileanu}{raileanu@cs.nyu.edu}

\icmlkeywords{Machine Learning, ICML}

\vskip 0.3in
]



\printAffiliationsAndNotice{} 

\begin{abstract}
Standard RL algorithms assume fixed environment dynamics and require a significant amount of interaction to adapt to new environments. We introduce Policy-Dynamics Value Functions (PD-VF), a novel approach for rapidly adapting to dynamics different from those previously seen in training. PD-VF explicitly estimates the cumulative reward in a space of policies and environments. An ensemble of conventional RL policies is used to gather experience on training environments, from which embeddings of both policies and environments can be learned. Then, a value function conditioned on both embeddings is trained. At test time, a few actions are sufficient to infer the environment embedding, enabling a policy to be selected by maximizing the learned value function (which requires no additional environment interaction). We show that our method can rapidly adapt to new dynamics on a set of MuJoCo domains. Code available at \href{https://github.com/rraileanu/policy-dynamics-value-functions}{\nolinkurl{policy-dynamics-value-functions}}.

\end{abstract}

\section{Introduction}

Deep reinforcement learning (RL) has achieved impressive results on a wide range of complex tasks \citep{mnih2015human, silver2016mastering,  silver2017mastering, silver2018general, jaderberg2019human, Berner2019Dota2W, vinyals2019grandmaster}. However, recent studies have pointed out that RL agents trained and tested on the same environment tend to overfit to that environment's idiosyncracies and are unable to generalize to even small perturbations \citep{Whiteson2011ProtectingAE, rajeswaran2017towards, zhang2018study, zhang2018dissection, henderson2018deep, Cobbe2019QuantifyingGI, Raileanu2020RIDERI, Song2020ObservationalOI}.
It is often the case that besides the test environments being different from the train environments, they will also have
costly interactions, scarce or unavailable feedback, and irreversible consequences. For example, a self-driving car might have to adjust its behavior depending on weather conditions, or a prosthetic control system might have to adapt to a new human. In these cases it is crucial for RL agents to find and execute appropriate policies as quickly as possible.

Our approach is inspired by \citet{Sutton2011HordeAS} who introduced the notion of general value functions (GVFs), which can be used to gather knowledge about the world in the form of predictions. A GVF estimates the expected return of an arbitrary policy on a certain task (as defined by a reward function, a termination function and a terminal-reward function). Similarly, in this work, we aim to learn a value function conditioned on elements of a space of policies and tasks, but here, a ``task'' is specified by the transition function of the MDP instead of the reward function. 

More specifically, we propose \pdvf{}, a novel framework for rapid adaptation to new environment dynamics. \pdvf{} consists of four phases: 
\begin{enumerate*}[label=(\roman*)] 
    \item a \textit{reinforcement learning phase} in which individual policies are learned for each environment in our training set using standard RL algorithms, 
    \item a \textit{self-supervised phase} in which trajectories generated by these policies are used to learn embeddings for both policies and environments, 
    \item a \textit{supervised training phase} in which a neural network is used to learn the value function of a certain policy acting in some environment. The network takes as inputs the initial state of the environment, as well as the corresponding policy and environment embeddings (as learned in the previous phase) and is trained with supervision of the cumulative reward obtained during an episode, and finally 
    \item an \textit{evaluation phase} in which, given a new environment, its dynamics embedding is inferred using the first few steps of an episode. Then, a policy is selected by finding the policy embedding that maximizes the learned value function. The selected policy is used to act in the environment until the episode ends.
\end{enumerate*}

Our framework uses self-supervised interactions with the environment to learn an embedding space of both dynamics and policies. By learning a value function in the policy-dynamics space, \pdvf{} can discover useful patterns in the complex relation between a family of environment dynamics, various behaviors, and the expected return.  The value function is designed to model non-optimal policies along with optimal policies in given environments so that it can understand how changes in dynamics relate to changes in the return of different policies. \pdvf{} uses the learned space of dynamics to rapidly embed a new environment in that space using only a few interactions.  At test time,  \pdvf{} can evaluate or rank policies (from a certain family) on unseen environments without the need of full rollouts (\ie{} it does not require full trajectories or rewards to update the policy). We evaluate our method on a set of continuous control tasks (with varying dynamics) in MuJoCo \citep{Todorov2012MuJoCoAP}. The dynamics of each task instance are determined by physical parameters such as wind direction or limb length and can be sampled from a continuous or discrete distribution. Performance is evaluated on a single episode at test time to emphasize rapid adaptation. We show that \pdvf{} outperforms other meta-learning and transfer learning approaches on new environments with unseen dynamics. 

\label{introduction}
\section{Related Work}
Our work draws inspiration from multiple research areas such as transfer learning \citep{taylor2009transfer, Higgins2017DARLAIZ}, skill and task embedding \citep{Devin2016LearningMN, Zhang2018DecouplingDA, Hausman2018LearningAE, Petangoda2019DisentangledSE}, and general value functions \citep{precup2001off, Sutton2011HordeAS, white2012scaling}.

\textbf{Multi-Task and Transfer Learning.}
\citet{taylor2009transfer} presents an overview of transfer learning methods in RL. A popular approach for transfer in RL is multi-task learning \citep{taylor2009transfer, teh2017distral}, a paradigm in which an agent is trained on a family of related tasks. By simultaneously learning about different tasks, the agent can exploit their common structure, which can lead to faster learning and better generalization to unseen tasks from the same family \citep{taylor2009transfer, lazaric2012transfer, ammar2012reinforcement, ammar2014automated, parisotto2015actor, borsa2016learning, gupta2017learning, andreas2017modular, oh2017zero, hessel2019multi}. A large body of work has been inspired by the Horde architecture \citep{Sutton2011HordeAS}, which consists of a number of RL agents with different policies and goals. Each agent is tasked with estimating the value function of a particular policy on a given task, thus collectively representing knowledge about the world. Building on these ideas, other methods leverage the shared dynamics of the tasks \citep{barreto2017successor, zhang2017deep, madjiheurem2019state2vec} or the similarity among value functions and the associated optimal policies \citep{schaul2015universal, borsa2018universal, hansen2019fast, siriwardhana2019vusfa}. These approaches assume the same underlying transition function for all tasks. In contrast, we focus on transferring knowledge across tasks with different dynamics.

\textbf{Meta-Learning and Robust Transfer.}
A popular approach for fast adaptation to new environments is meta reinforcement learning (meta RL) \citep{Cully2015RobotsTC, finn2017model, Wang2017RobustIO, Duan2016RL2FR, Xu2018MetaGradientRL, Houthooft2018EvolvedPG, saemundsson2018meta, nagabandi2018learning, humplik2019meta, rakelly2019efficient}. Meta RL methods have been designed to work well with dense reward and recent work has shown that they struggle to learn from a limited number of interactions and optimization steps at test time \citep{yang2019single}. In contrast, our framework is capable of rapid adaptation to new environment dynamics and does not require dense reward or a large number of interactions to find a good policy. Moreover, \pdvf{} does not update the model parameters at test time, which makes it less computationally expensive than meta RL. Another common approach for transfer across dynamics is model-based RL, which uses  Gaussian processes (GPs) or Bayesian neural networks (BNNs) to estimate the transition function \cite{DoshiVelez2013HiddenPM, Killian2017RobustAE}. However, such methods require fictional rollouts to train a policy from scratch at test time, which makes them computationally expensive and limits their applicability for real-world tasks. \citet{Yao2018DirectPT} uses a fully-trained BNN to further optimize latent variables during a single test episode, but requires an optimal policy for each training instance, which makes it harder to scale. Robust transfer methods either require a large number of interactions at test time \citep{rajeswaran2017towards} or assume that the distribution over hidden variables is known or controllable \citep{Paul2018FingerprintPO}. An alternative approach was proposed by \citet{Pinto2017RobustAR} who use an adversary to perturb the system, achieving robust transfer across physical parameters such as friction or mass.

\textbf{Skill and Task Embeddings.}
A large body of work proposes the use of learned skill and task embeddings for transfer in RL~\cite{da2012learning, sahni2017learning, oh2017zero, gupta2017learning, Hausman2018LearningAE, he2018zero}. For example, \citet{Hausman2018LearningAE} use approximate variational inference to learn a latent space of skills. Similarly, \citet{Arnekvist2018VPEVP} learn a stochastic embedding of optimal Q-functions for various skills and train a universal policy conditioned on this embedding. In both \citet{Hausman2018LearningAE} and \citet{Arnekvist2018VPEVP}, adaptation to a new task is done in the latent space with no further updates to the policy network. \citet{CoReyes2018SelfConsistentTA} learn a latent space of low-level skills that can be controlled by a higher-level policy, in the context of hierarchical reinforcement learning. This embedding is learned using a variational autoencoder \citep{Kingma2013AutoEncodingVB}
to encode state trajectories and decode states and actions.  \citet{Zintgraf2018FastCA} use a meta-learning approach to learn a deterministic task embedding. \citet{Wang2017RobustIO} and \citet{Duan2017OneShotIL} learn embeddings of expert demonstrations to aid imitation learning using variational and deterministic methods, respectively. More recently, \citet{Perez2018EfficientTL} learn dynamic models with auxiliary latent variables and use them for model-predictive control. \citet{Zhang2018DecouplingDA} use separate dynamics and reward modules to learn a task embedding. They show that conditioning a policy on this embedding helps transfer to changes in transition or reward function. While the above approaches might learn embeddings of skills or tasks, none of them leverage \textit{both} the latent space of policies and that of the environments for estimating the expected return and using it to select an effective policy at test time.

More similar to our work is that of \citet{yang2019single}, who also focus on fast adaptation to new environment dynamics and evaluate performance on a single episode at test time. \citet{yang2019single} train an inference model and a probe to estimate the underlying latent variables of the dynamics, which are then used as input to a universal control policy. While similar in scope, our approach is significantly different from that of \citet{yang2019single}. Importantly, \citet{yang2019single} does not learn a latent space of policies and instead trains a universal policy on all the environments. Learning a value function in a space of policies and dynamics allows the function approximator to capture relations among dynamics, behaviors (both optimal as well as non-optimal), and rewards that a universal policy cannot learn. Moreover, the learned structure can aid transfer to new dynamics.

\label{related}
\section{Policy-Dynamics Value Functions}
\label{method}

In this work, we aim to design an approach that can quickly find a good policy in an environment with new and unknown dynamics, after being trained on a family of environments with related dynamics.
The problem can be formalized as a family of Markov decision processes (MDPs) defined by $(\mathcal{S}, \mathcal{A}, \mathcal{T}, \mathcal{R}, \gamma)$, where $(\mathcal{S}, \mathcal{A}, \mathcal{R}, \gamma)$ are the corresponding state space, action space, reward function, and discount factor. Each instance of the family is a stationary MDP with transition function $\mathcal{T}_d(s'|s, a) \in \mathcal{T}$. Each $\mathcal{T}_d$ has a hidden parameter $d$ that is sampled once from a distribution $\mathcal{D}$ and held constant for that instance (\ie{} episode). $\mathcal{T}_d$ can be continuous or discrete in $d$. By design, the latent variable $d$ that defines the MDP's dynamics cannot be observed from individual states. 

We present Policy-Dynamics Value Functions (\pdvf{}), a novel framework for rapid adaptation across 
such MDPs with different dynamics. \pdvf{} is an extension of a value function that not only conditions on a state, but also on a policy and a transition function. 

A conventional value function $V: \mathcal{S} \rightarrow \mathcal{R}$ is defined as the expected future return from state $s$ of policy $\pi$: 
\begin{equation*}
    V(s) = \mathbb{E_{}} \left[ G_{t} | S_t = s \right] = \mathbb{E_{}} \left[ \sum_{k = t+1}^{T} \gamma^k r_{k} | S_t = s \right]. 
\end{equation*}

Formally, we define a \textit{policy-dynamics value function} or \pdvf{} as a function $W: \mathcal{S} \times \Pi \times \mathcal{T} \rightarrow \mathcal{R}$ with two auxiliary inputs representing the policy $\pi$ and the dynamics $d$:
\begin{equation*}
    W(s, \pi, d) = \mathbb{E} \left[ G_{t} | S_t = s, A_t \sim \pi, S_{t+1} \sim \mathcal{T}_d \right].
\end{equation*}

\subsection{Problem setup}
The dynamics distribution $\mathcal{D}$ is partitioned into two disjoint sets $\mathcal{D}_{train}$ and $\mathcal{D}_{test}$. These are used to generate a set of training and test environments, each having different transition functions, drawn from their respective distributions. 

Our model is learned on the training environments in three stages:
\begin{enumerate*}[label=(\roman*)] 
    \item a reinforcement learning phase,
    \item a self-supervised phase and 
    \item a supervised phase.
\end{enumerate*}
The resulting PD-VF model is evaluated on test environments, where it only experiences a single episode in each. This evaluation setting probes PD-VF's ability to very quickly adapt to previously unseen dynamics. 

\subsection{Reinforcement Learning Phase}
\label{sec:rltrain}
The first phase of training uses standard model-free RL algorithms to acquire experience in the training environments. An ensemble of $N$ policies are trained, each with a different random seed on one of the training environments. For each policy, we save a number of checkpoints at different stages throughout training. Then, we collect trajectories using each of these checkpoints in each of our training environments. This results in experience from a diverse set of policies (some good, some bad) across environments with different dynamics. Importantly, this dataset contains the behaviors of policies in environments they haven't been trained on. In the next section, we describe how the collected trajectories are used to learn policy and dynamics embeddings. 

\subsection{Self-Supervised Learning Phase}
The goal of this phase is to learn an embedding space of the dynamics that captures variations in the transition function, as well as an embedding space of the policies that captures variations in the agent behavior. 
\begin{figure}[ht!]
    \centering
    \includegraphics[width=0.7\columnwidth]{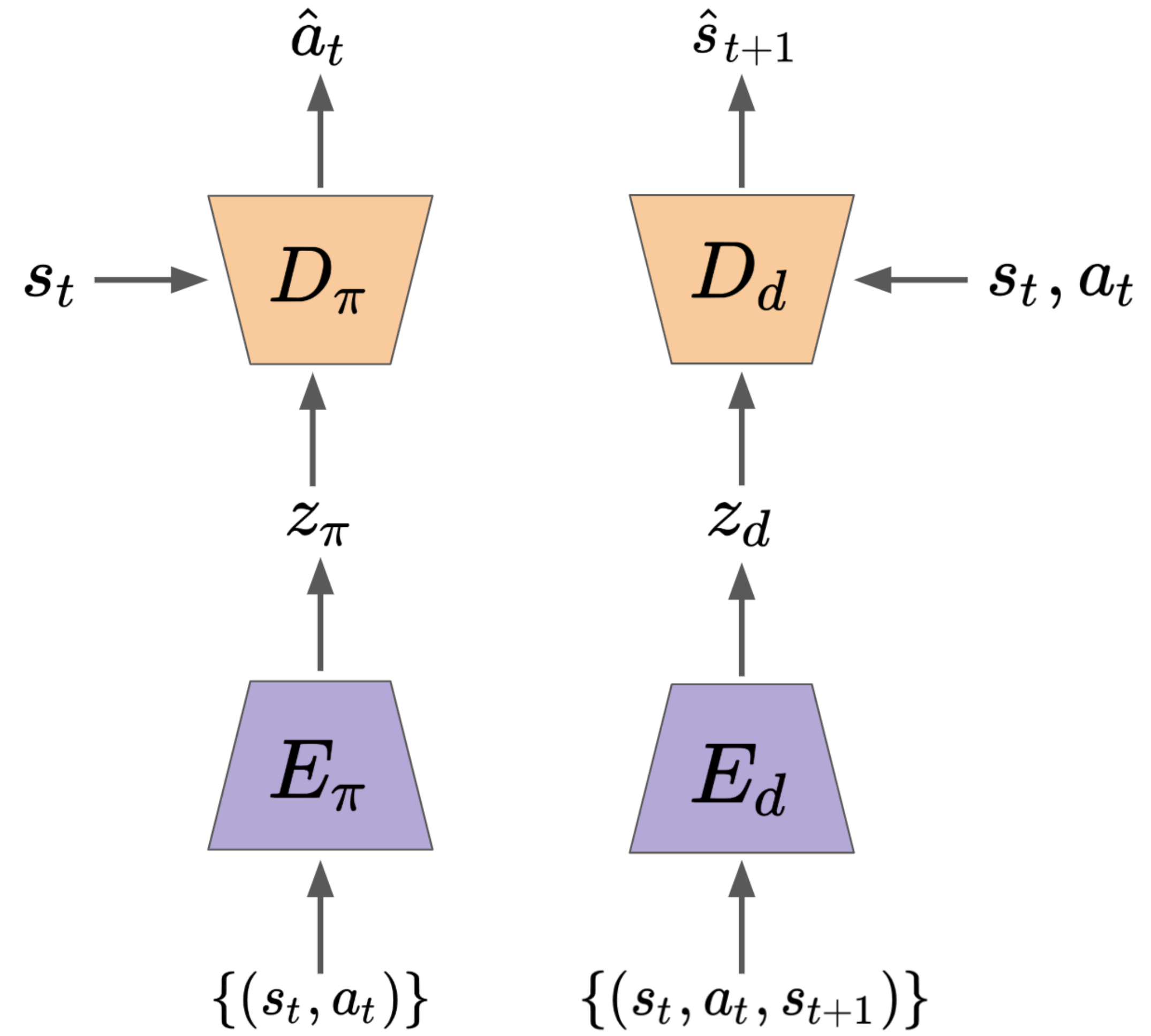}
    \caption{In the \textbf{self-supervised learning phase}, a pair of autoencoders is trained using transitions generated by a diverse set of policies in a set of environments with different dynamics. By exploiting the Markov property of the environment, distinct latent embeddings of the dynamics $z_d$ and policy $z_\pi$ are produced.}   
    \label{fig:emb_diagram}
\end{figure}
The space of dynamics is learned using an encoder $E_{d}$ parameterised as a Transformer \cite{vaswani2017attention}, and a decoder $D_{d}$ parameterised as a feed-forward network. The encoder takes as input a {\it set} of transitions $\{(s_t, a_t, s_{t+1})\}$ from the first $N_d$ steps in each episode and outputs a vector embedding for the dynamics $z_{d}$. The decoder takes as inputs the state $s_t$, action $a_t$ and dynamics embedding $z_{d}$, and predicts the next state $\hat{s}_{t+1}$. The parameters $\theta_d$ and $\phi_d$ of the encoder and decoder are trained to minimize the $\ell_2$ error of $\hat{s}_{t+1}$ and ${s}_{t+1}$. Formally,
\begin{equation*}
    z_{d} = E_{d}(\{(s_t, a_t, s_{t+1})\} ; \; \theta_{d})
\end{equation*}
\begin{equation*}
    \hat{s}_{t+1} = D_{d}(s_t, a_t, z_{d} ;  \; \phi_{d}).
\end{equation*}
This arrangement exploits the inductive bias that, conditioned on $d$, the environment is Markovian. By using no positional encoding in the Transformer, the input transitions lack any temporal ordering, thus preserving the Markov property. The decoder receives no historical information (since it is unnecessary in a Markovian setting), so it is forced to embed information about the dynamics into $z_d$ to make good predictions. Because the input set contains the actions in each triple, the encoder has no incentive to encode policy information into $z_d$. This modeling choice encourages $z_d$ to only contain information about the dynamics, rather than the policy used to generate the transitions.

Similarly, the space of policies is learned using an encoder $E_{\pi}$ parameterised as a Transformer and a decoder $D_{\pi}$ parameterised as a feed-forward network. The encoder takes as input a set (again using the Markov property as an inductive bias) of state-action pairs $\{(s_t, a_t)\}$ from a full episode and outputs a vector embedding for the policy $z_{d}$. The decoder takes as inputs the state $s_t$ and the policy embedding $z_{\pi}$ to predict the action taken by the policy $\hat{a}_{t}$. Since the policy encoder does not have direct access to full environment transitions, $z_{\pi}$ is constrained to capture information about the policy without elements of the dynamics. The parameters $\theta_{\pi}$ and $\phi_{\pi}$ of the encoder and decoder are trained to minimize the $\ell_2$ error of $\hat{a}_{t}$ and ${a}_{t}$. Formally,   
\begin{equation*}
    z_{\pi} = E_{\pi}(\{(s_t, a_t)\} ;  \; \theta_{\pi})
\end{equation*}
\begin{equation*}
    \hat{a}_t = D_{\pi}(s_t, z_{\pi} ;  \; \phi_{\pi}).
\end{equation*}

Both the policy and the dynamics embeddings are normalized to have unit $\ell_2$-norm. 

See Figure~\ref{fig:emb_diagram} for an overview of the self-supervised learning phase.

\begin{figure}[ht!]
    \centering
    \includegraphics[width=0.8\columnwidth]{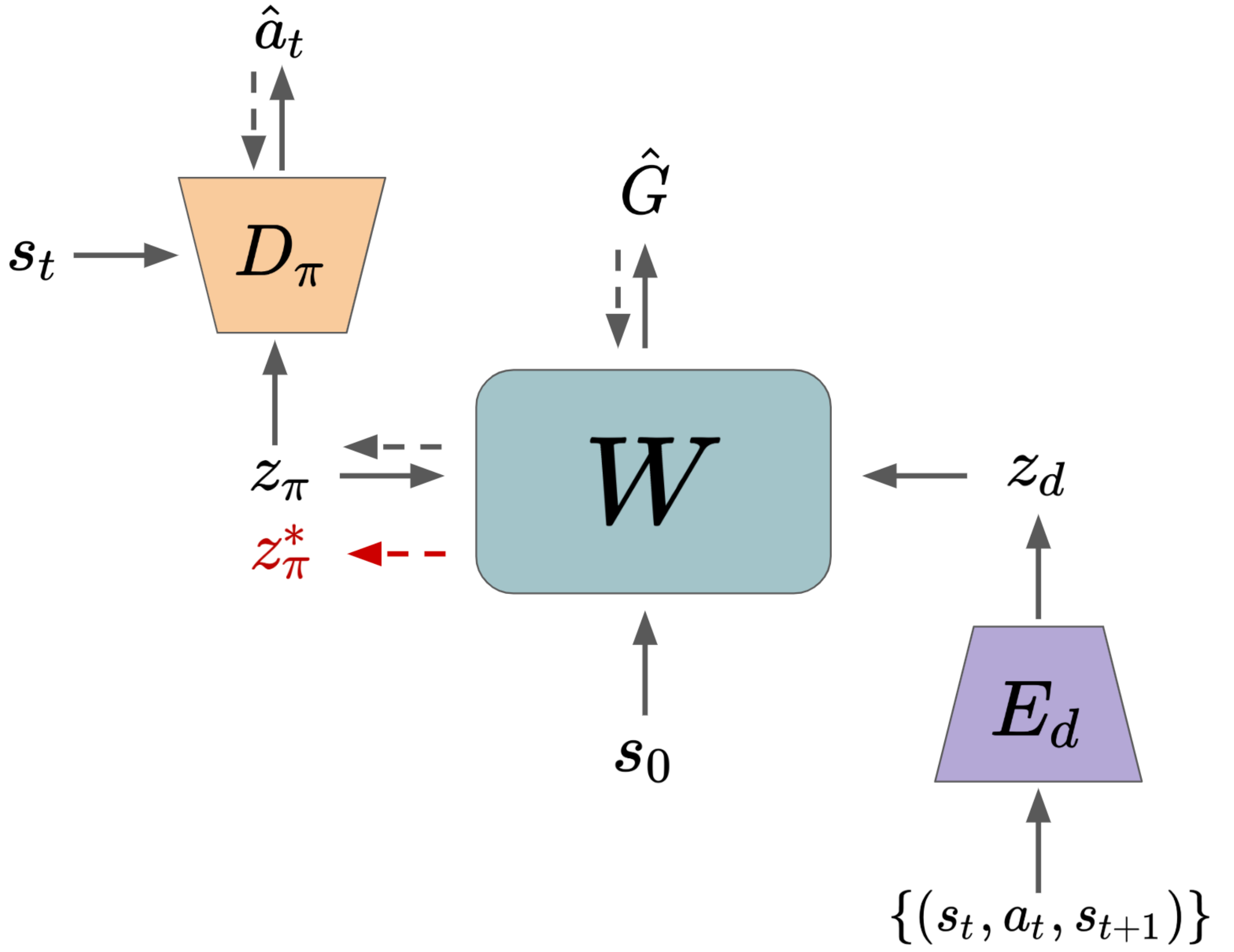}
    \caption{In the \textbf{supervised learning phase}, a parametric value function $W$ is trained to predict the expected return $G$ for an entire space of policies and dynamics. W takes as inputs the initial state $s_0$, policy embedding $z_\pi$, and dynamics embedding $z_d$ (estimated from a small set of transitions). We train $W$ in a supervised fashion, using Monte-Carlo estimates of the expected return $G$ for policy $\pi$ in environment with dynamics $\mathcal{T}_d$. At test time, $z_\pi$ is optimized to maximize $\hat{G}$ (red dashed arrow), resulting in $z^*_\pi$ which is then decoded to an actual policy via $D_\pi$.}
\label{fig:pdvf_diagram}
\end{figure}

\subsection{Supervised Learning Phase}
\label{sec:suptrain}

In this phase, the goal is to train an estimator $W$ of the expected return $\hat{G}$ for a space of policies and dynamics. More specifically, $W$ is a function approximator conditioned on the learned policy and dynamics embeddings, $z_\pi$ and $z_d$.

A central idea of our PD-VF framework is that $W$ provides a scoring function over the policy embedding space. It thus provides a mechanism to allow on-the-fly optimization of $z_\pi$ with respect to the estimated return $\hat{G}$, without the need for any environment interaction, given an estimate (or embedding) of the environment's dynamics. This is key to PD-VF's ability to rapidly find an effective policy in a new environment, only requiring enough environment interaction to give a reliable estimate of the dynamics embedding $z_d$ (just a few steps in practice). We choose $W$ to have a quadratic form to permit easy optimization with respect to $z_\pi$:
\begin{equation*}
    \hat{G} = W(s_0, z_{\pi}, z_{d}) = z_{\pi}^T \, A(s_0, z_{d}; \psi) \, z_{\pi}.
\end{equation*}
The matrix $A(s_0, z_d; \psi)$ is a function of the initial environment state $s_0$ as well as the dynamics embedding $z_d$. Note that $A$ only needs to model the initial state $s_0$ rather than an arbitrary state $s$ since the optimization w.r.t $z_\pi$ occurs only once, at the start of an episode. Since $A$ must be Hermitian positive-definite, a feed-forward network with parameters $\psi$ is first used to obtain a lower triangular matrix $L(s_0, z_d; \psi)$. Then $A$ is constructed from $L L^T$. 

{\noindent \bf Optimizing the policy embedding $z_\pi$}: The optimization of the policy embedding $z_{\pi}$ has a closed-form solution which is achieved by performing a singular value decomposition, $A = U S V^T$, and taking the top singular vector of this decomposition $z_{\pi}^{*}$. Unit $\ell_2$ normalization is then applied to $z_{\pi}^{*}$. We refer to this vector $z_{\pi}^{*}$ as the {\em optimal policy embedding} (OPE) of the \pdvf{}.

{\noindent \bf Learning $\psi$ -- Initial stage}: We collect training data for the \pdvf{} in the following manner. First, we randomly select a policy and an environment from our training set (described in Section~\ref{sec:rltrain}). Second, we generate full trajectories of that policy in the selected environment and cache the average return obtained across all episodes. This gives us a Monte-Carlo estimate for the expected return of the corresponding policy in that particular environment. Then, we use the first $N_{d}$ steps of that trajectory to infer the dynamics embedding. Similarly, we use the full trajectory to infer the policy embedding (via $E_\pi$, not the above optimization procedure). After collecting this data into a buffer, we train the estimator $W$ in a supervised fashion by predicting the expected return $G$ given an initial state $s_0$, a policy embedding $z_{\pi}$ and a dynamics embedding $z_{d}$. 

{\noindent \bf Learning $\psi$ -- Data Aggregation for the Value Function}:
For the method to work well, it is important that the learned value function $W$ makes accurate predictions for the entire policy space, and especially for the OPE $z_{\pi}^{*}$ (which correspond to the policies selected to act in the environment). One way to ensure that these estimates are accurate is by adding the OPEs to the training data. After initial training of the \pdvf{} on the original dataset of policy and dynamics embeddings, we use an iterative algorithm that alternates between collecting a new dataset of OPEs and training the \pdvf{} on the aggregated data (including the original data as well as data added from all previous iterations). We use early stopping to select the best value function (\ie{} the one with the lowest loss) to be used at test time.

{\noindent \bf Learning $\psi$ -- Data Aggregation for the Policy Decoder}:
Similarly, the policy decoder may poorly estimate an agent's actions in states not seen during training. Thus, we iteratively train the policy decoder using a combination of the original set of states as well as new states generated by the policy embeddings that maximize the current value function. More specifically, we use the current OPEs (corresponding to the policies that \pdvf{} thinks are best) as inputs to the policy decoder to generate actions and interact with the environment. Then, we add the states visited by this policy to the data. The policy decoder is trained using the aggregated collection of states which includes both the states visited by the original collection of policies as well as the states visited by the current OPEs selected by the \pdvf{}.

See Figure~\ref{fig:pdvf_diagram} for an overview of the supervised learning phase.

\subsection{Evaluation Phase}

At test time, we want to find a policy that performs well on a single episode of an environment with unseen dynamics. This proceeds as follows: (i) the agent uses one of the pretrained RL policies to act for $N_{d}$ steps; (ii) the generated transitions are then used to infer the dynamics embedding $z_d$; (iii) once an estimate of the dynamics is obtained, the matrix $A(s_0, z_{d};\; \psi)$ can be computed; (iv) we employ the closed-form optimization described above to compute the optimal policy embedding $z_{\pi}^{*}$; (v) the policy decoder, conditioned on the $z_{\pi}^{*}$ embedding, is then used to take actions in the environment until the end of the episode. Note that only a small number of interactions with a new environment is needed in order to adapt, the policy selection being performed internally within the PD-VF model. Performance is evaluated on a single trajectory of each environment instance.

\section{Experiments}
\label{experiments}

\begin{figure}[ht!]
        \subfigure[Spaceship]{\label{fig:space}\includegraphics[width=0.32\columnwidth]{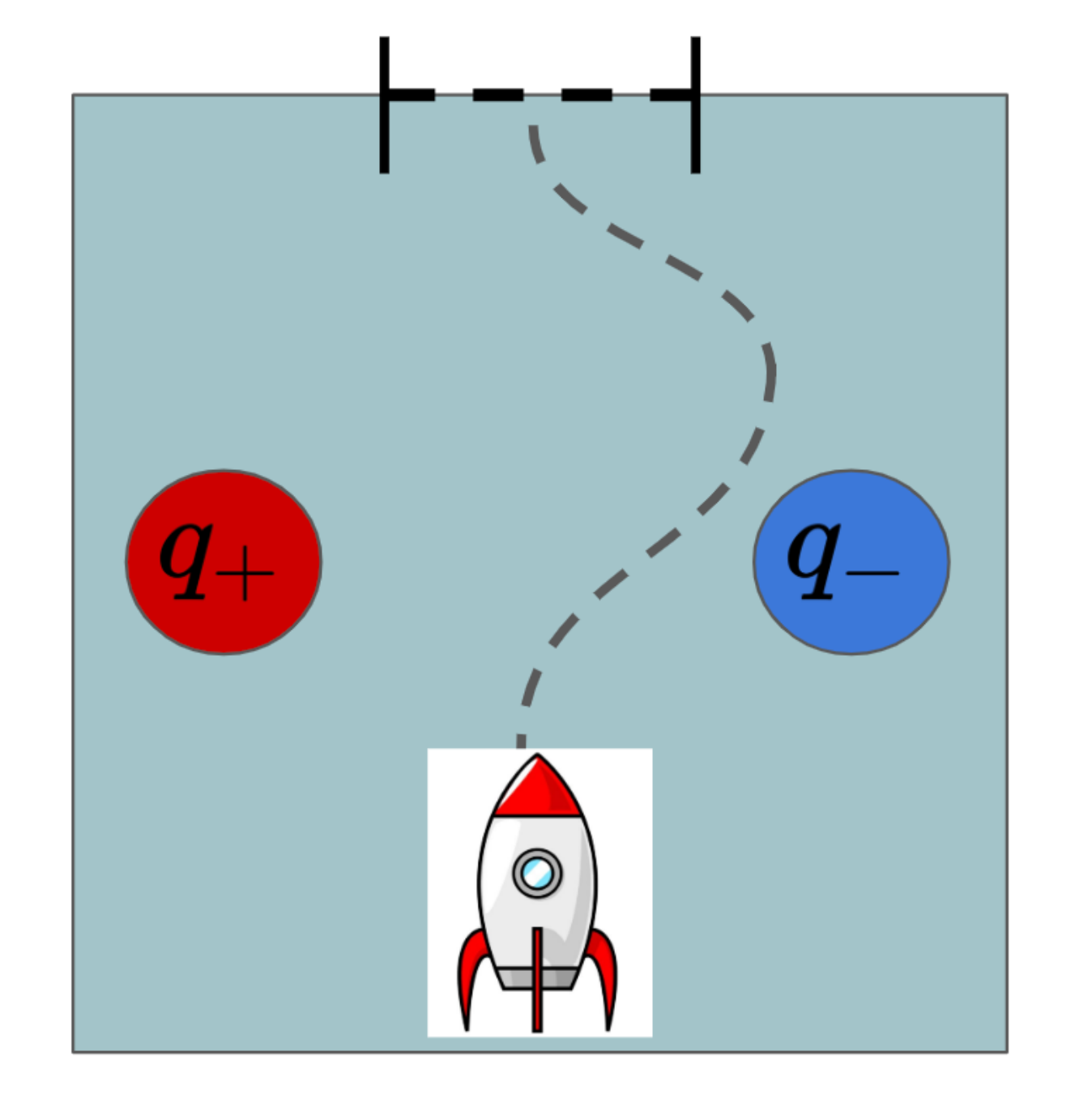}}
        \subfigure[Swimmer]{\label{fig:swim}\includegraphics[width=0.32\columnwidth]{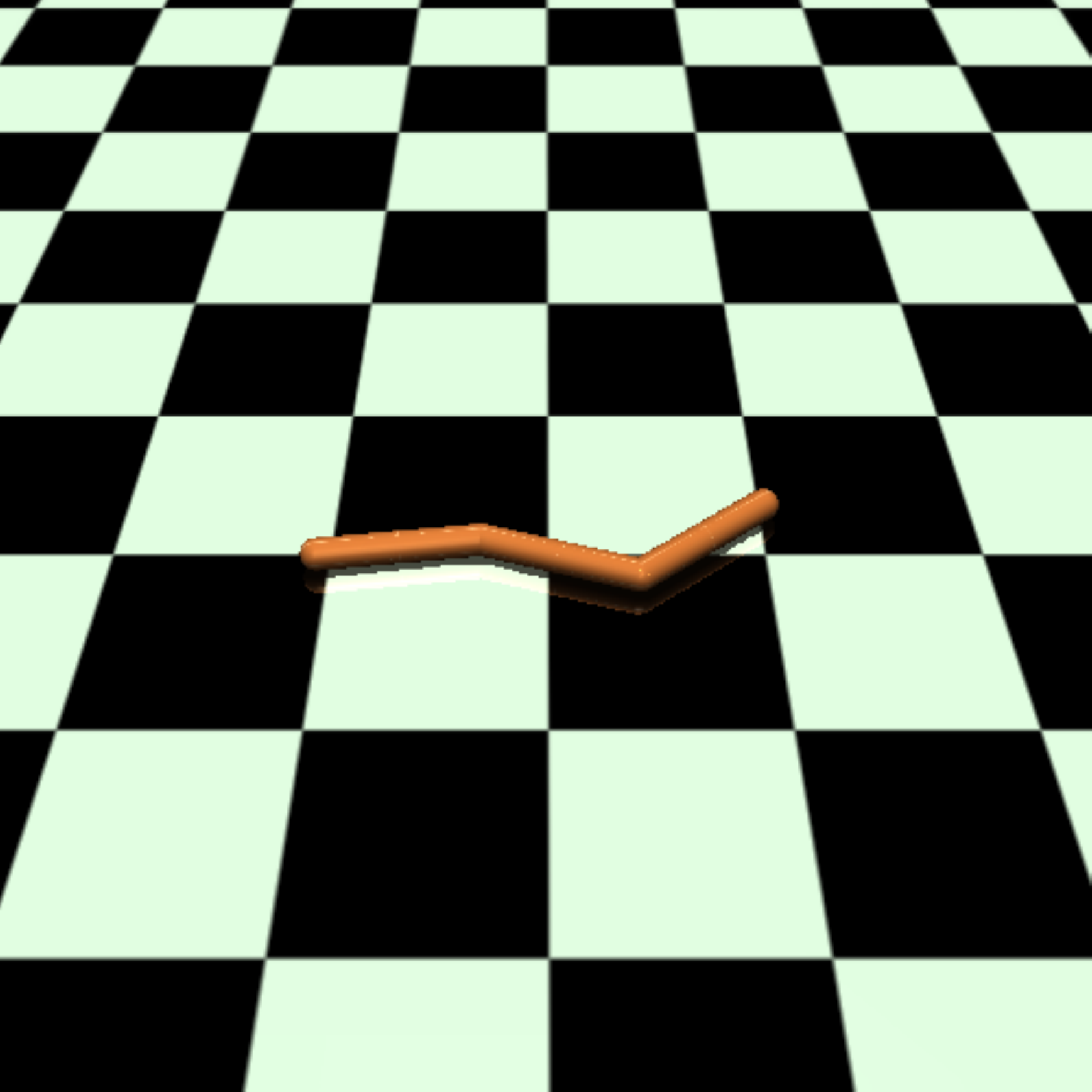}}
        \subfigure[Ant-wind]{\label{fig:antwind}\includegraphics[width=0.32\columnwidth]{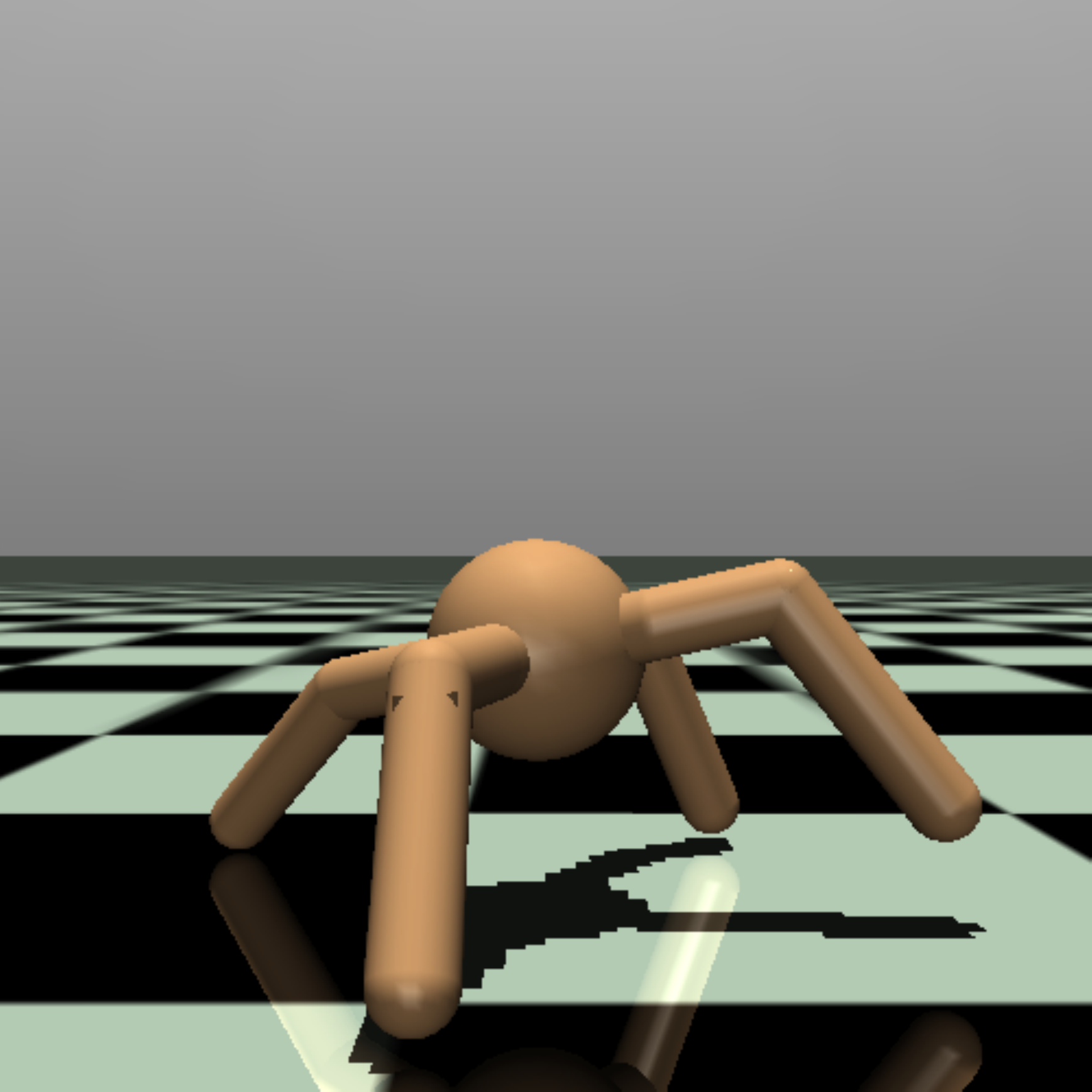}}

        \subfigure[Dynamics]{\label{fig:dynamics}\includegraphics[width=0.32\columnwidth]{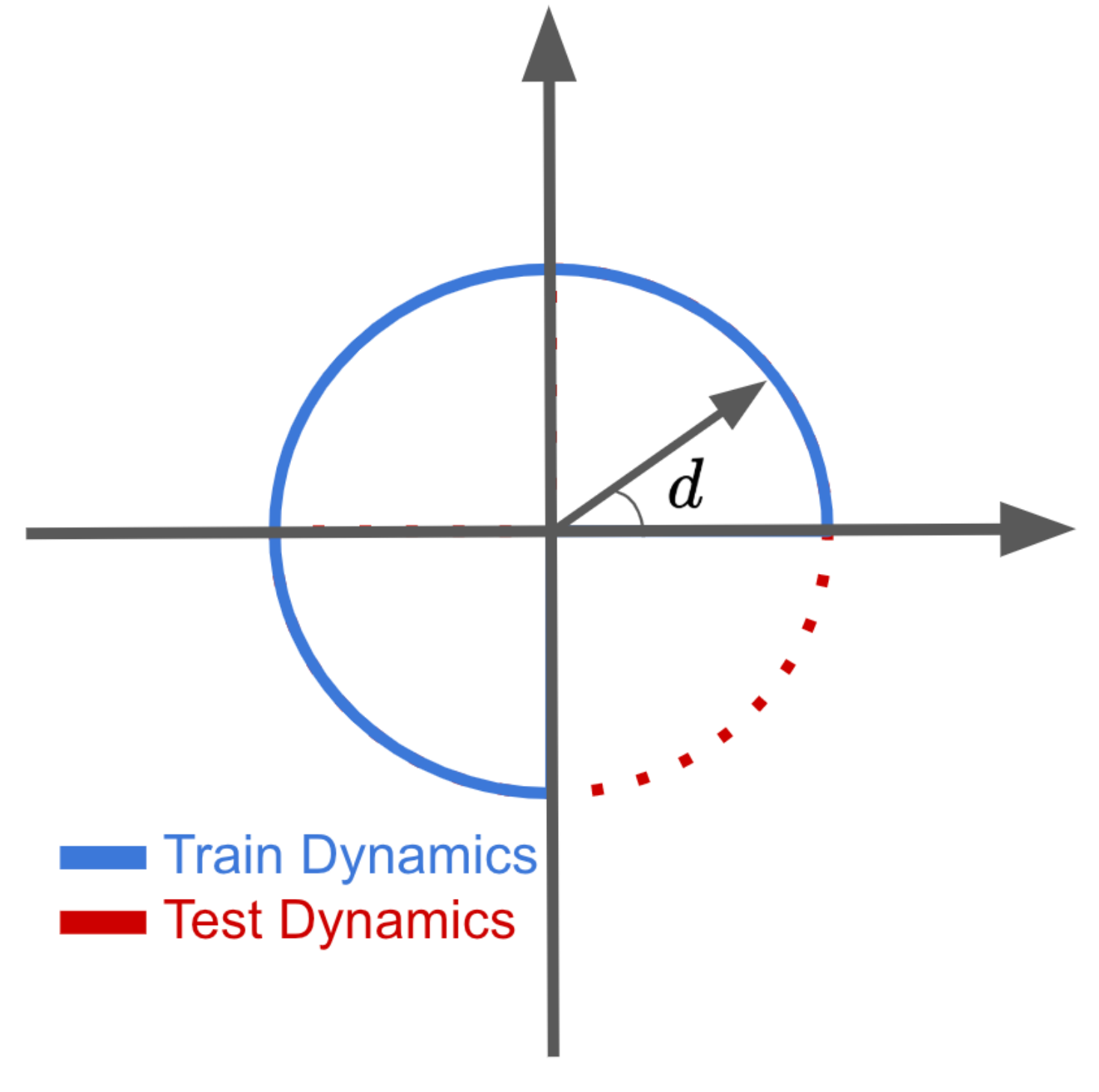}}
        \subfigure[Ant-legs-v1]{\label{fig:antleg1}\includegraphics[width=0.32\columnwidth]{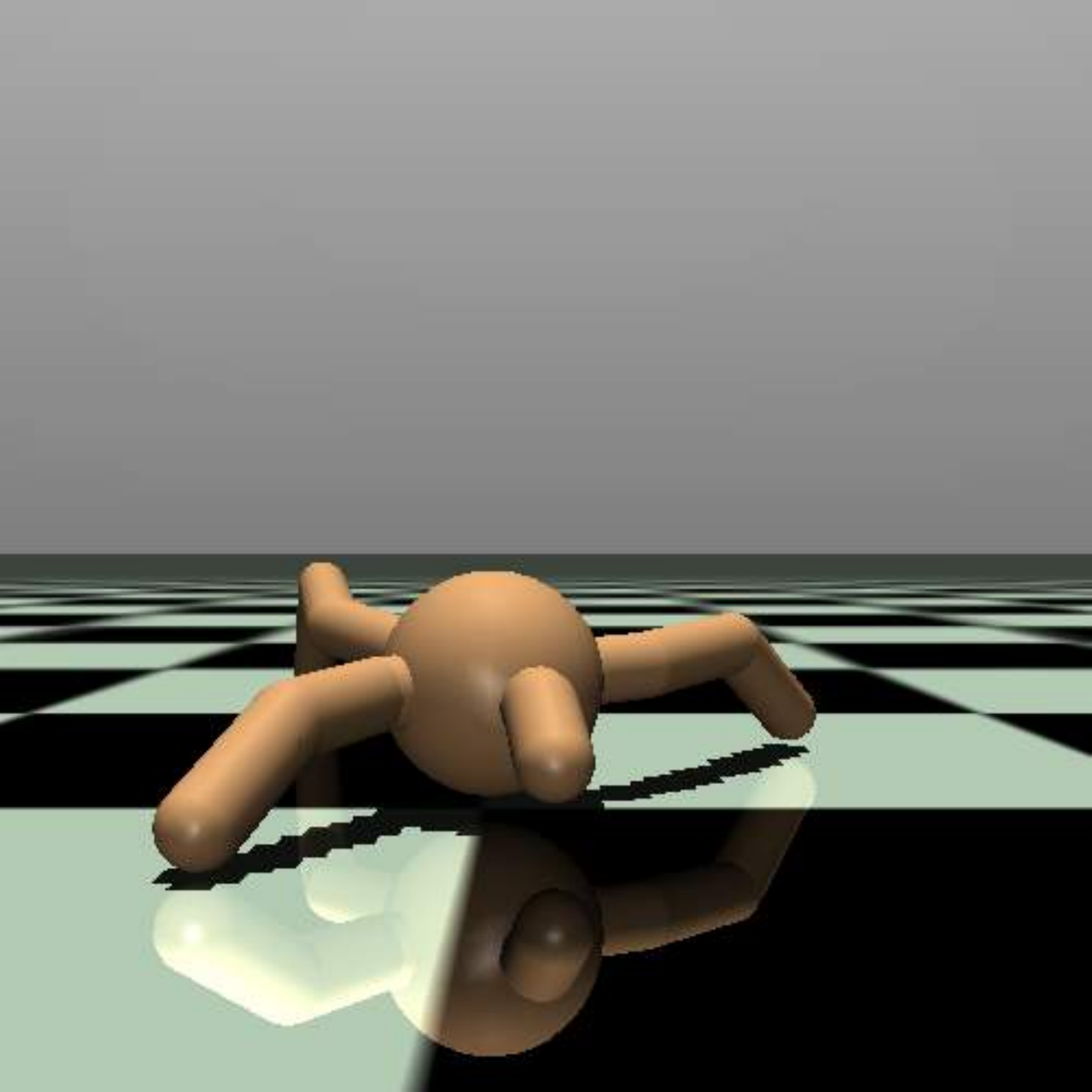}}
        \subfigure[Ant-legs-v2]{\label{fig:antleg2}\includegraphics[width=0.32\columnwidth]{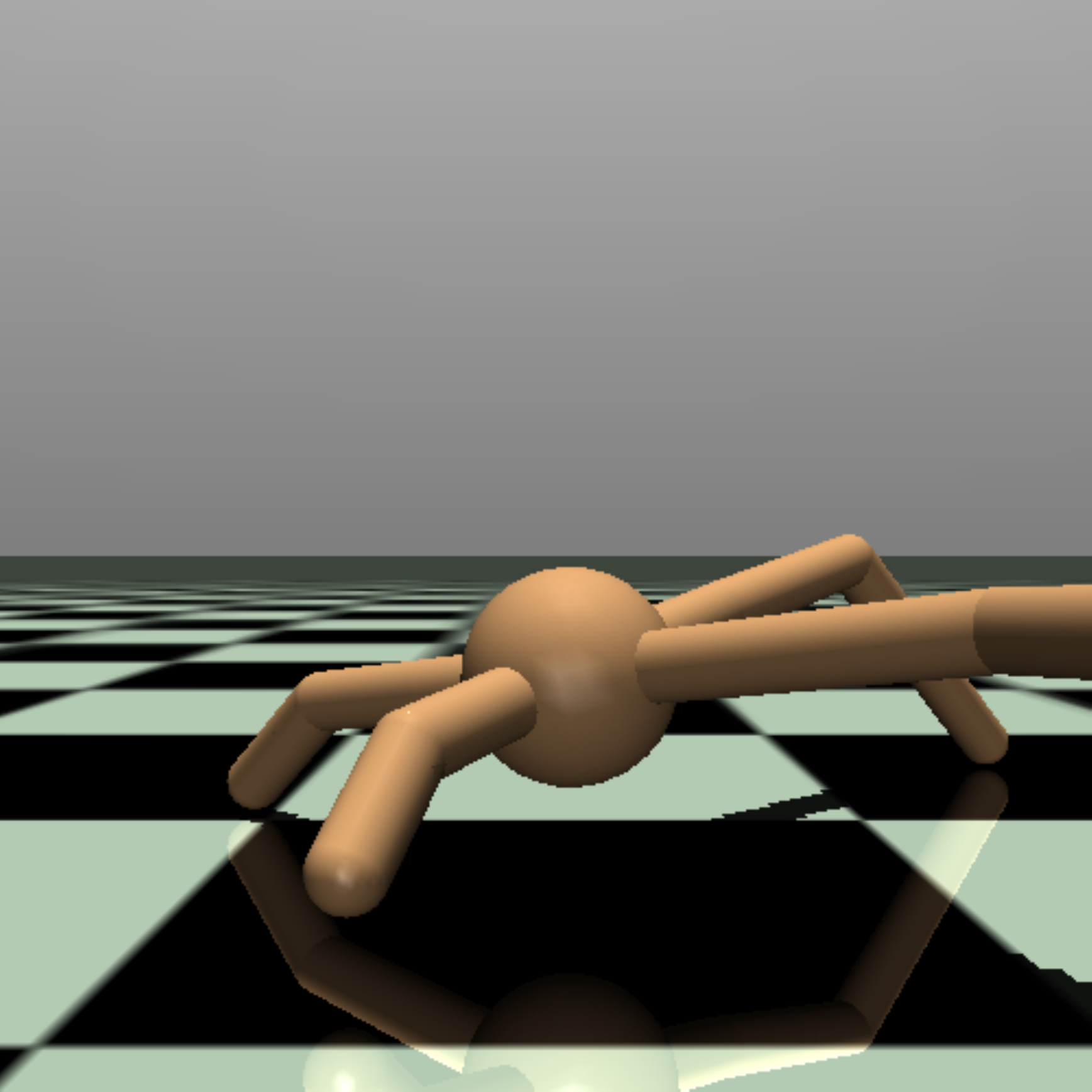}}
    \caption{(a) - (c) illustrate the continuous control domains used for testing adaptation to unseen environment dynamics. In Spaceship, Swimmer, and Ant-wind, the train and test distribution of the dynamics is continuous as illustrated in (d). (e) and (f) show two instances of the Ant-legs task in which limb lengths sampled from a discrete distribution determine the dynamics.}
    \label{fig:diagrams}
\end{figure}

\begin{figure*}[ht!]
        \subfigure{\includegraphics[width=0.50\textwidth]{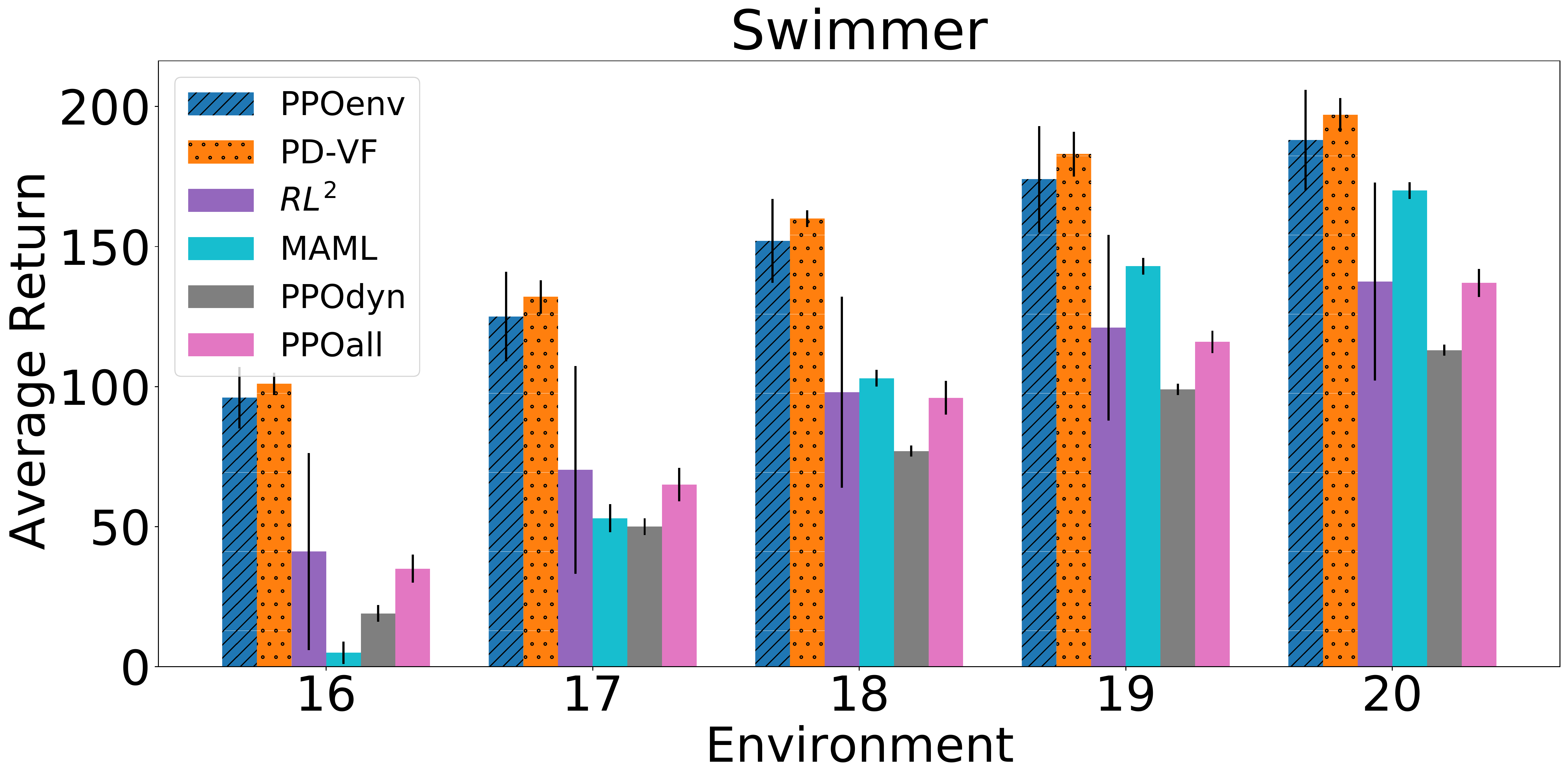}}
        \subfigure{\includegraphics[width=0.50\textwidth]{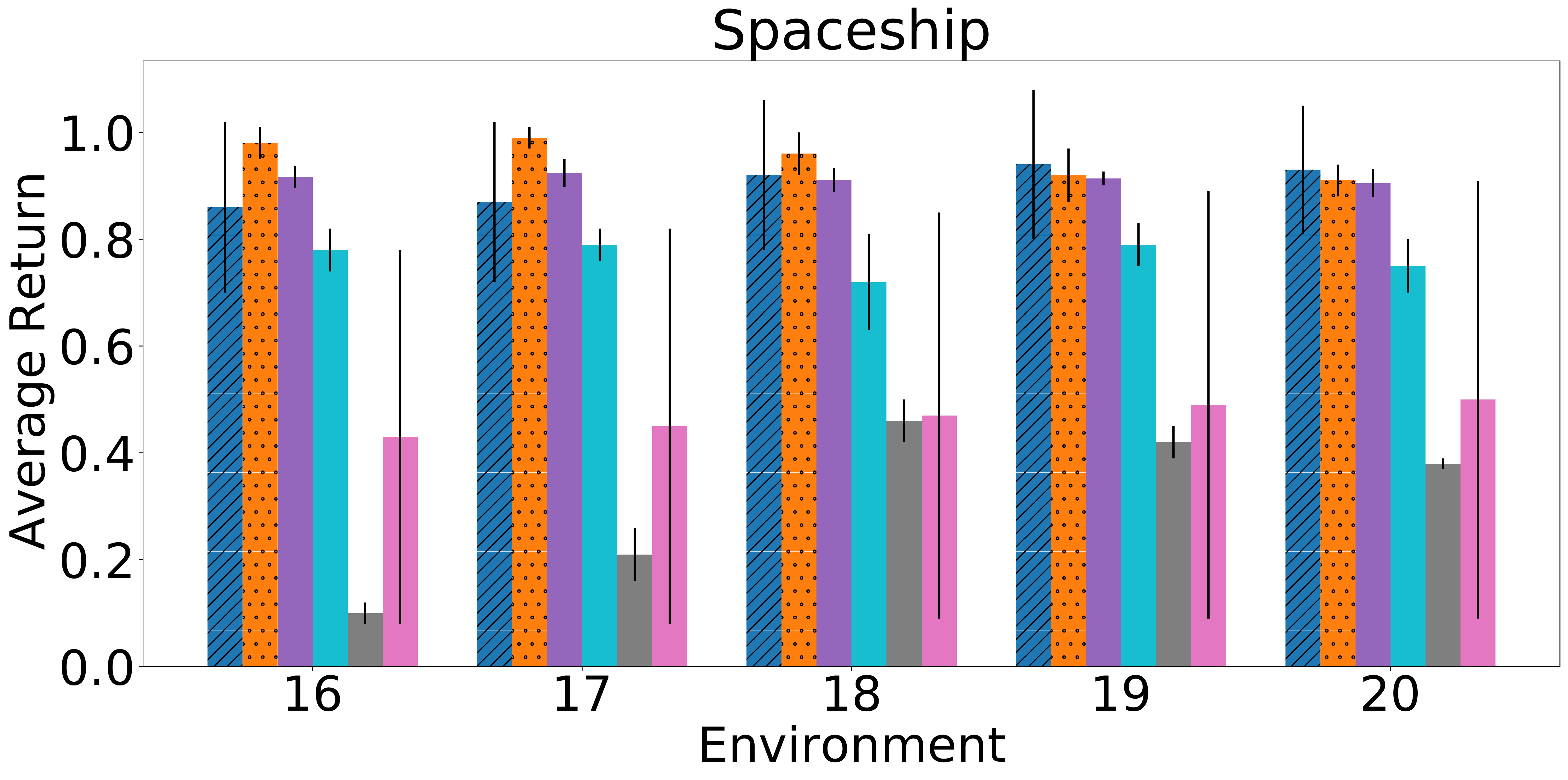}}
       
        \subfigure{\includegraphics[width=0.50\textwidth]{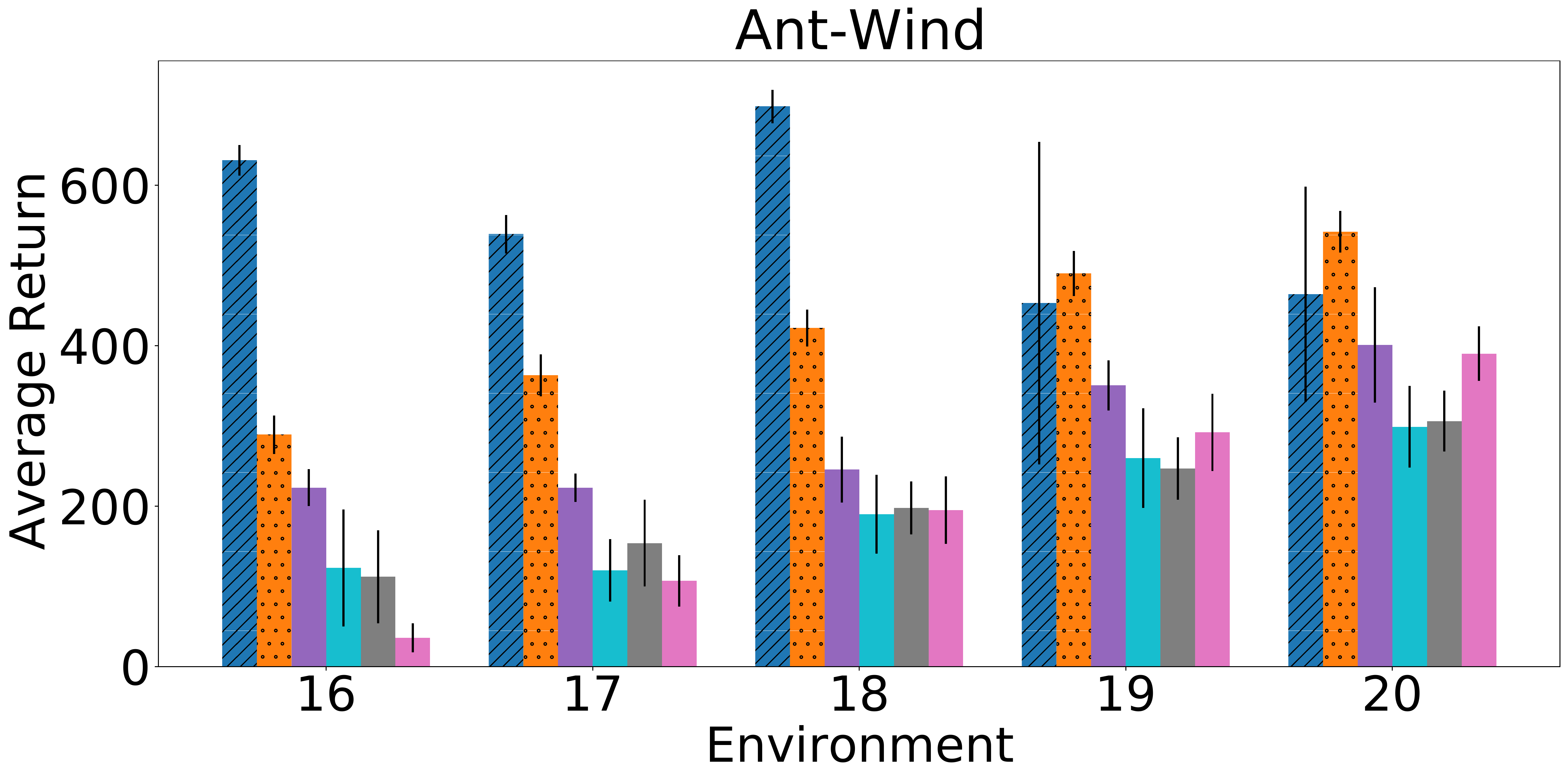}}
        \subfigure{\includegraphics[width=0.50\textwidth]{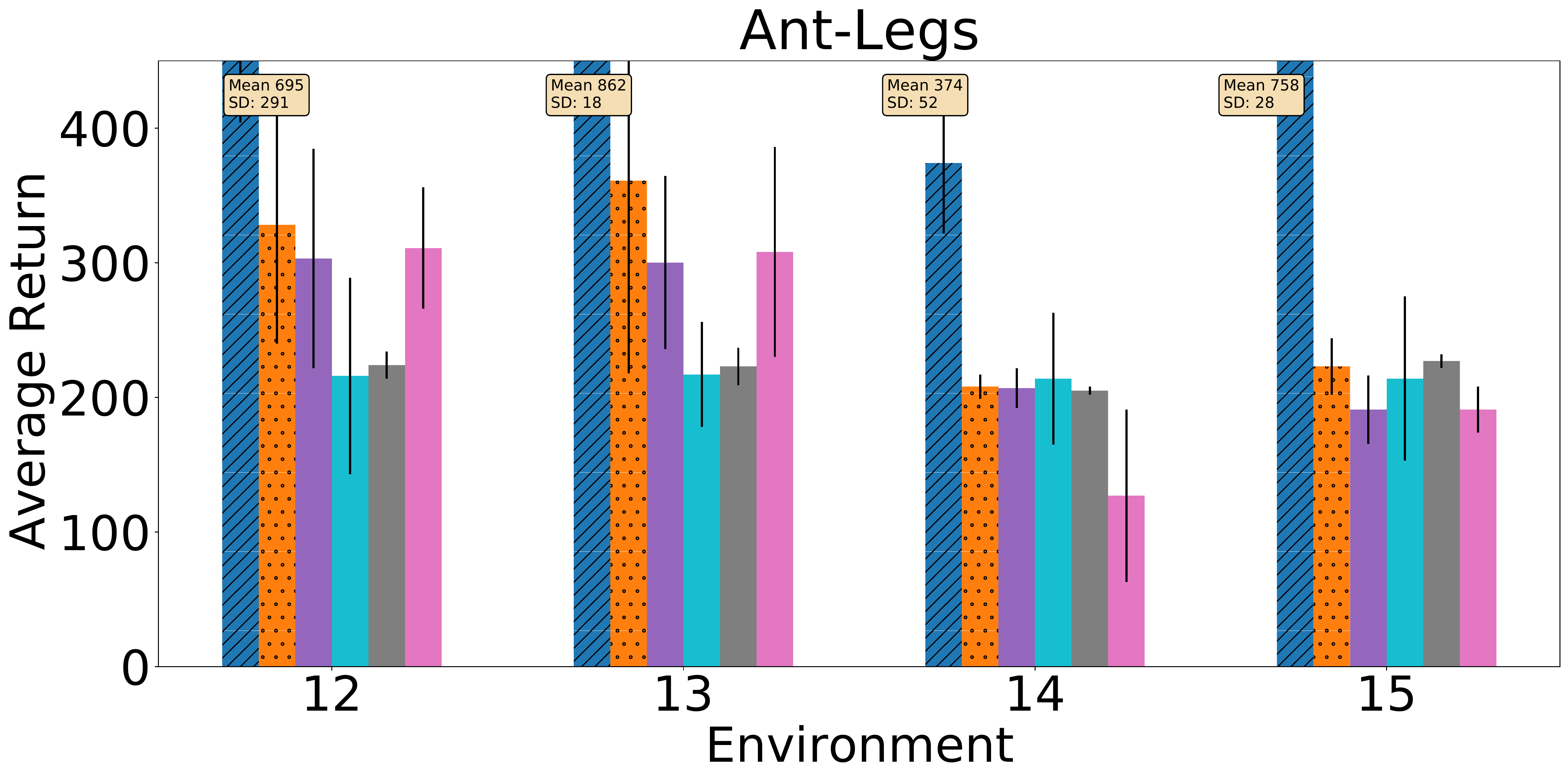}}
    \caption{\textbf{Test Performance.} Average return on test environments with unseen  dynamics in Swimmer (top-left), Spaceship (top-right), Ant-wind (bottom-left), and Ant-legs (bottom-right) obtained by \pdvf{}, the upper bound PPOenv, and baselines $RL^2$, MAML, PPOdyn, and PPOall. \pdvf{} outperforms these baselines on most test environments and, in some cases, it is comparable with PPOenv (which was trained directly on the test environments).}
    \label{fig:eval_results}
\end{figure*}

\subsection{Experimental Setup}
We evaluate \pdvf{} on four continuous control domains, and compare it with an upper bound, four baselines, and four ablations. For each domain, we create a number of environments with different dynamics. Then, we split the set of environments into training and test subsets, so that at test time, the agent has to find a policy that behaves well on unseen dynamics. For all our experiments, we show the mean and standard deviation of the average return (over 100 episodes) across 5 different seeds of each model. The dynamics embeddings are inferred using at most $N_d = 4$ interactions with the environment. 

\subsection{Environments}

\textbf{Spaceship} is a new continuous control domain designed by us. The task consists of moving a spaceship with a unit point charge from one end of a 2D room through a door at the other end. The action space consists of a fixed-magnitude force vector that is applied at each timestep. The room contains two fixed electric charges that deflect / attract the ship as it moves through the environment (see Figure~\ref{fig:space}). The polarity and magnitude of these charges are parameterised by $d$ and determine the environment dynamics. The distribution of dynamics $\mathcal{D}$ is chosen to be circular and centered (see Figure~\ref{fig:dynamics}). Samples $d$ are drawn at intervals of $\pi/10$, each forming a different environment instance with charge  configuration $(\cos(d), \sin(d))$. The 5 samples in the range $[\frac{3}{4} 2\pi, \ldots, 2\pi]$ are held out as evaluation environments, the rest being used for training. 

\begin{figure*}[ht!]
        \subfigure{\includegraphics[width=0.50\textwidth]{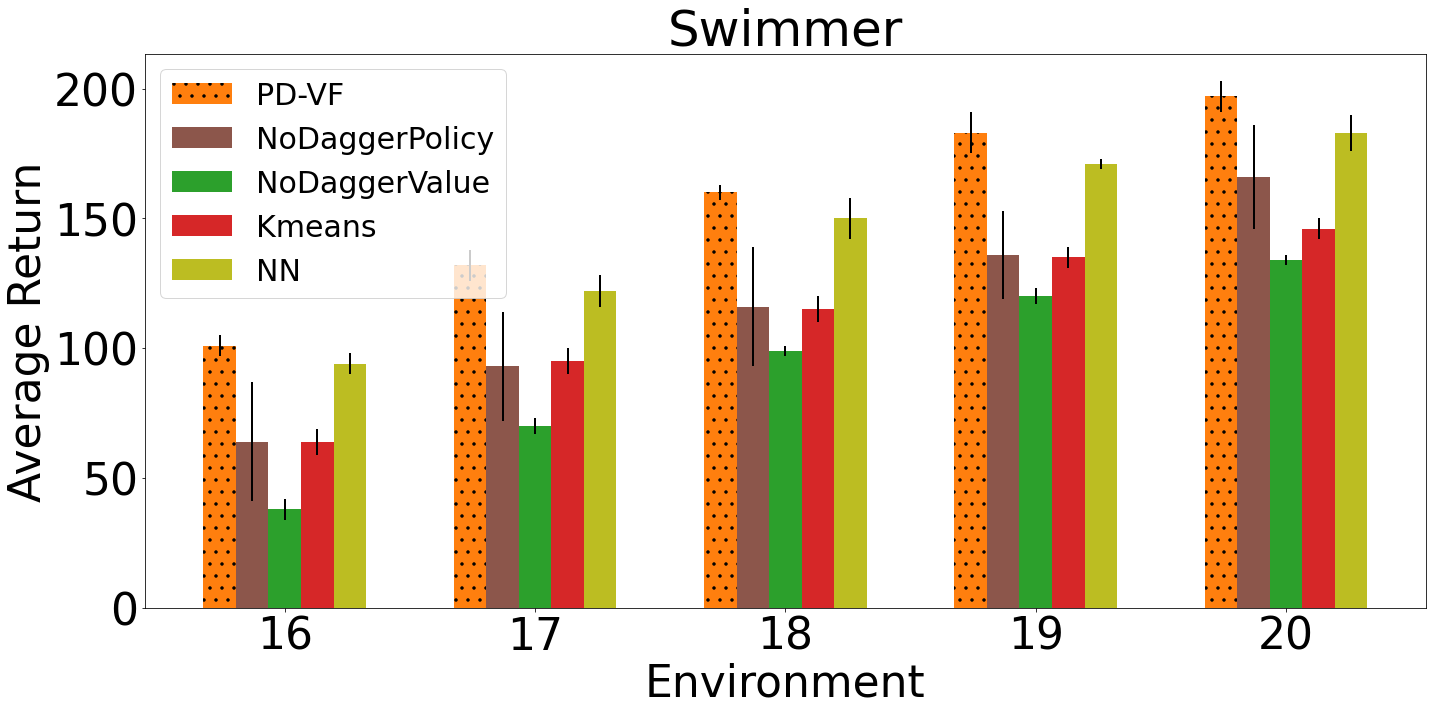}}
        \subfigure{\includegraphics[width=0.50\textwidth]{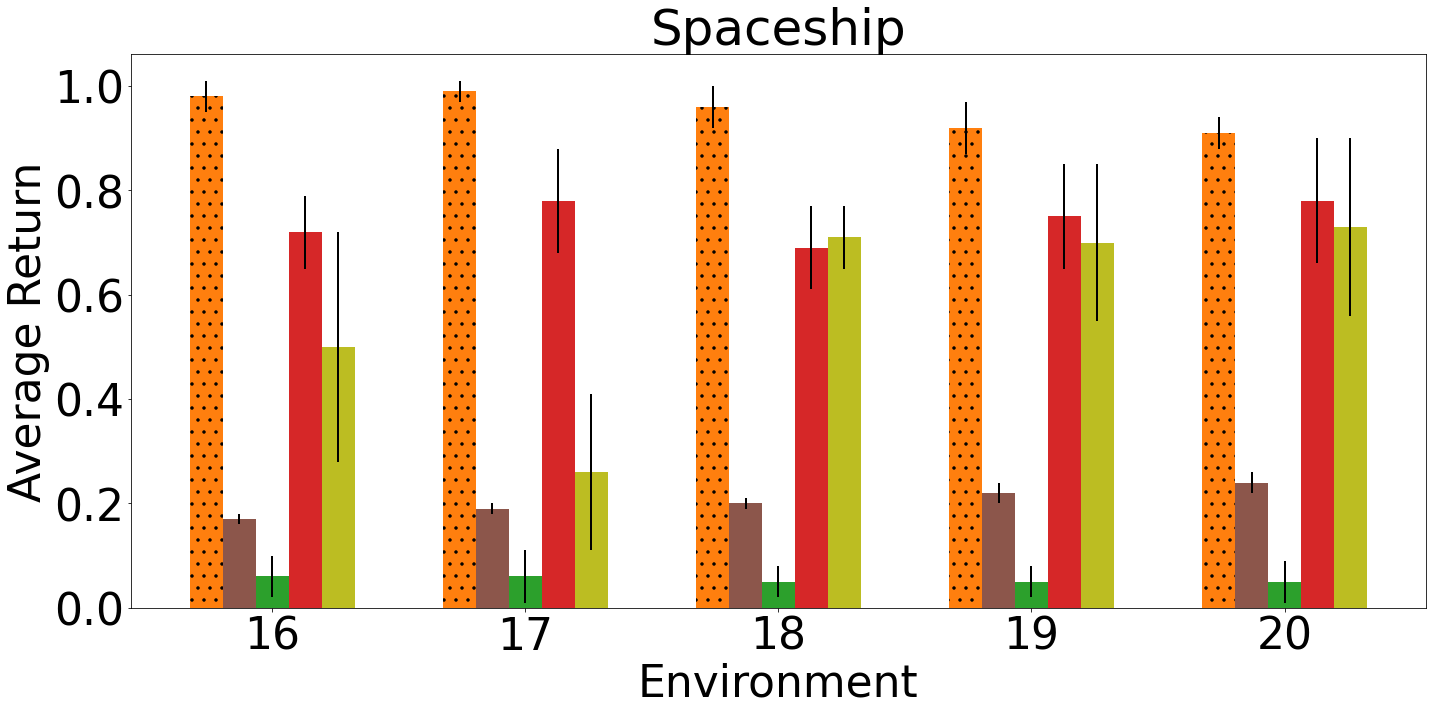}}
       
        \subfigure{\includegraphics[width=0.50\textwidth]{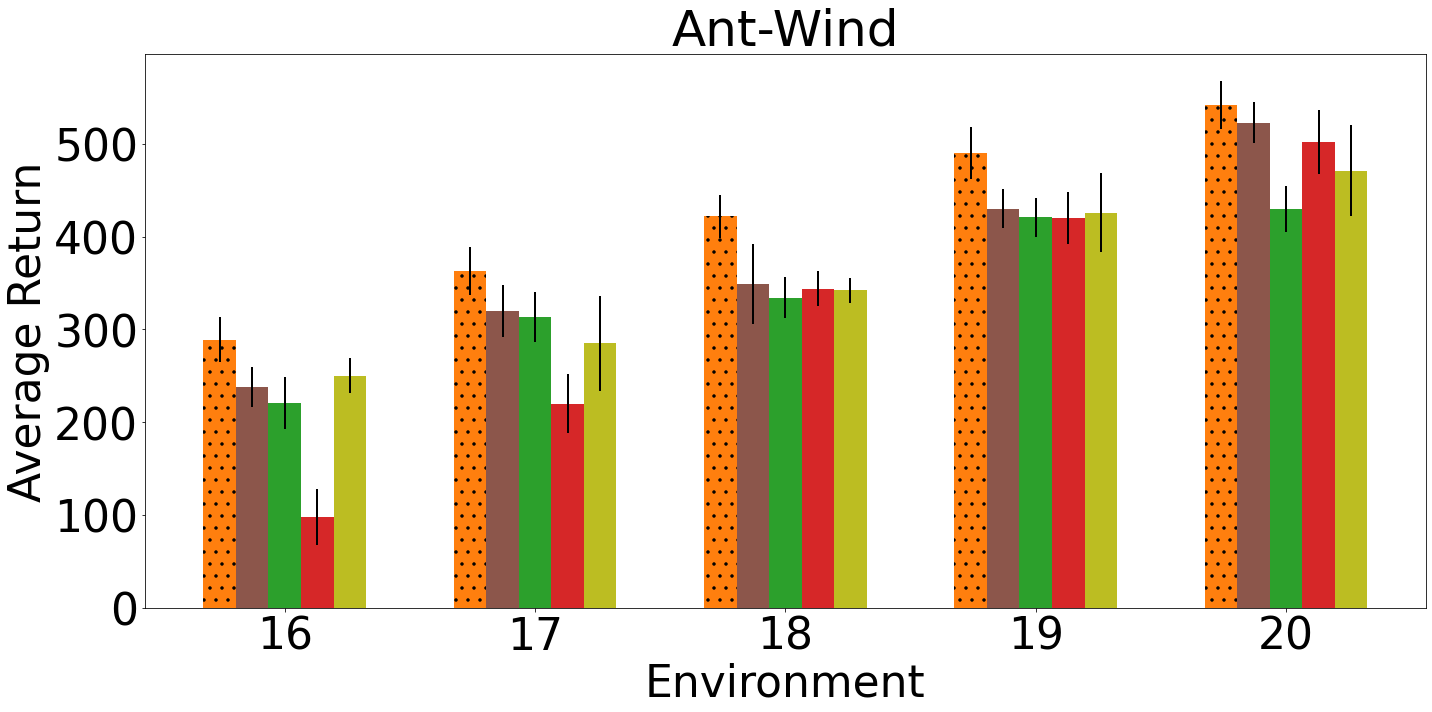}}
        \subfigure{\includegraphics[width=0.50\textwidth]{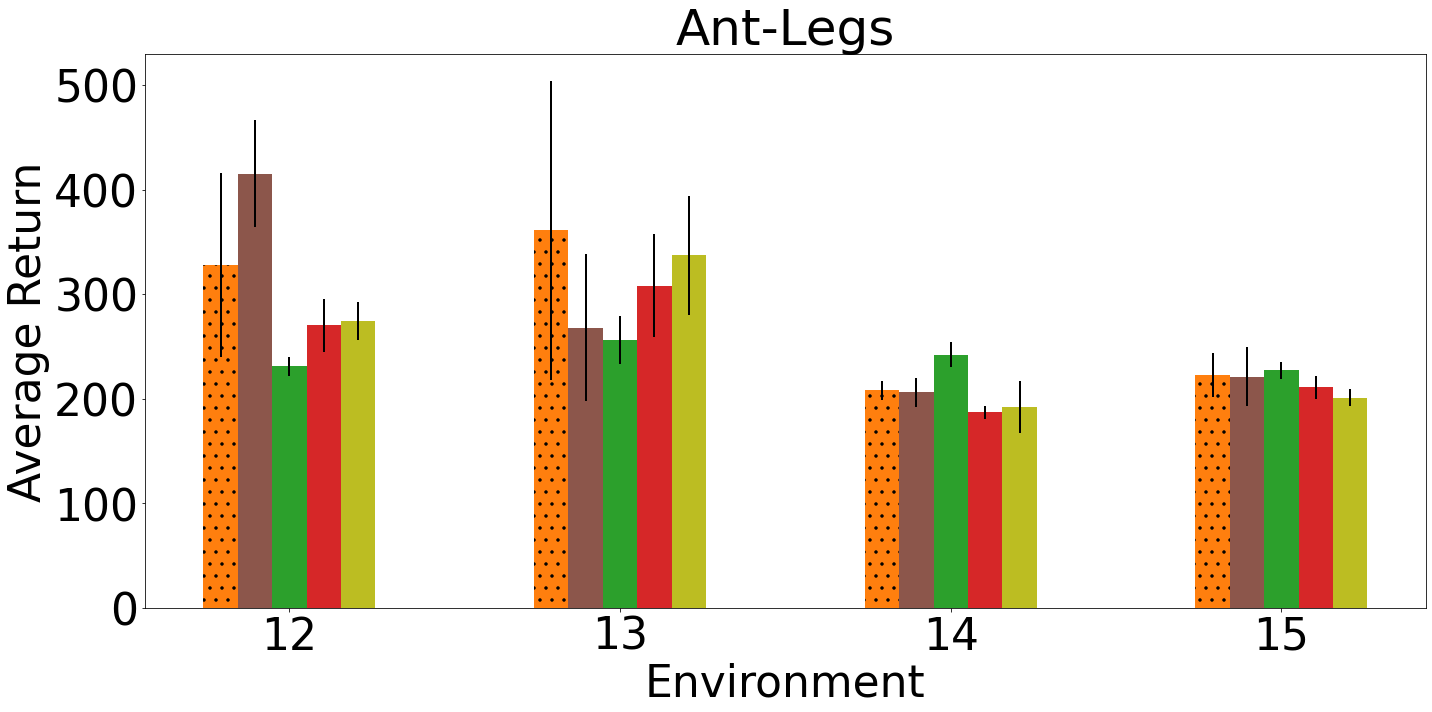}}
    \caption{\textbf{Test Performance.} Average return in Swimmer (top-left), Spaceship (top-right), Ant-wind (bottom-left), and Ant-legs (bottom-right) obtained by \pdvf{}, NoDaggerPolicy, NoDaggerValue, Kmeans, and NN. \pdvf{} is better than these ablations overall.}
    \label{fig:eval_ablations}
\end{figure*}

\textbf{Swimmer} is a family of environments with varying dynamics based on MuJoCo's Swimmer-v3 domain \citep{Todorov2012MuJoCoAP}. The goal is to control a three-link robot in a viscuous fluid to swim forward as fast as possible (Figure~\ref{fig:swim}). The dynamics are determined by a 2D current within the fluid, whose direction changes between environments (but has fixed magnitude). The current direction is determined by an angle $d$, which is sampled in the same manner as for Spaceship above, \ie{} train on $3/4$ of all possible directions and hold out the other $1/4$ for evaluation.

\textbf{Ant-wind} is a family of environments based on MuJoCo's Ant-v3 domain in which the goal is to make a four-legged creature walk forward as fast as possible (Figure~\ref{fig:antwind}). The environment dynamics are determined by the direction of a wind $d$, which is sampled from a continuous distribution in the same way as for Swimmer.

\textbf{Ant-legs} is a second task based on MuJoCo's Ant-v3 domain, in which the dynamics are sampled from a discrete distribution. The training environments are generated by fixing three ankle lengths (short, medium, and long) and generating all possible permutations for the four legs. The length of the ant leg is fixed to medium across all training environments. Symmetries in the training environments are removed by considering ants with the same number of short, medium, or long legs to be the same and choosing one ant from each equivalency class. There are four test environments with both the leg and ankle lengths being either short or long. Note that the test environments are significantly different from all the training ones, thus making Ant-legs a challenging setting for our method. Figures~\ref{fig:antleg1} and~\ref{fig:antleg2} show two instances of this environment.

\subsection{Baselines}
We use PPO \citep{schulman2017proximal} as the base RL algorithm for all the baselines and for the reinforcement learning phase of training the \pdvf{} (Sec.~\ref{sec:rltrain}). We use Adam \citep{kingma2014adam} for optimization. All models use the same network architecture for the policy and value functions. For a given environment, all methods use the same number of steps $N_d$ (at the beginning of each episode) to infer the embedding of the environment dynamics. Then, they each use a single policy network to act in the environment until the end of the episode. We report the cumulative reward obtained by each method throughout an episodes (in which they first infer the environment dynamics which determines the policy used for acting until the end of the episode). We compare with the following baselines:

\begin{figure*}[h!]
        \subfigure{\includegraphics[width=0.50\textwidth]{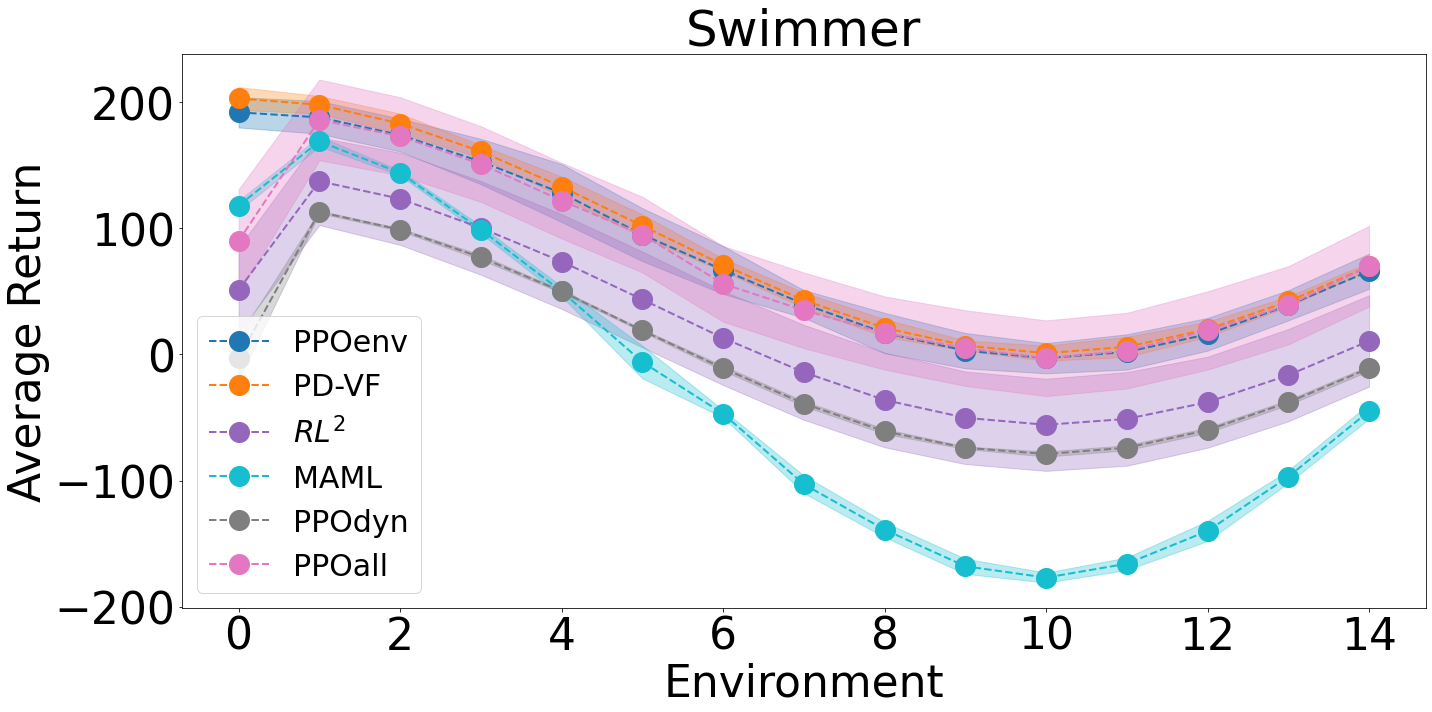}}
        \subfigure{\includegraphics[width=0.50\textwidth]{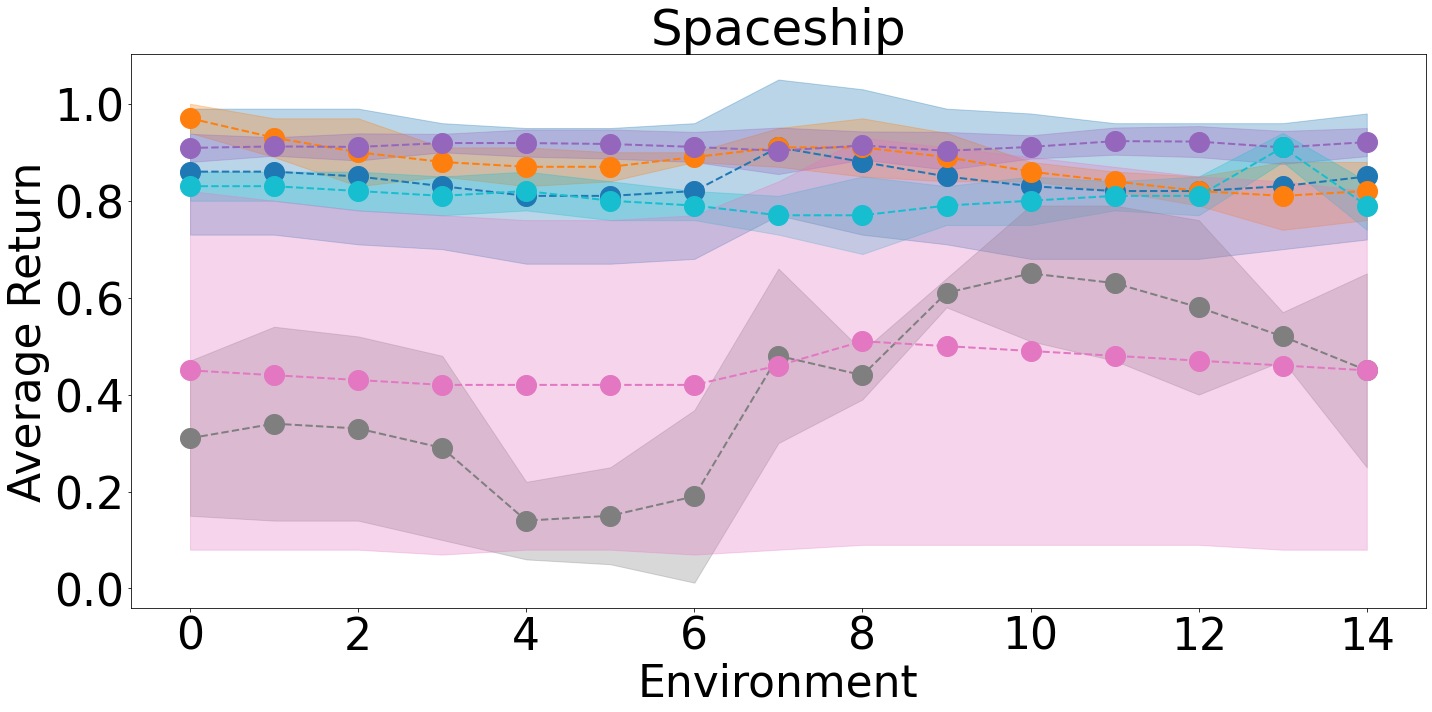}}
       
        \subfigure{\includegraphics[width=0.50\textwidth]{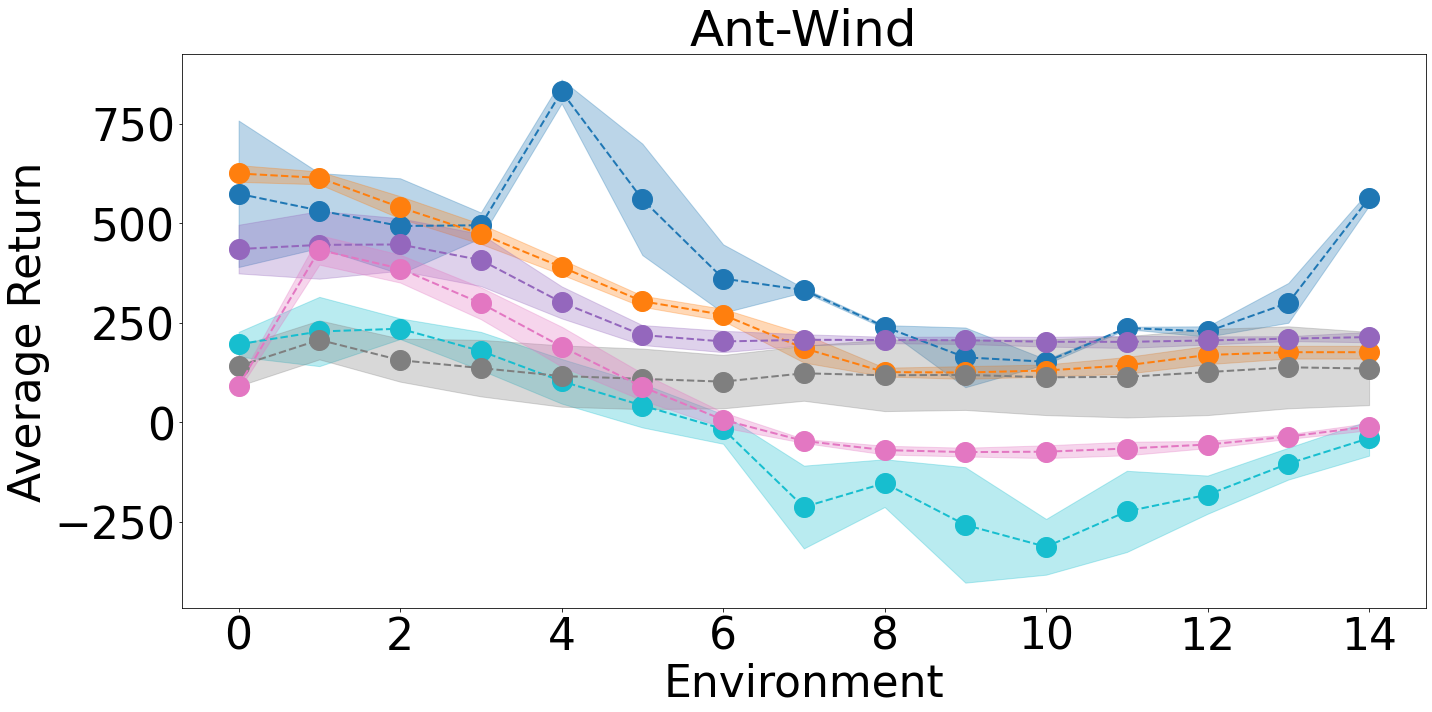}}
        \subfigure{\includegraphics[width=0.50\textwidth]{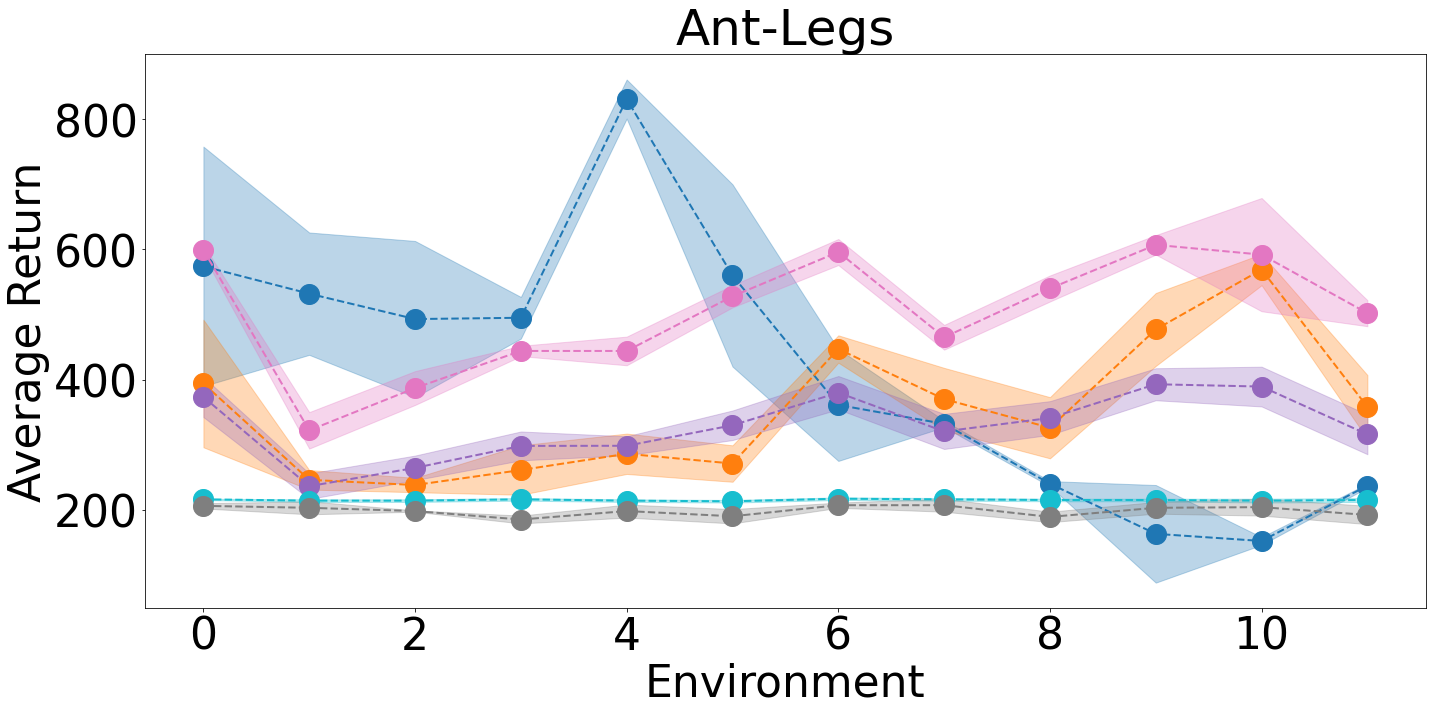}}
     \caption{\textbf{Train Performance.} Average return on train environments in Swimmer (top-left), Spaceship (top-right), Ant-wind (bottom-left), and Ant-legs (bottom-right) obtained by \pdvf{}, the upper bound PPOenv, and baselines $RL^2$, MAML, PPOdyn, and PPOall. \pdvf{} outperforms the baselines and ablations on most test environments and, in some cases, it is comparable with PPOenv (which was trained directly on the test environments). While other methods also perform reasonably well on the training environments, they generalize poorly to new environments with unseen dynamics.}
    \label{fig:train_results}
\end{figure*}

\textbf{\ppoenv} trains a PPO policy for each environment in our set. This is used as an upper bound for the other models. 

\textbf{\maml} is the meta-learning algorithm from \citet{finn2017model}. \maml{} generally requires some amount of training on the test environments, so to make it more comparable to our method and the other baselines, we allow one gradient step using a trajectory of length $N_d$ (i.e. the same length as the one used by \pdvf{} to infer the embedding of the environment dynamics). Thus, \maml{} has an advantage over \pdvf{} which does not make any parameter updates at test time.

$\textbf{RL}^2$ is the meta-learning algorithm from \citet{Wang2016LearningTR} and \citet{Duan2016RL2FR}, which uses a recurrent policy that takes as input the previous action and reward.

\textbf{\condpi} trains (using PPO) a single policy network conditioned on the dynamics embedding. At test time, it first infers the dynamics embedding and then conditions the pretrained policy network on that vector. This is a close implementation of the approach in \citet{yang2019single}\footnote{An exact match was not feasible as code for \citet{yang2019single} was not available.}. 

\textbf{\ppoall} trains a single PPO policy on all the training environments and uses it on the test environments without any additional fine-tuning. 

We also compare \pdvf{} with four ablations:

\textbf{\nn} finds the environment that is closest (in Euclidean metric) to the test environment's embedding and uses the \ppoenv{} policy trained on that environment to act. This ablation aims to tease out the effect of using both the learned space of policies and that of dynamics to adapt to new environments, from that of only using the learned dynamics space.

\textbf{\kmeans} clusters the environment embeddings (using trajectories collected in Section~\ref{sec:rltrain}) into $K$ clusters. Then, for each cluster, we train a new PPO policy on all the environments assigned to that cluster. At test time, we find the closest cluster for the given environment embedding and use the policy corresponding to that cluster to act in the environment.

\textbf{\ndv} trains a \pdvf{} without using dataset aggregation for the value function (see Section~\ref{sec:suptrain}).

\textbf{\ndp} uses \pdvf{} without using dataset aggregation for the policy decoder (see Section~\ref{sec:suptrain}).

\section{Results}
\label{results}

\subsection{Adaptation to New Environment Dynamics}
As seen in Figures~\ref{fig:eval_results} and~\ref{fig:eval_ablations}, \pdvf{} outperforms all other methods on test environments with new dynamics. In some cases (particularly on Spaceship and Swimmer), our approach is comparable to the PPOenv upper bound which was directly trained on the respective test environment (in contrast, \pdvf{} has never interacted with that environment before). While the strength of \pdvf{} lies in quickly adapting to new dynamics, its performance on training environments is still comparable to that of the other baselines, as shown in Figure~\ref{fig:train_results}. This result is not surprising since current state-of-the-art RL algorithms such as PPO can generally learn good policies for the environments they are trained on, given enough interactions, updates, and the right hyperparamters. However, as predicted, standard model-free RL methods such as the baseline PPOall do not generalize well to environments with dynamics different from the ones experiences during training. Even meta-learning approaches like MAML or $RL^2$ struggle to adapt when they are allowed to use only a short trajectory for updating the policy at test time, as is the case here. 

But most importantly, \pdvf{} also outperforms the approaches that use the dynamics embedding such as NN, Kmeans, and PPOdyn. This supports our claim that learning a value function for an entire space of policies (rather than for a single optimal policy as standard RL methods do) can be beneficial for adapting to unseen dynamics. By simultaneously estimating the return of a collection of policies in a family of environments with different but related dynamics, \pdvf{} can learn how variations in dynamics relate to differences in the performance of various policies. This allows the model to rank different policies and understand that sub-optimal behaviors in certain environments might be optimal in others. Thus, at least in theory, \pdvf{} has the ability to find policies that are better than the ones seen during training. Our empirical results indicate that this might also hold true in practice. Overall, \pdvf{} proves to be more robust to changes in dynamics relative to the other methods, especially in completely new environments.

\subsection{Analysis of Learned Embeddings}

The performance of \pdvf{} relies on learning useful policy and dynamics embeddings that capture variations in agent behaviors and transition functions, respectively. In this section, we analyze the learned embeddings. 

\begin{figure}[ht!]
        \subfigure[]{\label{fig:env_space}\includegraphics[width=0.31\columnwidth]{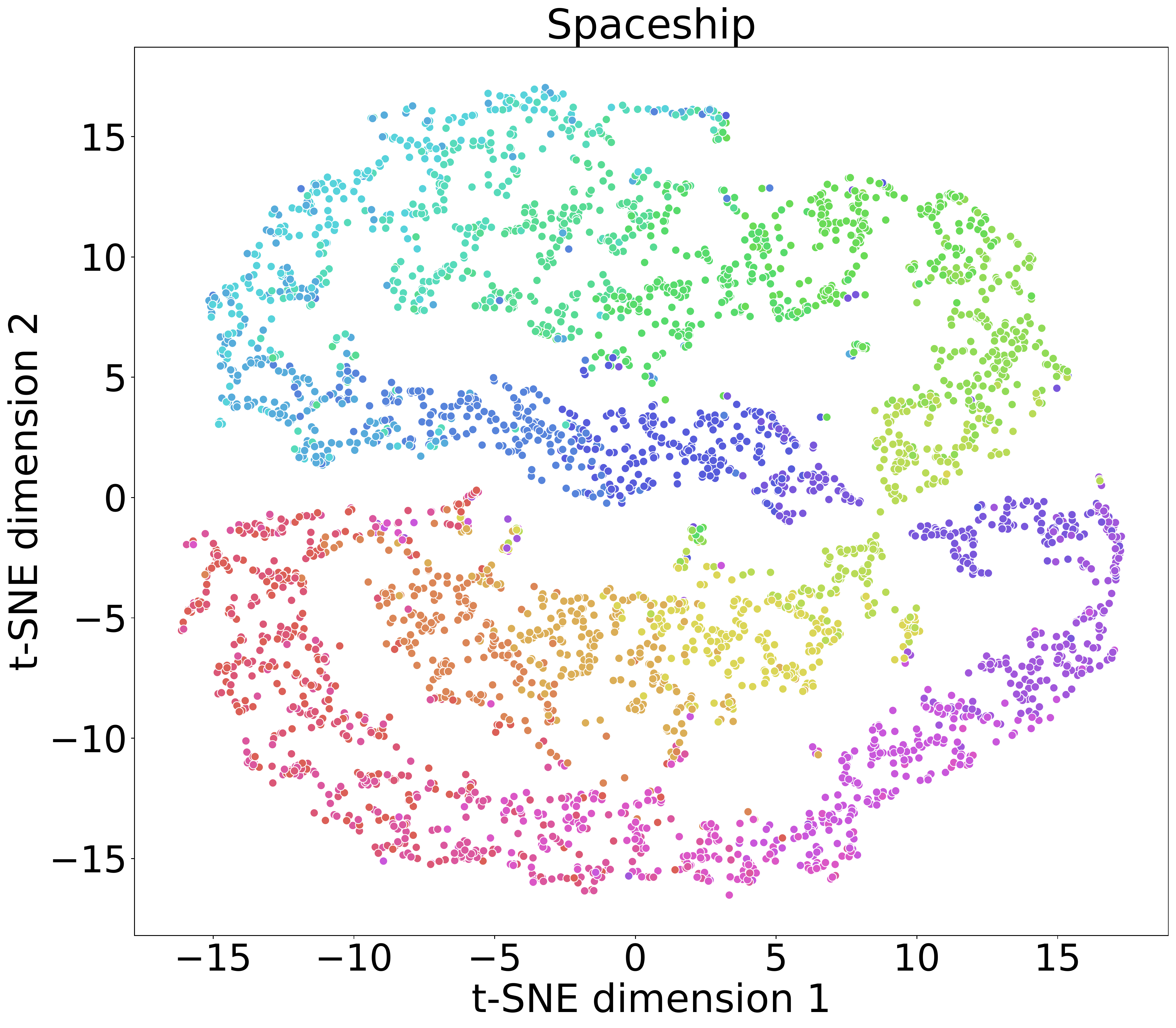}}
        \subfigure[]{\label{fig:env_swim}\includegraphics[width=0.31\columnwidth]{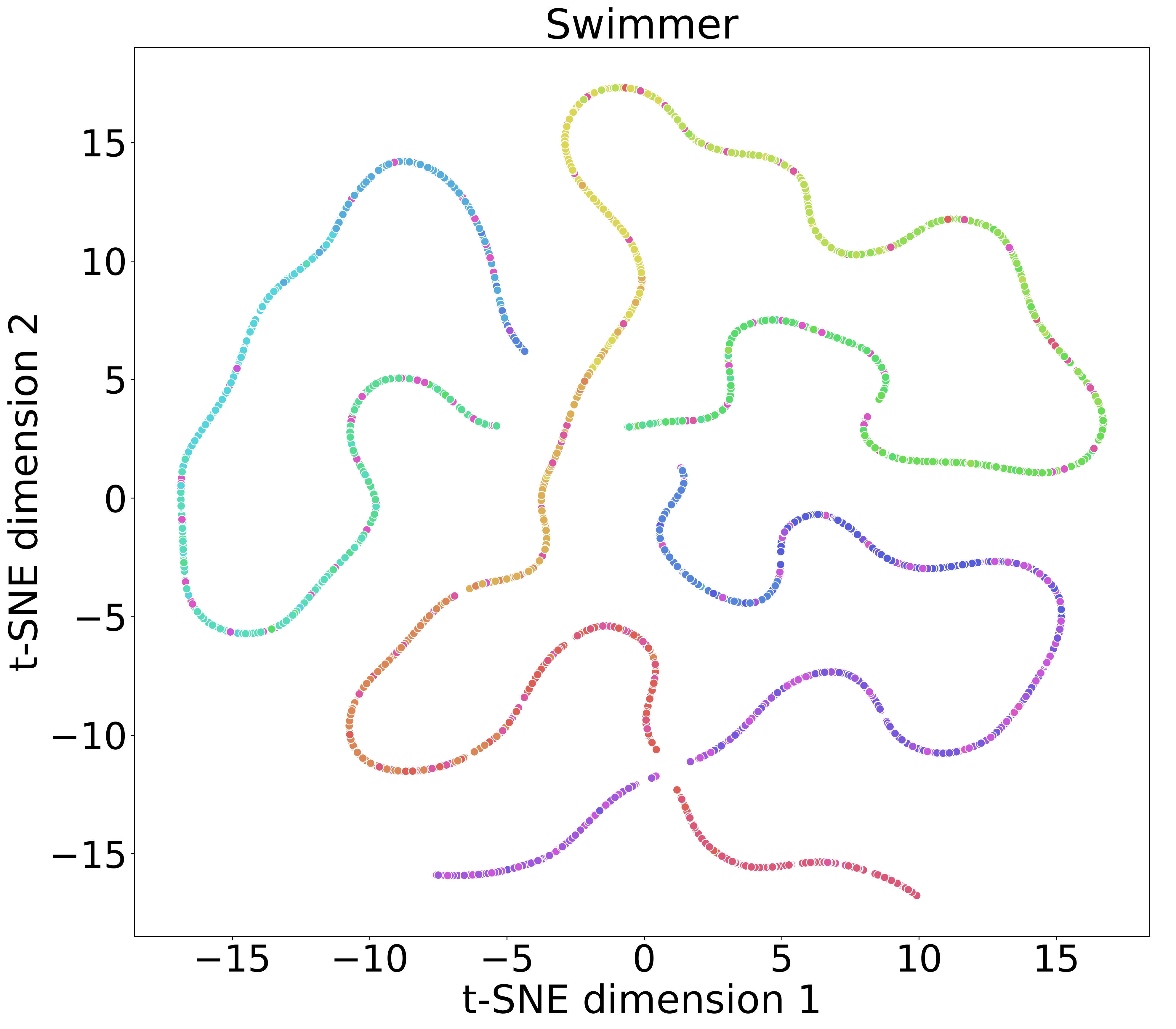}}
        \subfigure[]{\label{fig:env_ant}\includegraphics[width=0.36\columnwidth]{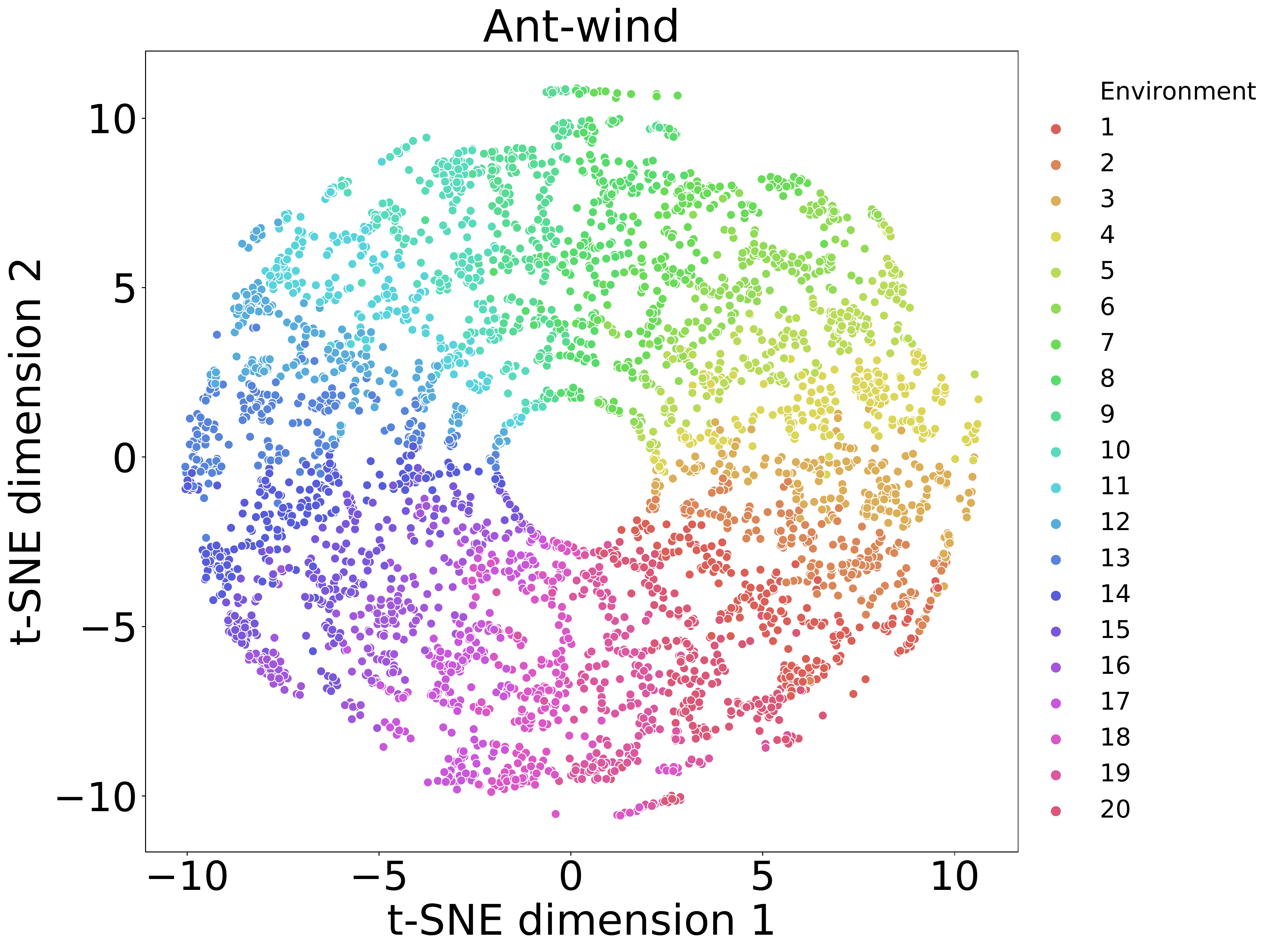}}
    \caption{t-SNE plots of the learned \textbf{environment embeddings} $z_{d}$ for Spaceship (a), Swimmer (b), and Ant-wind (c). The color corresponds to the \textit{environment} that generated the transitions used to encode the corresponding dynamics embeddings. The plot contains embeddings of both train and test environments.}
    \label{fig:env_embeddings}
\end{figure}

\begin{figure}[ht!]
        \subfigure[]{\label{fig:pi_space}\includegraphics[width=0.31\columnwidth]{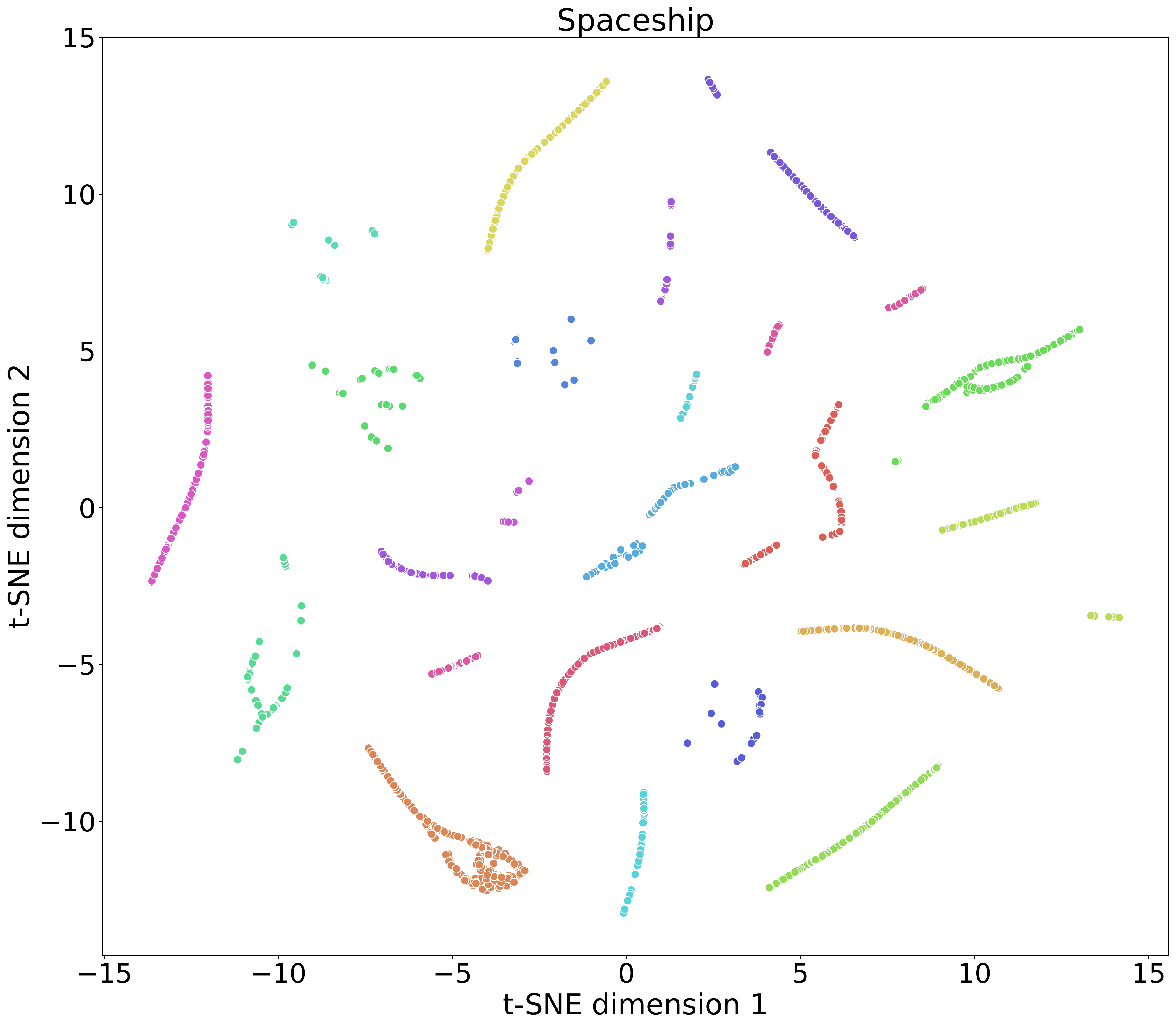}}
        \subfigure[]{\label{fig:pi_swim}\includegraphics[width=0.31\columnwidth]{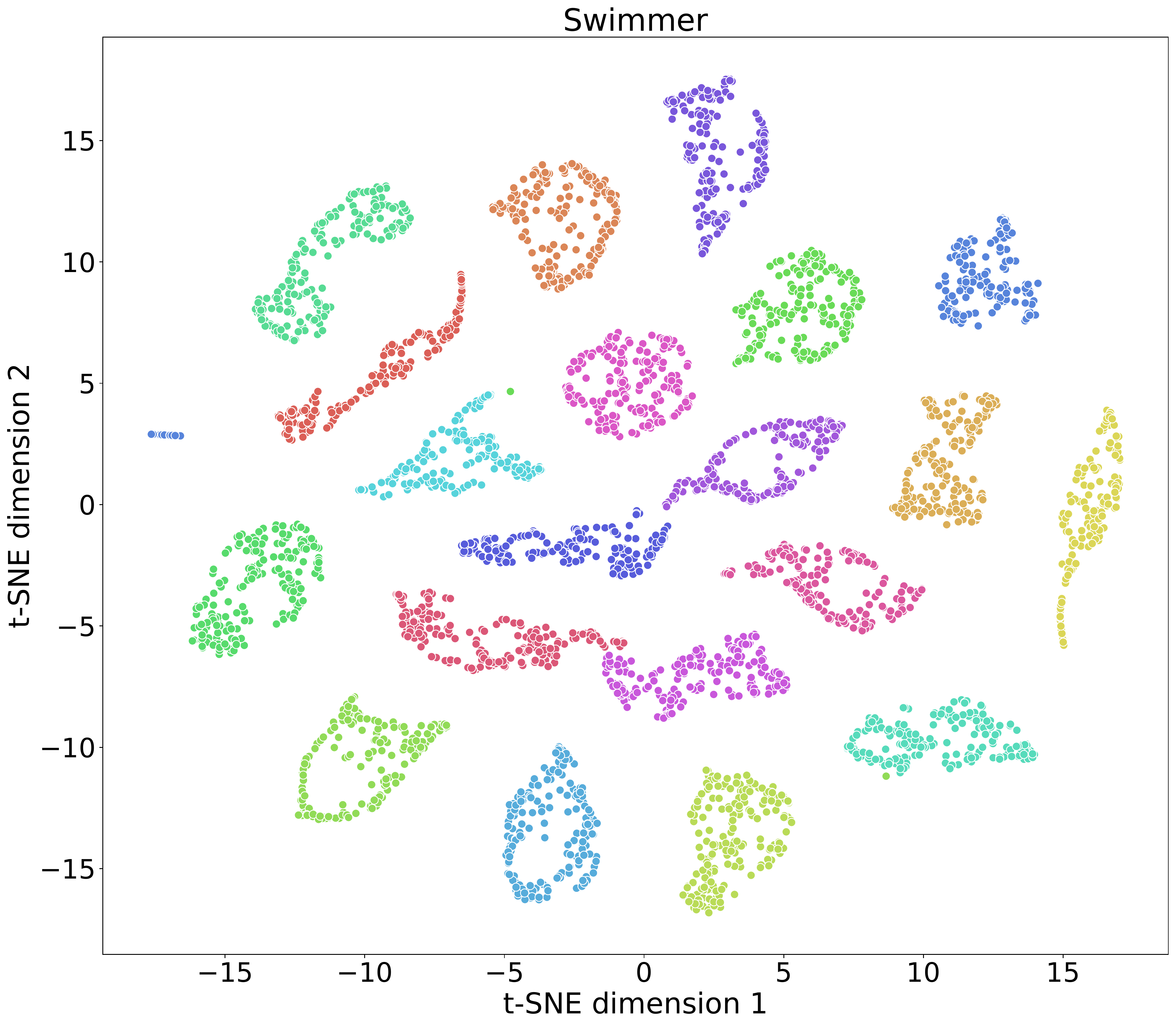}}
        \subfigure[]{\label{fig:pi_ant}\includegraphics[width=0.34\columnwidth]{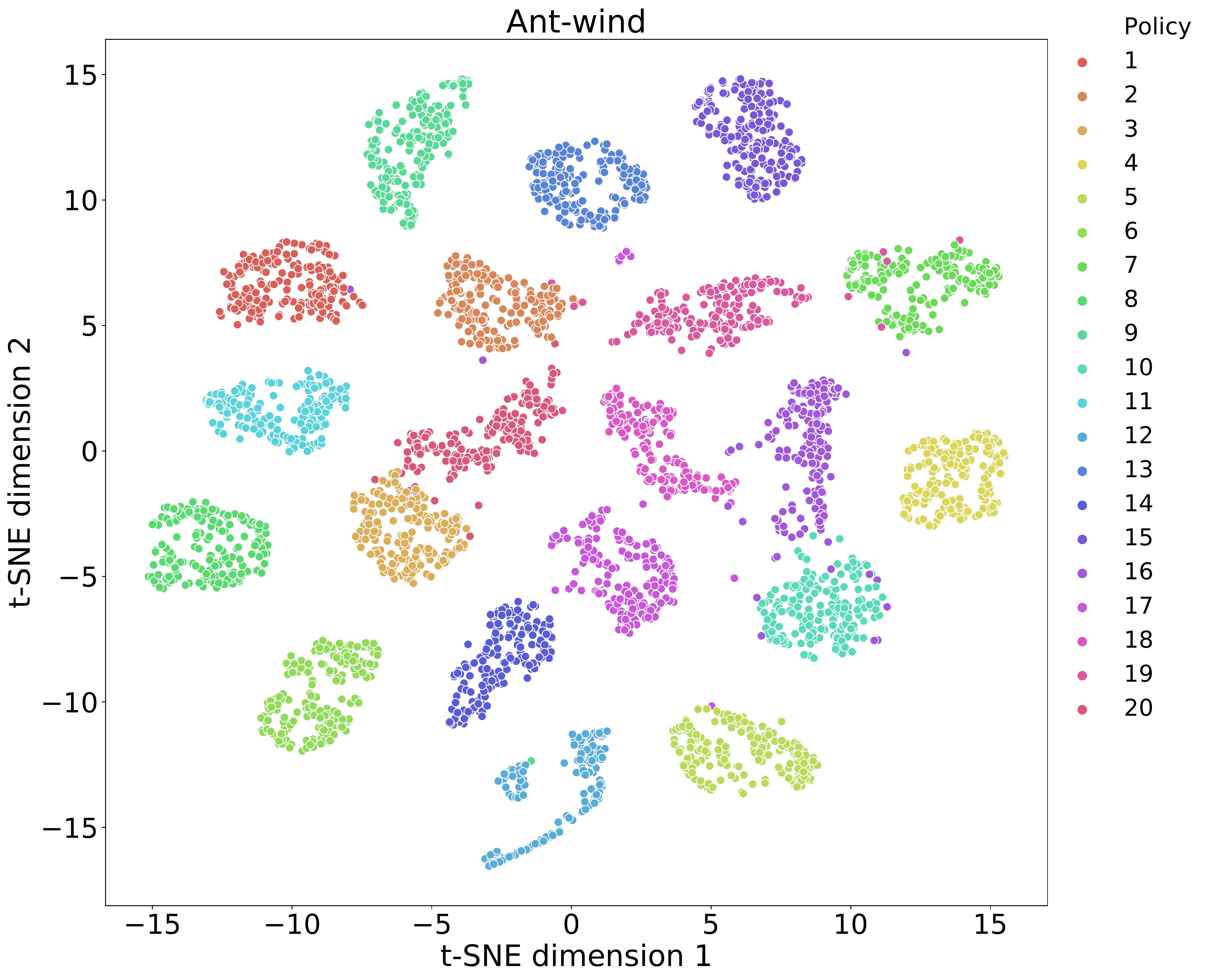}}
    \caption{t-SNE plots of the learned \textbf{policy embeddings} $z_{\pi}$ for Spaceship (a), Swimmer (b), and Ant-wind (c). The color corresponds to the \textit{policy} that generated the transitions used to encode the corresponding policy embeddings. The plot contains embeddings of policies trained on both train and test environments.}
    \label{fig:policy_embeddings}
\end{figure}

Figure~\ref{fig:env_embeddings} shows a t-SNE plot \citep{Maaten2008VisualizingDU} of the learned dynamics embeddings on the three continuous control domains used for evaluating our method. Environment $i$ corresponds to dynamics defined by $d = i \times \pi / 10$ (\ie{} the direction of the wind in Swimmer's environment 1 is at $\pi/10$ degrees). Environments 1 - 15  are used for training, while 16 - 20 are used for evaluation. The latent space captures the continuous nature of the distribution used to generate the environment dynamics. For example, in Figure~\ref{fig:env_ant}, one can see the wind direction corresponding to a particular environment, indicating that the learned embedding space uncovers the manifold structure of the true dynamics distribution. Even if, during training, the dynamics model never sees trajectories through the test environments, it is still able to embed them within the 1D manifold, thus preserving smoothness in the latent space. 

Similarly, Figure~\ref{fig:policy_embeddings} shows the corresponding t-SNE \citep{Maaten2008VisualizingDU} of the learned policy embeddings for Spaceship, Swimmer, and Ant. The embeddings are clustered according to the policy that generated them.

\label{results}
\section{Discussion and Future Work}
In this work, we propose policy-dynamics value functions (\pdvf{}), a novel framework for fast adaptation to new environment dynamics. The key idea is to learn a value function conditioned on both a policy and a dynamics embedding which are learned in a self-supervised way. At test time, the environment embedding can be inferred from only a few interactions, which allows the selection of a policy that maximizes the learned value function. \pdvf{} has a number of desirable properties: it leverages the structure in both the policy and the dynamics space to estimate the expected return, it only needs a small number of steps to adapt to unseen dynamics, it does not update any parameters at test time, and it does not require dense reward or long rollouts to find an effective policy in a new environment. Empirical results on a set of continuous control domains show that \pdvf{} outperforms other methods on unseen dynamics, while being competitive on training environments. 

\pdvf{} opens up many promising directions for future research. First of all, the formulation can be extended to estimate the value function not only for a family of policies and environment dynamics, but also for a family of reward functions. Another avenue for future research is to use a more general class of function approximators (such as neural networks) to parameterise the value estimator instead of a quadratic form. The \pdvf{} framework can, in principle, also be used to evaluate a family of policies and environments on other metrics of interest besides the expected return, such as, for example, reward variance, agent prosociality, deviation from expert behavior, and so on. Another interesting direction is to integrate additional constraints (or prior knowledge) to the optimization problem (\eg{} maximize expected return while only using policies in a certain region of the policy space). As noted by \citet{precup2001off}, \citet{Sutton2011HordeAS}, and \citet{white2012scaling}, learning about multiple policies in parallel via general value functions can be useful for lifelong learning. Similarly, \pdvf{} can be a useful tool for an agent to continually gather knowledge about various policies and dynamics in the world. Finally, \pdvf{} can also be applied to multi-agent settings for adapting to different opponents or teammates whose behaviors determine the environment dynamics.

\label{conclusion}

\section*{Acknowledgements}
Roberta and Max were supported by the DARPA L2M grant. 

\bibliography{main}

\begin{thebibliography}{72}
\providecommand{\natexlab}[1]{#1}
\providecommand{\url}[1]{\texttt{#1}}
\expandafter\ifx\csname urlstyle\endcsname\relax
  \providecommand{\doi}[1]{doi: #1}\else
  \providecommand{\doi}{doi: \begingroup \urlstyle{rm}\Url}\fi

\bibitem[Ammar et~al.(2012)Ammar, Tuyls, Taylor, Driessens, and
  Weiss]{ammar2012reinforcement}
Ammar, H.~B., Tuyls, K., Taylor, M.~E., Driessens, K., and Weiss, G.
\newblock Reinforcement learning transfer via sparse coding.
\newblock In \emph{Proceedings of the 11th international conference on
  autonomous agents and multiagent systems}, volume~1, pp.\  383--390.
  International Foundation for Autonomous Agents and Multiagent Systems~…,
  2012.

\bibitem[Ammar et~al.(2014)Ammar, Eaton, Taylor, Mocanu, Driessens, Weiss, and
  Tuyls]{ammar2014automated}
Ammar, H.~B., Eaton, E., Taylor, M.~E., Mocanu, D.~C., Driessens, K., Weiss,
  G., and Tuyls, K.
\newblock An automated measure of mdp similarity for transfer in reinforcement
  learning.
\newblock In \emph{Workshops at the Twenty-Eighth AAAI Conference on Artificial
  Intelligence}, 2014.

\bibitem[Andreas et~al.(2017)Andreas, Klein, and Levine]{andreas2017modular}
Andreas, J., Klein, D., and Levine, S.
\newblock Modular multitask reinforcement learning with policy sketches.
\newblock In \emph{International Conference on Machine Learning}, pp.\
  166--175, 2017.

\bibitem[Arnekvist et~al.(2018)Arnekvist, Kragic, and
  Stork]{Arnekvist2018VPEVP}
Arnekvist, I., Kragic, D., and Stork, J.~A.
\newblock Vpe: Variational policy embedding for transfer reinforcement
  learning.
\newblock \emph{2019 International Conference on Robotics and Automation
  (ICRA)}, pp.\  36--42, 2018.

\bibitem[Barreto et~al.(2017)Barreto, Dabney, Munos, Hunt, Schaul, van Hasselt,
  and Silver]{barreto2017successor}
Barreto, A., Dabney, W., Munos, R., Hunt, J.~J., Schaul, T., van Hasselt,
  H.~P., and Silver, D.
\newblock Successor features for transfer in reinforcement learning.
\newblock In \emph{Advances in neural information processing systems}, pp.\
  4055--4065, 2017.

\bibitem[Berner et~al.(2019)Berner, Brockman, Chan, Cheung, Debiak, Dennison,
  Farhi, Fischer, Hashme, Hesse, J{\'o}zefowicz, Gray, Olsson, Pachocki,
  Petrov, de~Oliveira~Pinto, Raiman, Salimans, Schlatter, Schneider, Sidor,
  Sutskever, Tang, Wolski, and Zhang]{Berner2019Dota2W}
Berner, C., Brockman, G., Chan, B., Cheung, V., Debiak, P., Dennison, C.,
  Farhi, D., Fischer, Q., Hashme, S., Hesse, C., J{\'o}zefowicz, R., Gray, S.,
  Olsson, C., Pachocki, J.~W., Petrov, M., de~Oliveira~Pinto, H.~P., Raiman,
  J., Salimans, T., Schlatter, J., Schneider, J., Sidor, S., Sutskever, I.,
  Tang, J., Wolski, F., and Zhang, S.
\newblock Dota 2 with large scale deep reinforcement learning.
\newblock \emph{ArXiv}, abs/1912.06680, 2019.

\bibitem[Borsa et~al.(2016)Borsa, Graepel, and Shawe-Taylor]{borsa2016learning}
Borsa, D., Graepel, T., and Shawe-Taylor, J.
\newblock Learning shared representations in multi-task reinforcement learning.
\newblock \emph{arXiv preprint arXiv:1603.02041}, 2016.

\bibitem[Borsa et~al.(2018)Borsa, Barreto, Quan, Mankowitz, Munos, van Hasselt,
  Silver, and Schaul]{borsa2018universal}
Borsa, D., Barreto, A., Quan, J., Mankowitz, D., Munos, R., van Hasselt, H.,
  Silver, D., and Schaul, T.
\newblock Universal successor features approximators.
\newblock \emph{arXiv preprint arXiv:1812.07626}, 2018.

\bibitem[Co-Reyes et~al.(2018)Co-Reyes, Liu, Gupta, Eysenbach, Abbeel, and
  Levine]{CoReyes2018SelfConsistentTA}
Co-Reyes, J.~D., Liu, Y., Gupta, A., Eysenbach, B., Abbeel, P., and Levine, S.
\newblock Self-consistent trajectory autoencoder: Hierarchical reinforcement
  learning with trajectory embeddings.
\newblock In \emph{ICML}, 2018.

\bibitem[Cobbe et~al.(2019)Cobbe, Klimov, Hesse, Kim, and
  Schulman]{Cobbe2019QuantifyingGI}
Cobbe, K., Klimov, O., Hesse, C., Kim, T., and Schulman, J.
\newblock Quantifying generalization in reinforcement learning.
\newblock In \emph{ICML}, 2019.

\bibitem[Cully et~al.(2015)Cully, Clune, Tarapore, and
  Mouret]{Cully2015RobotsTC}
Cully, A., Clune, J., Tarapore, D., and Mouret, J.-B.
\newblock Robots that can adapt like animals.
\newblock \emph{Nature}, 521:\penalty0 503--507, 2015.

\bibitem[Da~Silva et~al.(2012)Da~Silva, Konidaris, and Barto]{da2012learning}
Da~Silva, B., Konidaris, G., and Barto, A.
\newblock Learning parameterized skills.
\newblock \emph{arXiv preprint arXiv:1206.6398}, 2012.

\bibitem[Devin et~al.(2016)Devin, Gupta, Darrell, Abbeel, and
  Levine]{Devin2016LearningMN}
Devin, C., Gupta, A., Darrell, T., Abbeel, P., and Levine, S.
\newblock Learning modular neural network policies for multi-task and
  multi-robot transfer.
\newblock \emph{2017 IEEE International Conference on Robotics and Automation
  (ICRA)}, pp.\  2169--2176, 2016.

\bibitem[Doshi-Velez \& Konidaris(2013)Doshi-Velez and
  Konidaris]{DoshiVelez2013HiddenPM}
Doshi-Velez, F. and Konidaris, G.
\newblock Hidden parameter markov decision processes: A semiparametric
  regression approach for discovering latent task parametrizations.
\newblock \emph{IJCAI : proceedings of the conference}, 2016:\penalty0
  1432--1440, 2013.

\bibitem[Duan et~al.(2016)Duan, Schulman, Chen, Bartlett, Sutskever, and
  Abbeel]{Duan2016RL2FR}
Duan, Y., Schulman, J., Chen, X., Bartlett, P.~L., Sutskever, I., and Abbeel,
  P.
\newblock Rl$^2$: Fast reinforcement learning via slow reinforcement learning.
\newblock \emph{ArXiv}, abs/1611.02779, 2016.

\bibitem[Duan et~al.(2017)Duan, Andrychowicz, Stadie, Ho, Schneider, Sutskever,
  Abbeel, and Zaremba]{Duan2017OneShotIL}
Duan, Y., Andrychowicz, M., Stadie, B.~C., Ho, J., Schneider, J., Sutskever,
  I., Abbeel, P., and Zaremba, W.
\newblock One-shot imitation learning.
\newblock In \emph{NIPS}, 2017.

\bibitem[Finn et~al.(2017)Finn, Abbeel, and Levine]{finn2017model}
Finn, C., Abbeel, P., and Levine, S.
\newblock Model-agnostic meta-learning for fast adaptation of deep networks.
\newblock In \emph{Proceedings of the 34th International Conference on Machine
  Learning-Volume 70}, pp.\  1126--1135. JMLR. org, 2017.

\bibitem[Gupta et~al.(2017)Gupta, Devin, Liu, Abbeel, and
  Levine]{gupta2017learning}
Gupta, A., Devin, C., Liu, Y., Abbeel, P., and Levine, S.
\newblock Learning invariant feature spaces to transfer skills with
  reinforcement learning.
\newblock \emph{arXiv preprint arXiv:1703.02949}, 2017.

\bibitem[Hansen et~al.(2019)Hansen, Dabney, Barreto, Van~de Wiele,
  Warde-Farley, and Mnih]{hansen2019fast}
Hansen, S., Dabney, W., Barreto, A., Van~de Wiele, T., Warde-Farley, D., and
  Mnih, V.
\newblock Fast task inference with variational intrinsic successor features.
\newblock \emph{arXiv preprint arXiv:1906.05030}, 2019.

\bibitem[Hausman et~al.(2018)Hausman, Springenberg, Wang, Heess, and
  Riedmiller]{Hausman2018LearningAE}
Hausman, K., Springenberg, J.~T., Wang, Z., Heess, N. M.~O., and Riedmiller,
  M.~A.
\newblock Learning an embedding space for transferable robot skills.
\newblock In \emph{ICLR}, 2018.

\bibitem[He et~al.(2018)He, Julian, Heiden, Zhang, Schaal, Lim, Sukhatme, and
  Hausman]{he2018zero}
He, Z., Julian, R., Heiden, E., Zhang, H., Schaal, S., Lim, J.~J., Sukhatme,
  G., and Hausman, K.
\newblock Zero-shot skill composition and simulation-to-real transfer by
  learning task representations.
\newblock \emph{arXiv preprint arXiv:1810.02422}, 2018.

\bibitem[Henderson et~al.(2018)Henderson, Islam, Bachman, Pineau, Precup, and
  Meger]{henderson2018deep}
Henderson, P., Islam, R., Bachman, P., Pineau, J., Precup, D., and Meger, D.
\newblock Deep reinforcement learning that matters.
\newblock In \emph{Thirty-Second AAAI Conference on Artificial Intelligence},
  2018.

\bibitem[Hessel et~al.(2019)Hessel, Soyer, Espeholt, Czarnecki, Schmitt, and
  van Hasselt]{hessel2019multi}
Hessel, M., Soyer, H., Espeholt, L., Czarnecki, W., Schmitt, S., and van
  Hasselt, H.
\newblock Multi-task deep reinforcement learning with popart.
\newblock In \emph{Proceedings of the AAAI Conference on Artificial
  Intelligence}, volume~33, pp.\  3796--3803, 2019.

\bibitem[Higgins et~al.(2017)Higgins, Pal, Rusu, Matthey, Burgess, Pritzel,
  Botvinick, Blundell, and Lerchner]{Higgins2017DARLAIZ}
Higgins, I., Pal, A., Rusu, A.~A., Matthey, L., Burgess, C., Pritzel, A.,
  Botvinick, M.~M., Blundell, C., and Lerchner, A.
\newblock Darla: Improving zero-shot transfer in reinforcement learning.
\newblock In \emph{ICML}, 2017.

\bibitem[Houthooft et~al.(2018)Houthooft, Chen, Isola, Stadie, Wolski, Ho, and
  Abbeel]{Houthooft2018EvolvedPG}
Houthooft, R., Chen, R.~Y., Isola, P., Stadie, B.~C., Wolski, F., Ho, J., and
  Abbeel, P.
\newblock Evolved policy gradients.
\newblock \emph{ArXiv}, abs/1802.04821, 2018.

\bibitem[Humplik et~al.(2019)Humplik, Galashov, Hasenclever, Ortega, Teh, and
  Heess]{humplik2019meta}
Humplik, J., Galashov, A., Hasenclever, L., Ortega, P.~A., Teh, Y.~W., and
  Heess, N.
\newblock Meta reinforcement learning as task inference.
\newblock \emph{arXiv preprint arXiv:1905.06424}, 2019.

\bibitem[Jaderberg et~al.(2019)Jaderberg, Czarnecki, Dunning, Marris, Lever,
  Castaneda, Beattie, Rabinowitz, Morcos, Ruderman, et~al.]{jaderberg2019human}
Jaderberg, M., Czarnecki, W.~M., Dunning, I., Marris, L., Lever, G., Castaneda,
  A.~G., Beattie, C., Rabinowitz, N.~C., Morcos, A.~S., Ruderman, A., et~al.
\newblock Human-level performance in 3d multiplayer games with population-based
  reinforcement learning.
\newblock \emph{Science}, 364\penalty0 (6443):\penalty0 859--865, 2019.

\bibitem[Killian et~al.(2017)Killian, Konidaris, and
  Doshi-Velez]{Killian2017RobustAE}
Killian, T.~W., Konidaris, G., and Doshi-Velez, F.
\newblock Robust and efficient transfer learning with hidden parameter markov
  decision processes.
\newblock \emph{Advances in neural information processing systems},
  30:\penalty0 6250--6261, 2017.

\bibitem[Kingma \& Ba(2014)Kingma and Ba]{kingma2014adam}
Kingma, D.~P. and Ba, J.
\newblock Adam: A method for stochastic optimization.
\newblock \emph{arXiv preprint arXiv:1412.6980}, 2014.

\bibitem[Kingma \& Welling(2013)Kingma and Welling]{Kingma2013AutoEncodingVB}
Kingma, D.~P. and Welling, M.
\newblock Auto-encoding variational bayes.
\newblock \emph{CoRR}, abs/1312.6114, 2013.

\bibitem[Lazaric(2012)]{lazaric2012transfer}
Lazaric, A.
\newblock Transfer in reinforcement learning: a framework and a survey.
\newblock In \emph{Reinforcement Learning}, pp.\  143--173. Springer, 2012.

\bibitem[Madjiheurem \& Toni(2019)Madjiheurem and
  Toni]{madjiheurem2019state2vec}
Madjiheurem, S. and Toni, L.
\newblock State2vec: Off-policy successor features approximators.
\newblock \emph{arXiv preprint arXiv:1910.10277}, 2019.

\bibitem[Mnih et~al.(2015)Mnih, Kavukcuoglu, Silver, Rusu, Veness, Bellemare,
  Graves, Riedmiller, Fidjeland, Ostrovski, et~al.]{mnih2015human}
Mnih, V., Kavukcuoglu, K., Silver, D., Rusu, A.~A., Veness, J., Bellemare,
  M.~G., Graves, A., Riedmiller, M., Fidjeland, A.~K., Ostrovski, G., et~al.
\newblock Human-level control through deep reinforcement learning.
\newblock \emph{Nature}, 518\penalty0 (7540):\penalty0 529--533, 2015.

\bibitem[Nagabandi et~al.(2018)Nagabandi, Clavera, Liu, Fearing, Abbeel,
  Levine, and Finn]{nagabandi2018learning}
Nagabandi, A., Clavera, I., Liu, S., Fearing, R.~S., Abbeel, P., Levine, S.,
  and Finn, C.
\newblock Learning to adapt in dynamic, real-world environments through
  meta-reinforcement learning.
\newblock \emph{arXiv preprint arXiv:1803.11347}, 2018.

\bibitem[Oh et~al.(2017)Oh, Singh, Lee, and Kohli]{oh2017zero}
Oh, J., Singh, S., Lee, H., and Kohli, P.
\newblock Zero-shot task generalization with multi-task deep reinforcement
  learning.
\newblock \emph{arXiv preprint arXiv:1706.05064}, 2017.

\bibitem[Parisotto et~al.(2015)Parisotto, Ba, and
  Salakhutdinov]{parisotto2015actor}
Parisotto, E., Ba, J.~L., and Salakhutdinov, R.
\newblock Actor-mimic: Deep multitask and transfer reinforcement learning.
\newblock \emph{arXiv preprint arXiv:1511.06342}, 2015.

\bibitem[Paul et~al.(2018)Paul, Osborne, and Whiteson]{Paul2018FingerprintPO}
Paul, S., Osborne, M.~A., and Whiteson, S.
\newblock Fingerprint policy optimisation for robust reinforcement learning.
\newblock In \emph{ICML}, 2018.

\bibitem[Perez et~al.(2018)Perez, Such, and Karaletsos]{Perez2018EfficientTL}
Perez, C.~F., Such, F.~P., and Karaletsos, T.
\newblock Efficient transfer learning and online adaptation with latent
  variable models for continuous control.
\newblock \emph{ArXiv}, abs/1812.03399, 2018.

\bibitem[Petangoda et~al.(2019)Petangoda, Pascual-Diaz, Adam, Vrancx, and
  Grau-Moya]{Petangoda2019DisentangledSE}
Petangoda, J.~C., Pascual-Diaz, S., Adam, V., Vrancx, P., and Grau-Moya, J.
\newblock Disentangled skill embeddings for reinforcement learning.
\newblock \emph{ArXiv}, abs/1906.09223, 2019.

\bibitem[Pinto et~al.(2017)Pinto, Davidson, Sukthankar, and
  Gupta]{Pinto2017RobustAR}
Pinto, L., Davidson, J., Sukthankar, R., and Gupta, A.
\newblock Robust adversarial reinforcement learning.
\newblock In \emph{ICML}, 2017.

\bibitem[Precup et~al.(2001)Precup, Sutton, and Dasgupta]{precup2001off}
Precup, D., Sutton, R.~S., and Dasgupta, S.
\newblock Off-policy temporal-difference learning with function approximation.
\newblock In \emph{ICML}, pp.\  417--424, 2001.

\bibitem[Raileanu \& Rockt{\"a}schel(2020)Raileanu and
  Rockt{\"a}schel]{Raileanu2020RIDERI}
Raileanu, R. and Rockt{\"a}schel, T.
\newblock Ride: Rewarding impact-driven exploration for procedurally-generated
  environments.
\newblock \emph{ArXiv}, abs/2002.12292, 2020.

\bibitem[Rajeswaran et~al.(2017)Rajeswaran, Lowrey, Todorov, and
  Kakade]{rajeswaran2017towards}
Rajeswaran, A., Lowrey, K., Todorov, E.~V., and Kakade, S.~M.
\newblock Towards generalization and simplicity in continuous control.
\newblock In \emph{Advances in Neural Information Processing Systems}, pp.\
  6550--6561, 2017.

\bibitem[Rakelly et~al.(2019)Rakelly, Zhou, Finn, Levine, and
  Quillen]{rakelly2019efficient}
Rakelly, K., Zhou, A., Finn, C., Levine, S., and Quillen, D.
\newblock Efficient off-policy meta-reinforcement learning via probabilistic
  context variables.
\newblock In \emph{International conference on machine learning}, pp.\
  5331--5340, 2019.

\bibitem[S{\ae}mundsson et~al.(2018)S{\ae}mundsson, Hofmann, and
  Deisenroth]{saemundsson2018meta}
S{\ae}mundsson, S., Hofmann, K., and Deisenroth, M.~P.
\newblock Meta reinforcement learning with latent variable gaussian processes.
\newblock \emph{arXiv preprint arXiv:1803.07551}, 2018.

\bibitem[Sahni et~al.(2017)Sahni, Kumar, Tejani, and Isbell]{sahni2017learning}
Sahni, H., Kumar, S., Tejani, F., and Isbell, C.
\newblock Learning to compose skills.
\newblock \emph{arXiv preprint arXiv:1711.11289}, 2017.

\bibitem[Schaul et~al.(2015)Schaul, Horgan, Gregor, and
  Silver]{schaul2015universal}
Schaul, T., Horgan, D., Gregor, K., and Silver, D.
\newblock Universal value function approximators.
\newblock In \emph{International conference on machine learning}, pp.\
  1312--1320, 2015.

\bibitem[Schulman et~al.(2017)Schulman, Wolski, Dhariwal, Radford, and
  Klimov]{schulman2017proximal}
Schulman, J., Wolski, F., Dhariwal, P., Radford, A., and Klimov, O.
\newblock Proximal policy optimization algorithms.
\newblock \emph{arXiv preprint arXiv:1707.06347}, 2017.

\bibitem[Silver et~al.(2016)Silver, Huang, Maddison, Guez, Sifre, Van
  Den~Driessche, Schrittwieser, Antonoglou, Panneershelvam, Lanctot,
  et~al.]{silver2016mastering}
Silver, D., Huang, A., Maddison, C.~J., Guez, A., Sifre, L., Van Den~Driessche,
  G., Schrittwieser, J., Antonoglou, I., Panneershelvam, V., Lanctot, M.,
  et~al.
\newblock Mastering the game of go with deep neural networks and tree search.
\newblock \emph{nature}, 529\penalty0 (7587):\penalty0 484, 2016.

\bibitem[Silver et~al.(2017)Silver, Schrittwieser, Simonyan, Antonoglou, Huang,
  Guez, Hubert, Baker, Lai, Bolton, et~al.]{silver2017mastering}
Silver, D., Schrittwieser, J., Simonyan, K., Antonoglou, I., Huang, A., Guez,
  A., Hubert, T., Baker, L., Lai, M., Bolton, A., et~al.
\newblock Mastering the game of go without human knowledge.
\newblock \emph{nature}, 550\penalty0 (7676):\penalty0 354--359, 2017.

\bibitem[Silver et~al.(2018)Silver, Hubert, Schrittwieser, Antonoglou, Lai,
  Guez, Lanctot, Sifre, Kumaran, Graepel, et~al.]{silver2018general}
Silver, D., Hubert, T., Schrittwieser, J., Antonoglou, I., Lai, M., Guez, A.,
  Lanctot, M., Sifre, L., Kumaran, D., Graepel, T., et~al.
\newblock A general reinforcement learning algorithm that masters chess, shogi,
  and go through self-play.
\newblock \emph{Science}, 362\penalty0 (6419):\penalty0 1140--1144, 2018.

\bibitem[Siriwardhana et~al.(2019)Siriwardhana, Weerasakera, Matthies, and
  Nanayakkara]{siriwardhana2019vusfa}
Siriwardhana, S., Weerasakera, R., Matthies, D.~J., and Nanayakkara, S.
\newblock Vusfa: Variational universal successor features approximator to
  improve transfer drl for target driven visual navigation.
\newblock \emph{arXiv preprint arXiv:1908.06376}, 2019.

\bibitem[Song et~al.(2020)Song, Jiang, Tu, Du, and
  Neyshabur]{Song2020ObservationalOI}
Song, X., Jiang, Y., Tu, S., Du, Y., and Neyshabur, B.
\newblock Observational overfitting in reinforcement learning.
\newblock \emph{ArXiv}, abs/1912.02975, 2020.

\bibitem[Sutton et~al.(2011)Sutton, Modayil, Delp, Degris, Pilarski, White, and
  Precup]{Sutton2011HordeAS}
Sutton, R.~S., Modayil, J., Delp, M., Degris, T., Pilarski, P.~M., White, A.,
  and Precup, D.
\newblock Horde: a scalable real-time architecture for learning knowledge from
  unsupervised sensorimotor interaction.
\newblock In \emph{AAMAS}, 2011.

\bibitem[Taylor \& Stone(2009)Taylor and Stone]{taylor2009transfer}
Taylor, M.~E. and Stone, P.
\newblock Transfer learning for reinforcement learning domains: A survey.
\newblock \emph{Journal of Machine Learning Research}, 10\penalty0
  (Jul):\penalty0 1633--1685, 2009.

\bibitem[Teh et~al.(2017)Teh, Bapst, Czarnecki, Quan, Kirkpatrick, Hadsell,
  Heess, and Pascanu]{teh2017distral}
Teh, Y., Bapst, V., Czarnecki, W.~M., Quan, J., Kirkpatrick, J., Hadsell, R.,
  Heess, N., and Pascanu, R.
\newblock Distral: Robust multitask reinforcement learning.
\newblock In \emph{Advances in Neural Information Processing Systems}, pp.\
  4496--4506, 2017.

\bibitem[Todorov et~al.(2012)Todorov, Erez, and Tassa]{Todorov2012MuJoCoAP}
Todorov, E., Erez, T., and Tassa, Y.
\newblock Mujoco: A physics engine for model-based control.
\newblock \emph{2012 IEEE/RSJ International Conference on Intelligent Robots
  and Systems}, pp.\  5026--5033, 2012.

\bibitem[van~der Maaten \& Hinton(2008)van~der Maaten and
  Hinton]{Maaten2008VisualizingDU}
van~der Maaten, L. and Hinton, G.~E.
\newblock Visualizing data using t-sne.
\newblock 2008.

\bibitem[Vaswani et~al.(2017)Vaswani, Shazeer, Parmar, Uszkoreit, Jones, Gomez,
  Kaiser, and Polosukhin]{vaswani2017attention}
Vaswani, A., Shazeer, N., Parmar, N., Uszkoreit, J., Jones, L., Gomez, A.~N.,
  Kaiser, {\L}., and Polosukhin, I.
\newblock Attention is all you need.
\newblock In \emph{Advances in neural information processing systems}, pp.\
  5998--6008, 2017.

\bibitem[Vinyals et~al.(2019)Vinyals, Babuschkin, Czarnecki, Mathieu, Dudzik,
  Chung, Choi, Powell, Ewalds, Georgiev, et~al.]{vinyals2019grandmaster}
Vinyals, O., Babuschkin, I., Czarnecki, W.~M., Mathieu, M., Dudzik, A., Chung,
  J., Choi, D.~H., Powell, R., Ewalds, T., Georgiev, P., et~al.
\newblock Grandmaster level in starcraft ii using multi-agent reinforcement
  learning.
\newblock \emph{Nature}, 575\penalty0 (7782):\penalty0 350--354, 2019.

\bibitem[Wang et~al.(2016)Wang, Kurth-Nelson, Soyer, Leibo, Tirumala, Munos,
  Blundell, Kumaran, and Botvinick]{Wang2016LearningTR}
Wang, J.~X., Kurth-Nelson, Z., Soyer, H., Leibo, J.~Z., Tirumala, D., Munos,
  R., Blundell, C., Kumaran, D., and Botvinick, M.~M.
\newblock Learning to reinforcement learn.
\newblock \emph{ArXiv}, abs/1611.05763, 2016.

\bibitem[Wang et~al.(2017)Wang, Merel, Reed, de~Freitas, Wayne, and
  Heess]{Wang2017RobustIO}
Wang, Z., Merel, J., Reed, S.~E., de~Freitas, N., Wayne, G., and Heess, N.
  M.~O.
\newblock Robust imitation of diverse behaviors.
\newblock In \emph{NIPS}, 2017.

\bibitem[White et~al.(2012)White, Modayil, and Sutton]{white2012scaling}
White, A., Modayil, J., and Sutton, R.~S.
\newblock Scaling life-long off-policy learning.
\newblock In \emph{2012 IEEE International Conference on Development and
  Learning and Epigenetic Robotics (ICDL)}, pp.\  1--6. IEEE, 2012.

\bibitem[Whiteson et~al.(2011)Whiteson, Tanner, Taylor, and
  Stone]{Whiteson2011ProtectingAE}
Whiteson, S., Tanner, B., Taylor, M.~E., and Stone, P.
\newblock Protecting against evaluation overfitting in empirical reinforcement
  learning.
\newblock \emph{2011 IEEE Symposium on Adaptive Dynamic Programming and
  Reinforcement Learning (ADPRL)}, pp.\  120--127, 2011.

\bibitem[Xu et~al.(2018)Xu, van Hasselt, and Silver]{Xu2018MetaGradientRL}
Xu, Z., van Hasselt, H., and Silver, D.
\newblock Meta-gradient reinforcement learning.
\newblock In \emph{NeurIPS}, 2018.

\bibitem[Yang et~al.(2019)Yang, Petersen, Zha, and Faissol]{yang2019single}
Yang, J., Petersen, B., Zha, H., and Faissol, D.
\newblock Single episode policy transfer in reinforcement learning.
\newblock \emph{arXiv preprint arXiv:1910.07719}, 2019.

\bibitem[Yao et~al.(2018)Yao, Killian, Konidaris, and
  Doshi-Velez]{Yao2018DirectPT}
Yao, J., Killian, T.~W., Konidaris, G., and Doshi-Velez, F.
\newblock Direct policy transfer via hidden parameter markov decision
  processes.
\newblock 2018.

\bibitem[Zhang et~al.(2018{\natexlab{a}})Zhang, Ballas, and
  Pineau]{zhang2018dissection}
Zhang, A., Ballas, N., and Pineau, J.
\newblock A dissection of overfitting and generalization in continuous
  reinforcement learning.
\newblock \emph{arXiv preprint arXiv:1806.07937}, 2018{\natexlab{a}}.

\bibitem[Zhang et~al.(2018{\natexlab{b}})Zhang, Satija, and
  Pineau]{Zhang2018DecouplingDA}
Zhang, A., Satija, H., and Pineau, J.
\newblock Decoupling dynamics and reward for transfer learning.
\newblock \emph{ArXiv}, abs/1804.10689, 2018{\natexlab{b}}.

\bibitem[Zhang et~al.(2018{\natexlab{c}})Zhang, Vinyals, Munos, and
  Bengio]{zhang2018study}
Zhang, C., Vinyals, O., Munos, R., and Bengio, S.
\newblock A study on overfitting in deep reinforcement learning.
\newblock \emph{arXiv preprint arXiv:1804.06893}, 2018{\natexlab{c}}.

\bibitem[Zhang et~al.(2017)Zhang, Springenberg, Boedecker, and
  Burgard]{zhang2017deep}
Zhang, J., Springenberg, J.~T., Boedecker, J., and Burgard, W.
\newblock Deep reinforcement learning with successor features for navigation
  across similar environments.
\newblock In \emph{2017 IEEE/RSJ International Conference on Intelligent Robots
  and Systems (IROS)}, pp.\  2371--2378. IEEE, 2017.

\bibitem[Zintgraf et~al.(2018)Zintgraf, Shiarlis, Kurin, Hofmann, and
  Whiteson]{Zintgraf2018FastCA}
Zintgraf, L.~M., Shiarlis, K., Kurin, V., Hofmann, K., and Whiteson, S.
\newblock Fast context adaptation via meta-learning.
\newblock In \emph{ICML}, 2018.

\end{thebibliography}
\bibliographystyle{include/icml2020}

\appendix
\newpage
\appendix  \onecolumn

\section{Network Architectures}
\label{app:arch}

\subsection{Autoencoders}
The policy and dynamics autoencoders are parameterised by Transformers using stacked self-attention and point-wise, fully connected layers for the encoder, and a fully connected feed-forward network for the decoder. 

\textbf{Encoders:} The encoders consist of one layer composed of two sublayers, followed by another fully connected layer. The first sublayer, is a single-head self-attention mechanism, and the second is a simple fully connected feed-forward network. We employ a residual connection around each of the two sub-layers, followed by layer normalization and dropout. We use a dropout of 0.1 for all experiments. To facilitate these residual connections, all sublayers in the model, as well as the embedding layers, produce outputs of dimension $d_{model} = 64$. The second layer of the encoder projects the output of the first layer into the embedding space (from $d_{model}$ to $d_{emb}$). 

The \textbf{policy encoder} takes as input a set of state-action pairs $(s_t, a_t)$ from an full trajectory and outputs an embedding for the policy.  

The \textbf{dynamics encoder} takes as input a set of state-action-next-state tuples $(s_t, a_t, s_{t+1})$ from a full trajectory and outputs an embedding for the dynamics.

The dimension of both the policy and dynamics embedding is $d_{emb} = 8$ for all environments, with the exception of swimmer which uses a dynamics embedding of dimension 2.

\textbf{Decoders:} The decoder is a simple fully connected feed-forward network with three layers and ReLU activations after the first two layers.

The \textbf{policy decoder} takes as input the state of the environment and the policy embedding (outputted by the policy encoder) and outputs an action (i.e. the predicted action taken by the agent). 

The \textbf{environment decoder} takes as input the state of the environment, an action, the dynamics embedding (outputted by the dynamics encoder) and outputs a state (i.e. the predicted next state in the environment). 

The dimensions of the states and actions depend on the given environment. 

\subsection{The Policy-Dynamics Value Function}
\label{app:pdvf}
The Policy-Dynamics Value Function (PD-VF) takes as inputs the initial state of the environment, as well as a policy embedding and a dynamics embedding, and outputs a scalar representing the predicted expected return.

PD-VF is parameterised by a fully connected feed-forward network. First, the environment state and dynamics embedding are concatanated and passed through a linear layer with output dimension 64 followed by a ReLU nonlinearity. The second layer also has output dimension 64 but is followed by a hyperbolic tangent nonlinearity. The output of the second layer is then passed through another linear layer with output dimension equal to the square of the policy embedding dimension which is 64 in this case. Then, the output of this is rearranged in the form of a lower triangular matrix $L$. This matrix is used to construct a Hermitian positive-definite matrix A using the Cholesky decomposition, $A = L L^T$. Finally, the value outputted by the network is obtained by computing $z_{\pi}^T A z_{\pi}$, where $z_{pi}$ is the policy embedding. 

\subsection{Baselines}
\label{app:baselines}
All the pretrained PPO policies as well as all the baselines (except for CondPolicy as explained below) and the ablations use the same actor-critic network architecture. Both the actor and the critic are parameterised two-layer fully connected networks with hidden size 64 and hyperbolic tangent nonlinearities after each layer. Note that the weights are not shared by the two networks. The critic layer has another linear layer on top that outputs the estimated value. The actor network also has a linear layer on top that outputs a vector with the same number of dimensions as the action space. The actions are sampled from a Gaussian distribution with diagonal covariance matrix and means defined by the vector outputted by the actor network. The CondPolicy baseline has a similar architecture. The only difference is the first layer of both the actor and the critic, which has a larger input dimension due to the fact that these networks also take as input the policy embedding (along with the environment state).

\section{Training Details}
\label{app:training}
For experiments on Spaceship and Swimmer, we use only $N_d = 1$ steps to infer the dynamics embedding, while for Ant-wind we use $N_d = 2$ and for Ant-legs we use $N_d = 4$. Note that in all four domains, we only need a few steps to infer the environment dynamics, which allows us to quickly find a good policy for acting during the rest of the episode. Consequently, this results in good performance when evaluated on a single episode.  

First, we have the \textbf{reinforcement learning phase}, in which we pretrain 5 different initializations of PPO policies in each of the 20 environments in our distribution (both those used for training and those used for evaluation). We train all policies for $3e6$ environment interactions, which we have found to be enough for all of them to converge to a stable expected return. 

Then, in the \textbf{self-supervised learning phase}, we use the policies pretrained on the training environments (75 policies for each domain) to generate trajectories through the training environments. For each policy-environment pair, we generate 200 trajectories, half of which are used for training the policy and dynamics autoencodesr and the rest are used for evaluation. We train the autoencoders on this data for a maximum of 200 epochs and we save the models with the lowest evaluation loss. Note that the autoencoders are never trained on trajectories generated in the evaluation environments or by policies pretrained on those environments, but only on data produced by interactions with the training environments. 

Once we have the pretrained policy and dynamics autoencoders, we use them for learning the policy-dynamics value function in the \textbf{supervised training phase}. To do this, we again generate 40 trajectories in the training environments (using only the policies pretrained on those environments). Half of these trajectories are used for training the PD-VF, while the rest are used for evaluation. In our experiments, we have found 20 trajectories from each policy-environment pair to be enough for training the model. For each trajectory, a policy embedding is obtained by passing the full trajectory through the policy encoder. Similarly, a corresponding dynamics embedding is obtained for each trajectory by passing the first few $N_d$ transitions of that trajectory through the dynamics encoder. The initial state and the return of that trajectory are also recorded. Now we have all the data needed for training the PD-VF with supervision. The PD-VF takes as inputs the initial state, the policy and dynamics embeddings and outputs a prediction for the expected return (corresponding to acting with that policy in the given environment). It is trained with $\ell_2$ loss using the observed return. For the initial training stage of the PD-VF, we use 200 epochs, while for the second stage that include data aggregation for the value function and policy decoder, we use 100 epochs. The second stage is repeated a maximum of 20 times (each training for 100 epochs). We select the model that obtains the lowest loss on the evaluation data (out of all the models after each stage). We use this model for probing performace on the evaluation environments. 

\section{Hyperparameters}
\label{app:hyper}
For training the PPO policies, as well as the baselines and ablations, we searched for the learning rate in [0.0001, 0.0003, 0.0005, 0.001] 
and found 0.0003 to work best across the boad. The entropy coefficient was set to 0.0, value loss coefficient 0.5, number of PPO epochs 10, number of PPO steps 2048, number of mini batches 32, gamma 0.99, and generealized advantage estimator coefficient 0.95. We also linearly decay the learning rate.
These values were not searched over since they have been previously optimized for MuJoCo domains and have been shown to be robust across these environments.

For MAML, we used the best hyperparameters found in the original paper for MuJoCo, so meta batch size of 20, 10 batches, and 8 workers.

For the dynamics autoencoder, we did a grid search over the learning rate in [0.0001, 0.001, 0.01] and found 0.001 to be best for the dynamics and 0.01 to be best for the policy. We also searched for the right batch size in [8, 32, 256, 2048] and found 8 to work best for the dynamics and 2048 for the policy. We also did grid searches over $d_{emb} \in [2, 8, 32]$ and found $d_{emb} = 8$ for the policy autoencoders and $d_{emb} = 2$ for the dynamics autoencoders (except for Ant, in which we use $d_{emb} = 8$. We also searched for the hidden dimension of the transformers $d_{model} \in [32, 64, 128]$ and found $d_{model} = 64$ to work best for both the policy and the dynamics embeddings.

For the value function, we tried different values for the number of epochs for the initial training phase $N_{ep, 1} \in $ [1000, 500, 200, 100] and for the second training phase $N_{ep, 2} \in $ [500, 200, 100] and we found 200 and 100 (i.e. each of the 20 data aggregation stages has 100 epochs) to work best, respectively. We also tried different learning rates from [0.0005, 0.001, 0.005, 0.01] and found 0.005 to be the best. Similarly, we tried batch sizes in [64, 128, 256] and found 128 to be the best. 

All the results shown in this paper are obtained using the best hyperparameters found in our grid searches.

\section{Environments}
\label{app:envs}

\subsection{Spaceship Environment}
The source code contains the Spaceship domain that we designed, which is wrapped in a \textit{gym} environment, so it can be easily integrated with any RL algorithm and used to evaluate agents.    

The task consists of moving a spaceship with a unit point charge from one end of a 2D room through a door at the other end. The action space consists of a fixed-magnitude force vector that is applied at each timestep. The room contains two fixed electric charges that deflect/attract the ship as it moves through the environment. 

At the beginning of each episode, the agent\'s location is initialized at the center-bottom of the room with coordinates (2.5, 0.2). The target door is always located at the center-top of the room with coordinates (2.5, 5.0). The size of the room is 5, the size of the door is 1, and the temporal resolution is 0.3 (i.e. the time interval used to compute the next position of the spaceship given the current location and the applied force). The observation consists of the spaceship\'s 2D location in the room (whose coordinates can be any real number between 0 and 5, the size of the room) and the action consists of the 2D force applied by the agent. The episode ends either when the agent hits a wall, exits the room through the door, or the agent has taken more than 50 steps in the environment. At the end of an episode, the agent receives reward that decreases exponentially with its distance to the target door. The decay factor is set to 3.0. For all other steps, the agent receives no reward. 
The distribution of dynamics is a centered circle with radius 1.5.

\subsection{MuJoCo Environments}
For Swimmer we use a circle with radius 0.1 to sample the environment dynamics, while Ant-wind uses a radius of 4.0. For all three domains with continuous distribution of dynamics (i.e. Spaceship, Swimmer, and Ant-wind), we sample 15 environments for training and we hold out 5 for evaluation. The evaluation environments have dynamics covering a closed interval from the distribution thus testing the ability of the model to extrapolate (rather than intrapolate) to different dynamics. The Ant-wind domain has a total of 16 environmets, 4 of which are used for evaluation.

\section{Evaluation}
\label{app:eval}
In this section, we describe in detail the evaluation method and how the results reported here are obtained. For each trained model (i.e. PD-VF, a baseline or an ablation) and for each (unseen) test environment, we use that model to obtain a full trajectory through the given environment. This is repeated 10 times and the average return of the 10 runs is recorded. Then, we compute the mean and standard deviation (of this average return) across 5 different seeds for each model. These are the statistics shown in Figures 4 and 5.

To generate the t-SNE plots, we generated 10 trajectories for each policy-environment pair, including both the training and the evaluation ones. The encoders are used to obtain policy and dynamics embeddings corresponding to each trajectory. Then, t-Distributed Stochastic Neighbor Embedding (t-SNE) with perplexity 30 is applied to produce Figures 6 and 7. Figure 6 shows the t-SNE for the dynamics embedding, where each point is colored by the environment in which the corresponding trajectory (used to obtain that dynamics embedding) was collected. Conversely, Figure 7 shows the t-SNE for the policy embedding, where each point is colored by the policy which generated the corresponding trajectory (used to obtain that policy embedding). 


\begin{figure*}[h!]
        \subfigure{\includegraphics[width=0.50\textwidth]{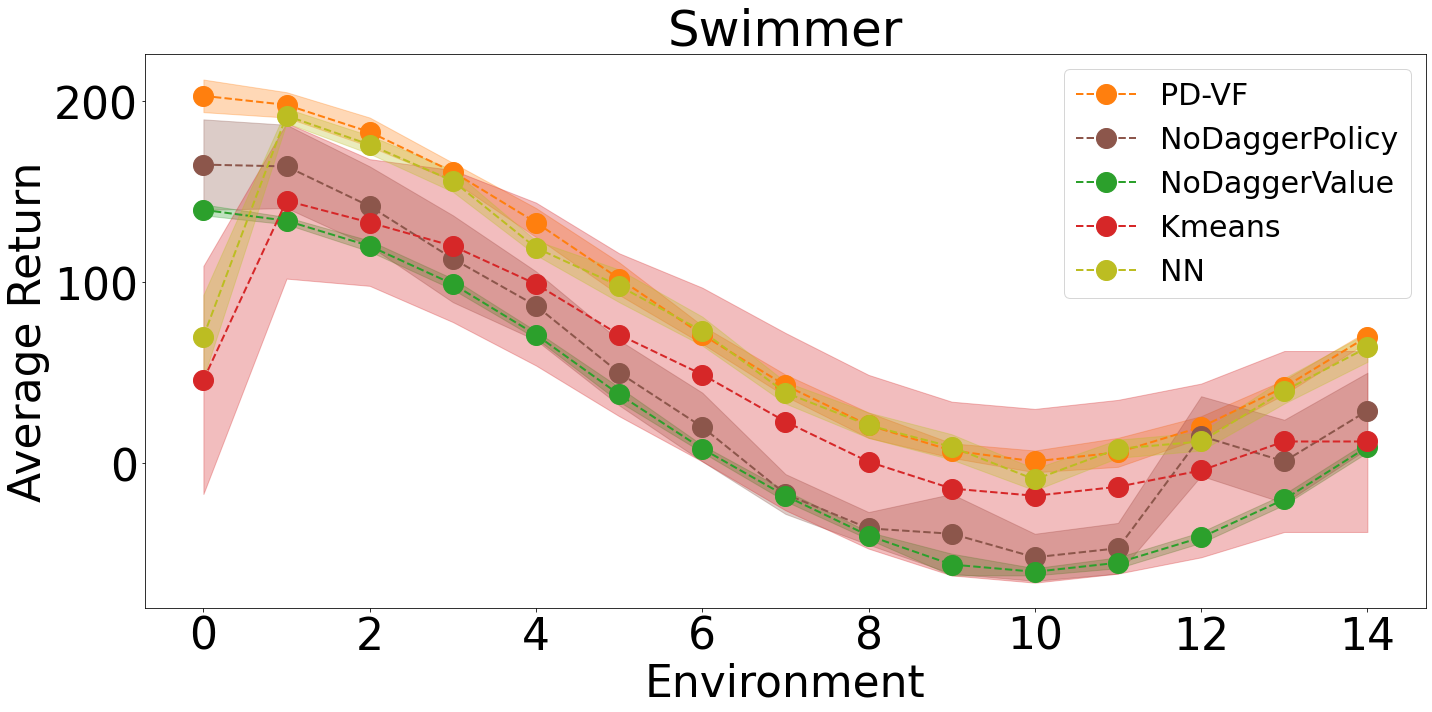}}
        \subfigure{\includegraphics[width=0.50\textwidth]{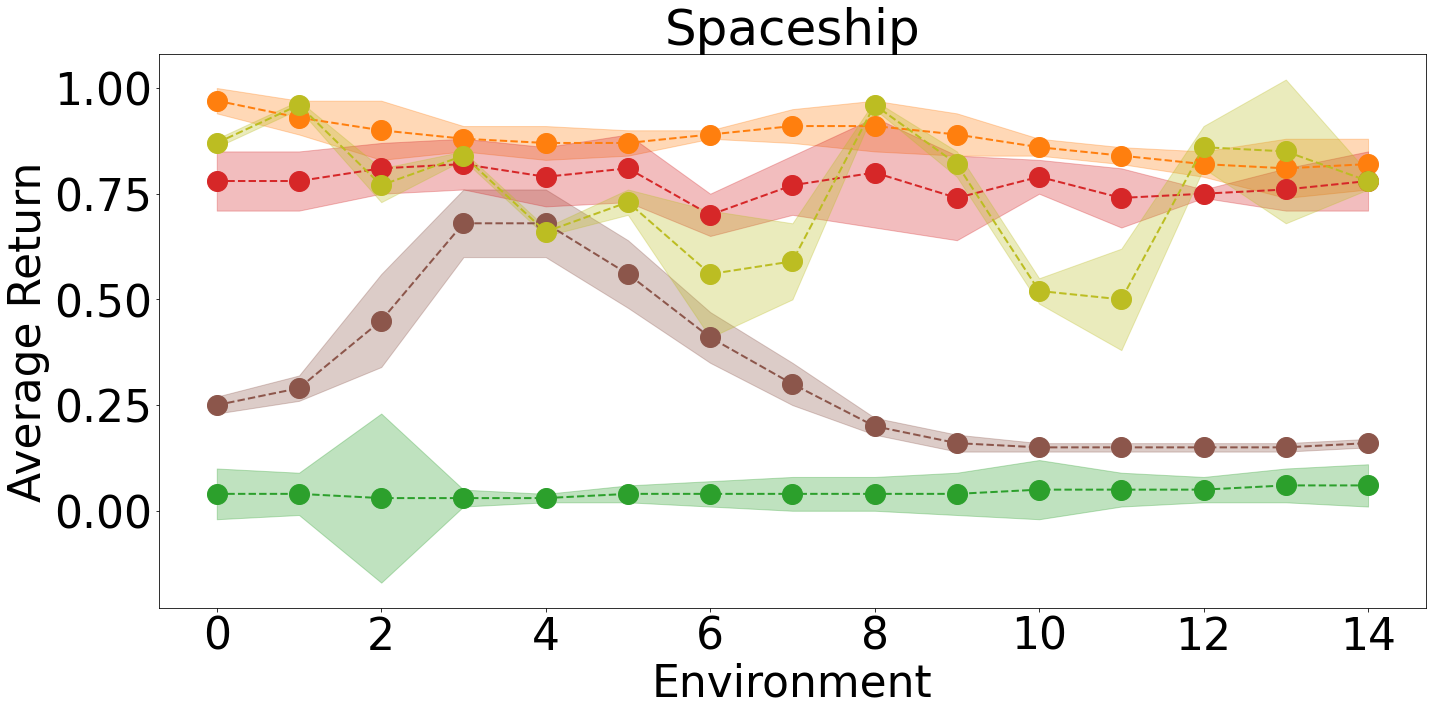}}
       
        \subfigure{\includegraphics[width=0.50\textwidth]{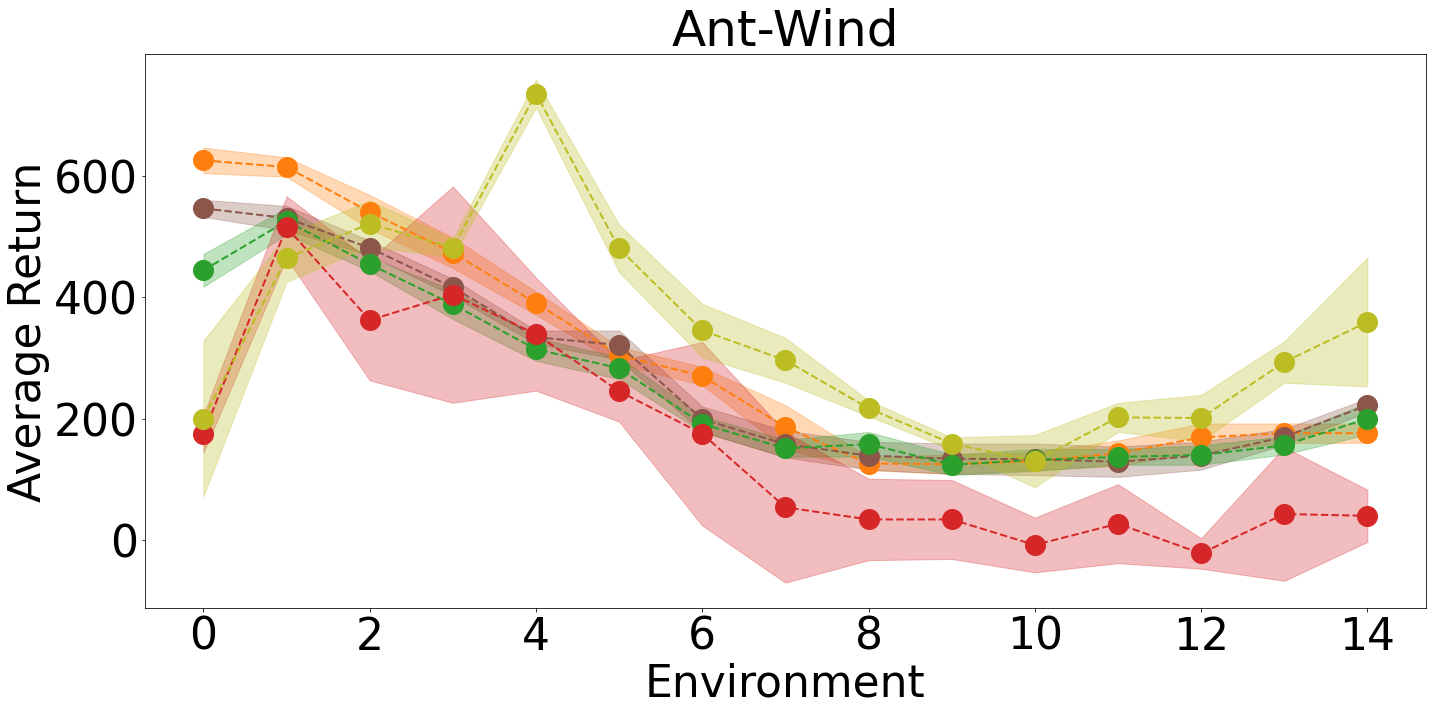}}
        \subfigure{\includegraphics[width=0.50\textwidth]{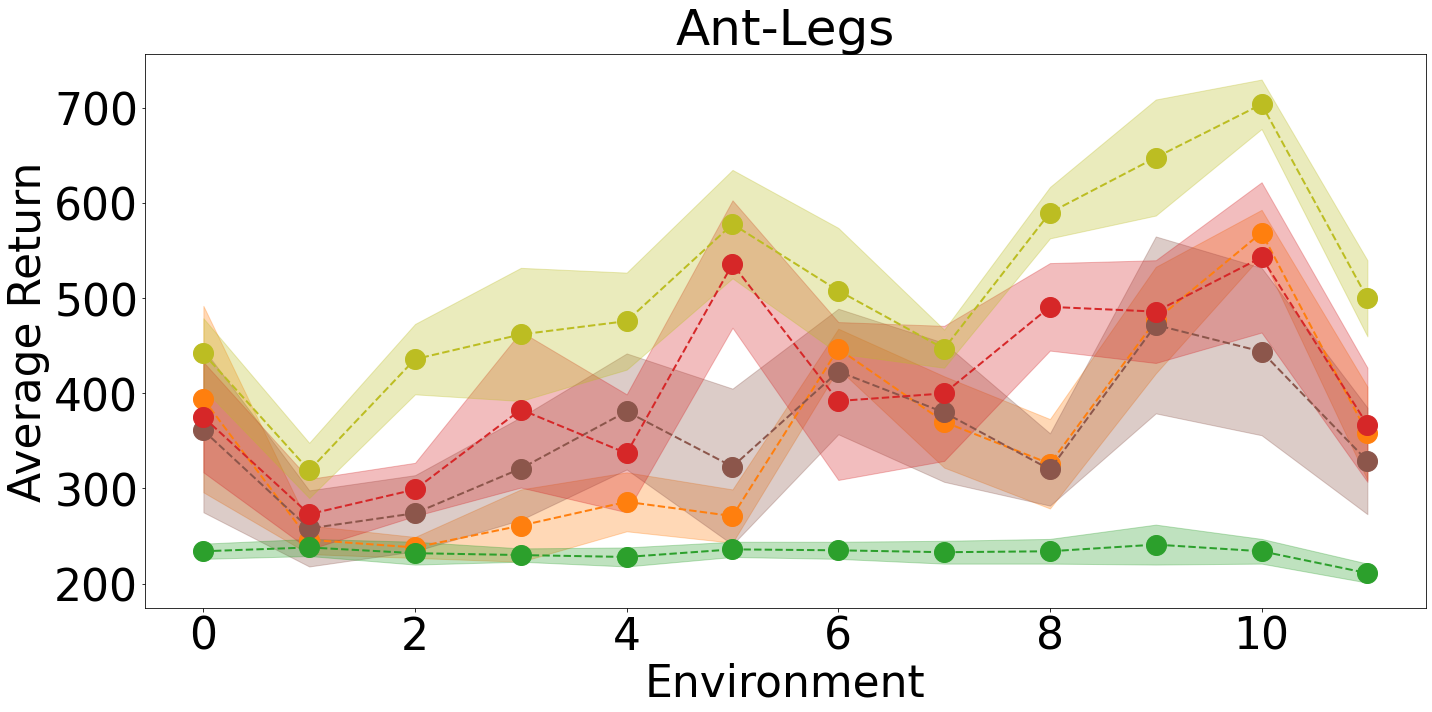}}
    \caption{\textbf{Train Performance.} Average return on train environments in Swimmer (top-left), Spaceship (top-right), Ant-wind (bottom-left), and Ant-legs (bottom-right) obtained by \pdvf{} and a few ablations, namely NoDaggerPolicy, NoDaggerValue, Kmeans, and NN. \pdvf{} is comparable with or outperforms the ablations on the train environments. While some of these ablations perform reasonably well on the environments they are trained on, they generalize poorly to unseen dynamics.}
    \label{fig:train_ablations}
\end{figure*}

\section{Analysis of Learned Embeddings}
\label{app:analysis}

Figure~\ref{fig:policy_embeddings_all} shows a t-SNE plot of the learned policy embeddings for Spaceship, Swimmer, and Ant (from left to right). The top and bottom rows color the embeddings by the policy and environment that generated the corresponding trajectory, respectively. Trajectories produced by the same policy have similar embeddings, while those generated in the same environment are not necessarily close in this embedding space. This shows that the policy embedding preserves information about the policy while disregarding elements of the environment (that generated the corresponding embedded trajectory).

\begin{figure}[h!]
        \subfigure{\includegraphics[width=0.31\columnwidth]{fig/tsne_space_env_env.pdf}}
        \subfigure{\includegraphics[width=0.31\columnwidth]{fig/tsne_swim_env_env.pdf}}
        \subfigure{\includegraphics[width=0.37\columnwidth]{fig/tsne_ant_env_env.pdf}}
        
        \subfigure{\includegraphics[width=0.31\columnwidth]{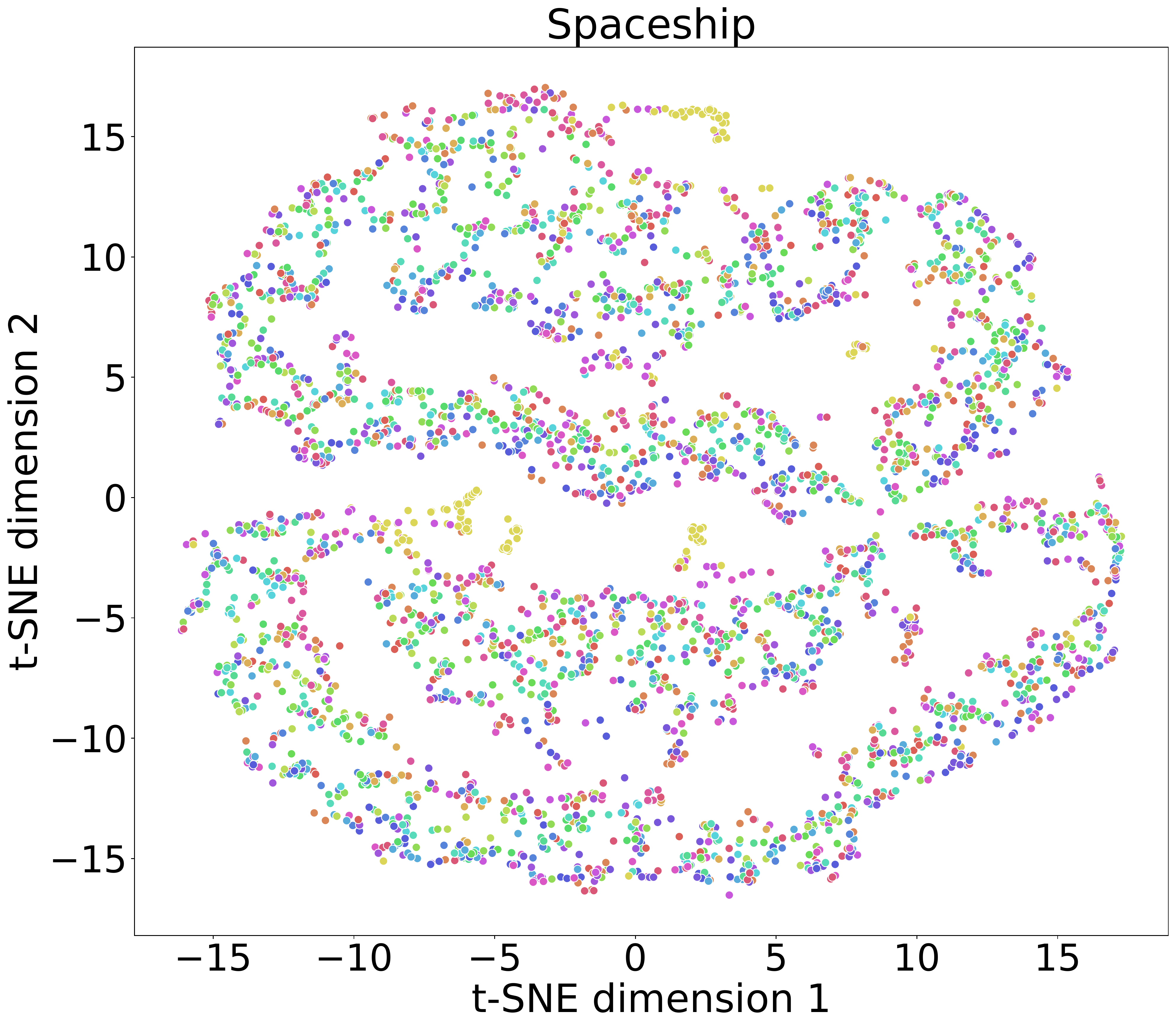}}
        \subfigure{\includegraphics[width=0.31\columnwidth]{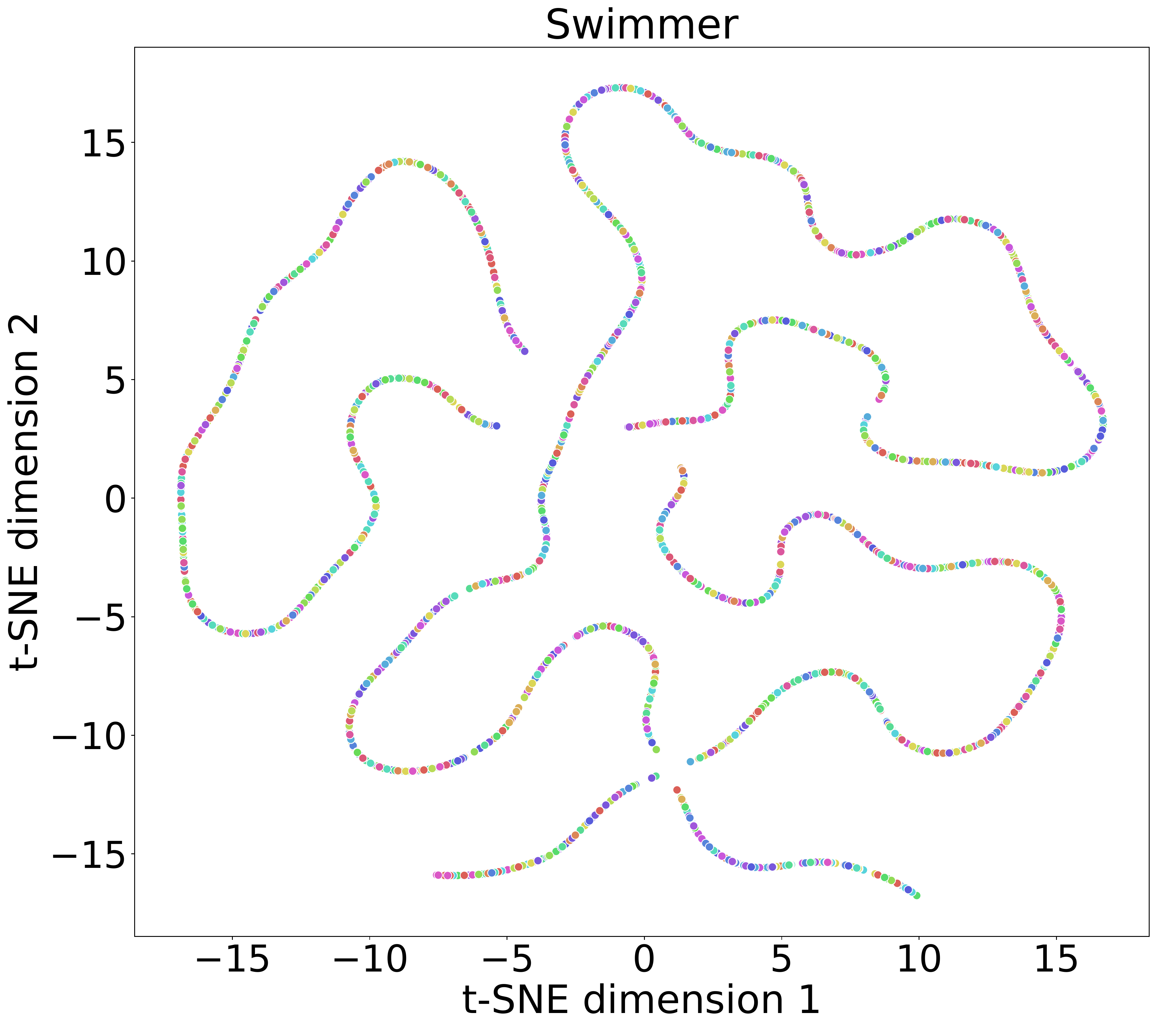}}
        \subfigure{\includegraphics[width=0.34\columnwidth]{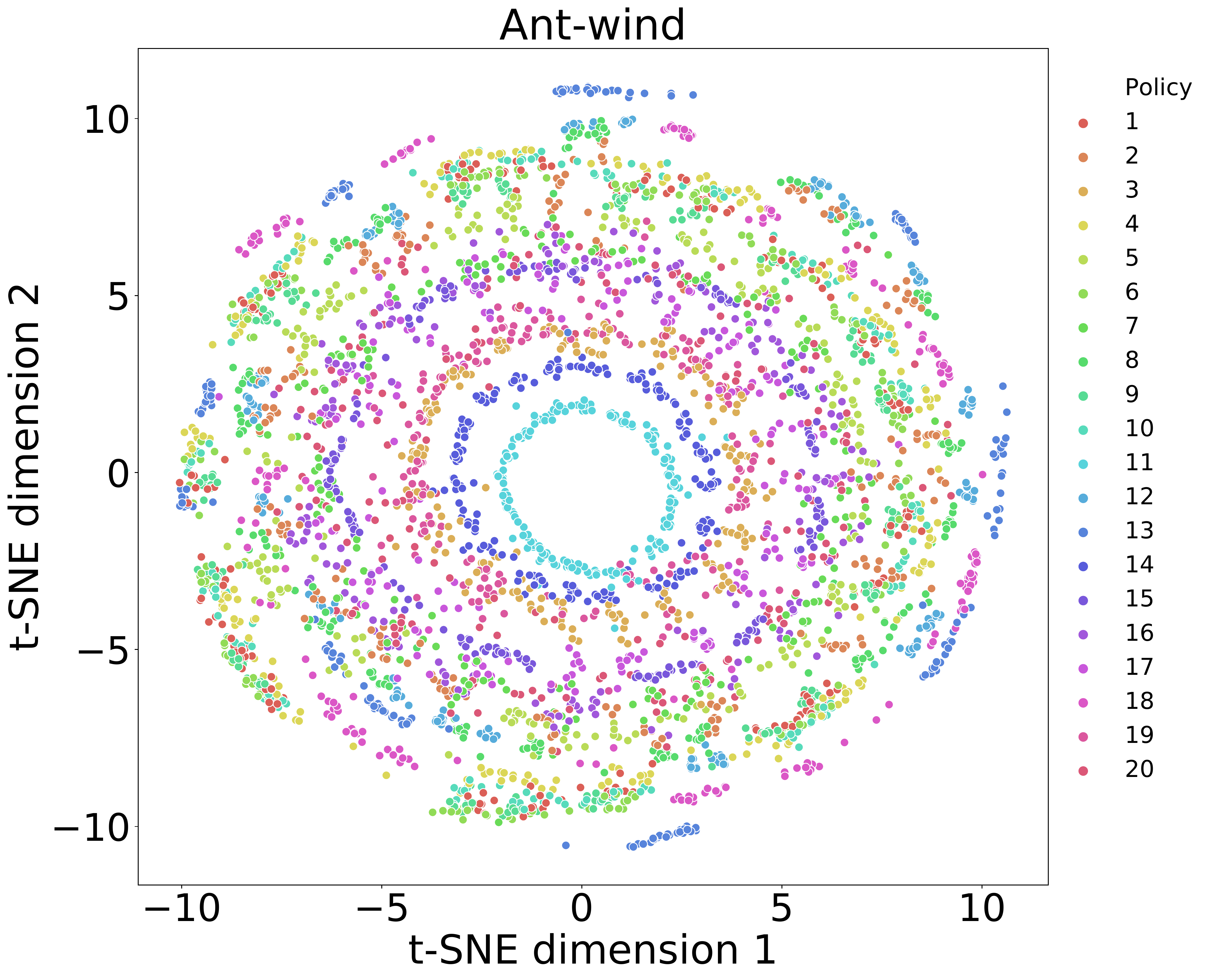}}
    \caption{t-SNE plots of the learned environment embeddings $z_{d}$ for Spaceship, Swimmer, and Ant-wind (from left to right). The points are colored by the \textit{environment} (top) and \textit{policy} (bottom) used to generate the trajectory of the corresponding dynamics embedding.}
    \label{fig:env_embeddings_all}
\end{figure}


Similarly, Figure~\ref{fig:env_embeddings_all} shows a t-SNE plot of the learned dynamics embeddings on the three continuous control domains used for evaluating our method. The top row colors each point by the corresponding environment used to generate the trajectory (from which the embedding is inferred), while the bottom row colors each point by the corresponding policy. One can see that the embedding space retrieves the true dynamics distribution and preserves the smoothness of the 1D manifold.

\begin{figure}[ht!]
        \subfigure{\includegraphics[width=0.31\columnwidth]{fig/tsne_space_pi_pi.pdf}}
        \subfigure{\includegraphics[width=0.31\columnwidth]{fig/tsne_swim_pi_pi.pdf}}
        \subfigure{\includegraphics[width=0.34\columnwidth]{fig/tsne_ant_pi_pi.pdf}}
        
        \subfigure{\includegraphics[width=0.31\columnwidth]{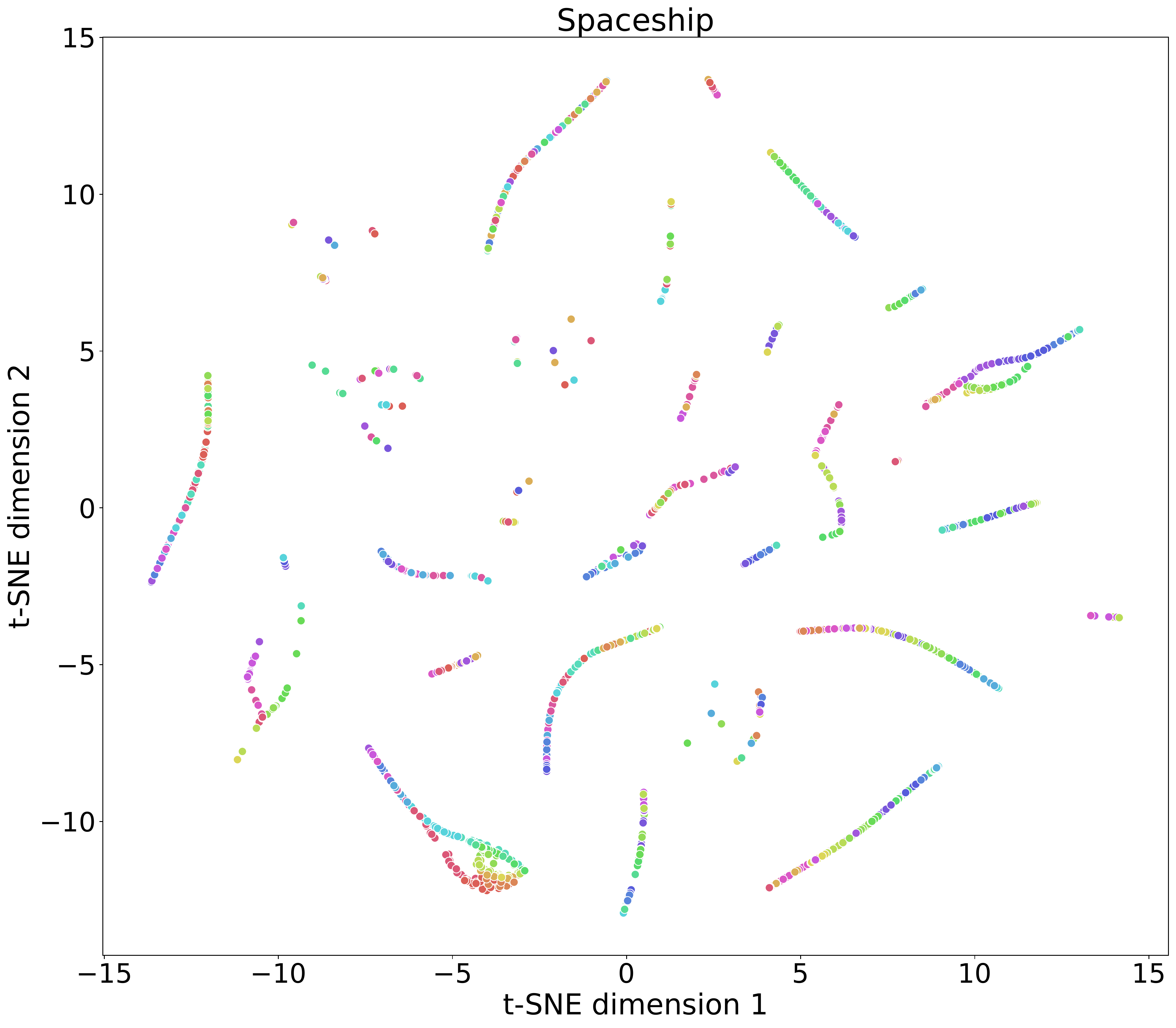}}
        \subfigure{\includegraphics[width=0.31\columnwidth]{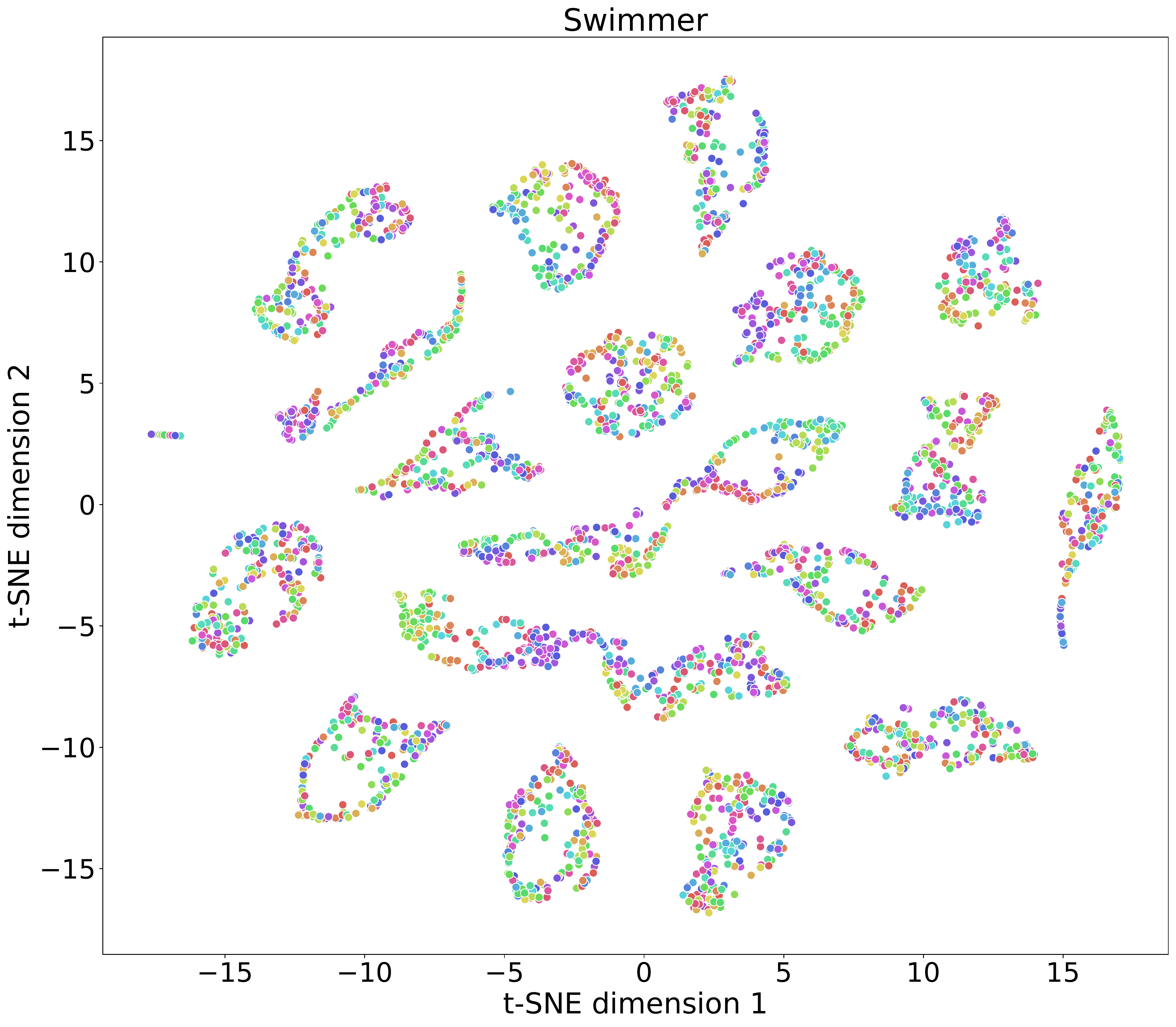}}
        \subfigure{\includegraphics[width=0.37\columnwidth]{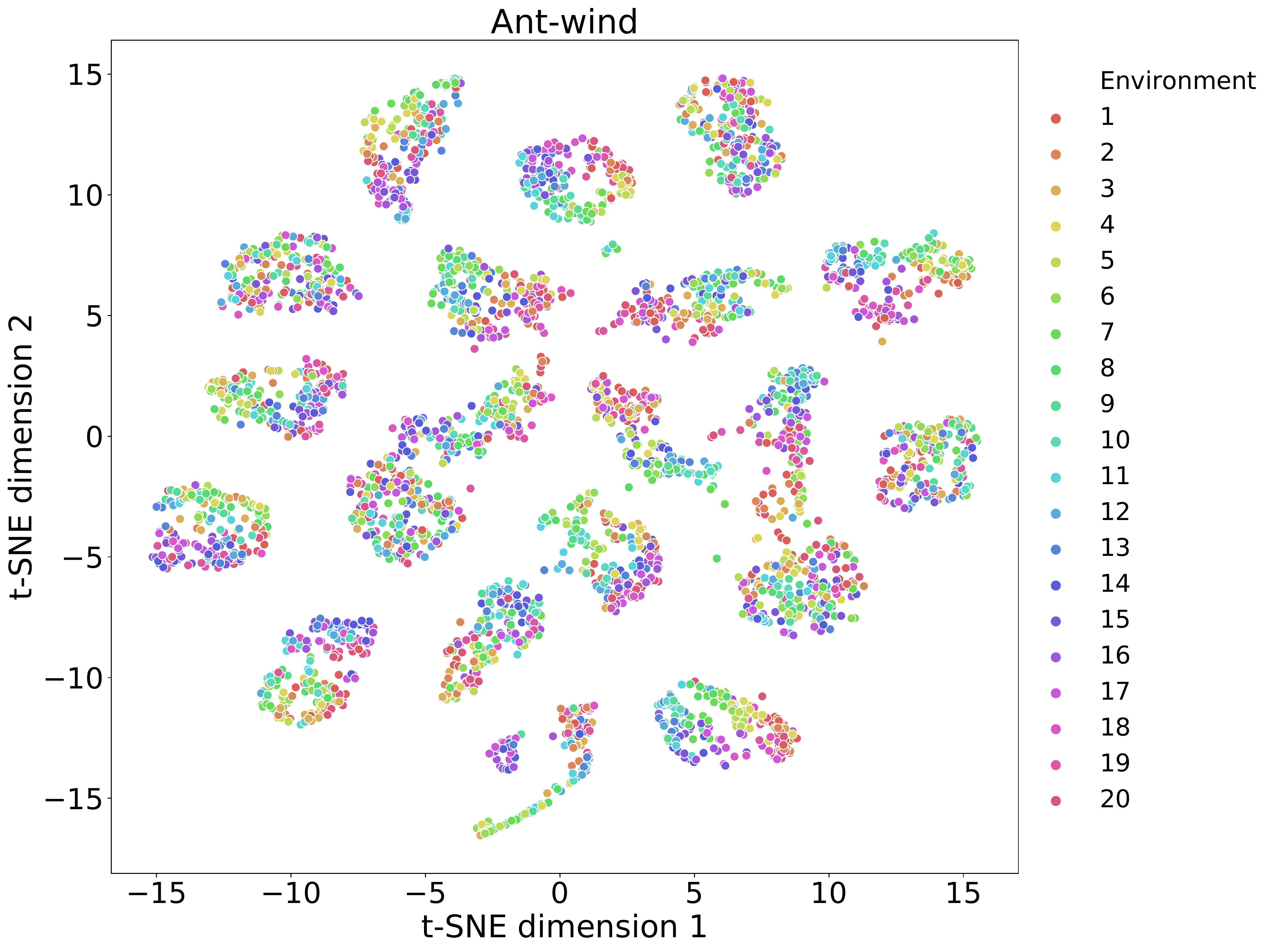}}
    \caption{t-SNE plots of the learned policy embeddings $z_{\pi}$ for Spaceship, Swimmer, and Ant-wind (from left to right). The points are colored by the \textit{policy} (top) and \textit{environment} (bottom) used to generate the trajectory of the corresponding policy embedding.}
    \label{fig:policy_embeddings_all}
\end{figure}


Importantly, this analysis shows that the learned policy and dynamics embeddings are generally disentangled (i.e. information about the dynamics is not contained in the policy space and vice versa). This is important as we want the dynamics space to mostly capture information about the transition function and similarly, we want the policy space to capture variation in the agent behavior. The only exception is the dynamics space of Ant-wind, which contains information about both the environment and the policy. This is because in this environment, the policy is dominated by the force applied to the body of the ant, whose goal is to move forward (while incurring a penalty proportional to the applied force). Thus, depending on the wind direction in the training environment, the agent learns to apply a force of a certain magnitude, a characteristic captured in the embedding space. When evaluated on environments with different dynamics, that policy will still apply a similar force. Our experiments indicate that even if the dynamics space is not fully disentangled (yet it contains information about the environment), the PD-VF is still able to make effective use of the embeddings to find good policies for unseen environments and even outperform other state-of-the-art RL methods. 

\section{The Challenge of Transfer}
\label{app:transfer}
In this section, we emphasize the fact that the family of environments we designed pose a significant challenge to current state-of-the-art RL methods. To do this, we train PPO policies on each of the environments in our set (until convergence) and evaluate them on all other environments. The results show that any of the policies trained in this way can drastically fail in other environments (with different dynamics) from our training and test sets. This demonstrates that our set of environments provides a wide range of dynamics and that a single policy trained in any of these environments does not generalize well to the other ones. Moreover, when evaluated on a single environments, the performance across the pretrained policies varies greatly, illustrating the diversity of collected behaviors (both optimal and suboptimal). This analysis further supports the need for learning about multiple policies (and their performance in various environments) in order to generalize across widely different scenarios (or dynamics in this case).

\begin{table*}[t!]
    \centering
    \small
    \begin{tabular}{c|c|c|c|c|c|c|c|c|c|c}
    \toprule
     & E1 & E2 & E3 & E4 & E5 & E6 & E7 & E8 & E9 & E10 \\ [0.5ex]
    \midrule
    P1  & 598 & 354 & 128 & 372 & 291 & 95.9 & -51.8 & -50.9 & -246 & -299 \\ 
    P2  & 512 & 461 & 503 & 349 & 228 & 135 & -1.44 & -46.8 & -28.1 & -271  \\ 
    P3  & 689 & 620 & 593 & 500 & 334 & 80 & 0.4 & -152 & -51 & -258 \\ 
    P4  & 654 & 665 & 557 & 519 & 29.9 & 180 & 177 & -9.41 & -90.5 & -218.5  \\ 
    P5  & 935 & 962 & 947 & 930 & 853 & 648 & 429 & 287 & 155 & -52.9 \\ 
    P6  & 811 & 838 & 794 & 778 & 710 & 600 & 386 & 247 & 123 & -5.14 \\ 
    P7  & 624 & 659 & 408 & 451 & 351 & 474 & 394 & 82.9 & 151 & 55.9 \\ 
    P8  & 500 & 470 & 442 & 393 & 321 & 468 & 410 & 315 & 209 & 124 \\ 
    P9  & 303 & 326 & 297 & 295 & 265 & 254 & 243 & 238 & 11.1 & 223 \\ 
  P10  & 293 & 54.8 & 294 & 293 & 250 & 226 & 200 & 200 & 180 & -1.28 \\ 
  P11  & 473 & 236 & 212 & 218 & 243 & 144 & 83.9 & 132 & 107 & 136 \\ 
  P12  & 266 & 264 & 268 & 242 & 214 & 181 & 55.6 & 72.7 & 239.4 & 240 \\ 
  P13  & 422 & 669 & 612 & 527 & 401 & 270 & 205 & 128 & 68.5 & 103 \\ 
  P14  & 436 & 362 & 424 & 366 & 259 & 296 & 97.3 & 55.7 & 24.9 & -2.44 \\ 
  P15  & 420 & 484 & 264 & 125 & 131 & 66.7 & 44.5 & 13.1 & 5.43 & 35.1 \\ 
  P16  & 671 & 769 & 573 & 270 & 189 & 212 & 153 & 96.4 & 19.7 & 0.290 \\ 
  P17  & 784 & 793 & 683 & 600 & 56.9 & 200 & 56.8 & 12.4 & 4.4 & 19.8  \\ 
  P18  & 755 & 703 & 564 & 213 & 170 & 129 & 58.1 & 2.17 & -103 & -43.7  \\ 
  P19  & 182 & 593 & 415 & 65.5 & 250 & 112 & 25.8 & 37.1 & -94.9 & -19.2 \\ 
  P20  & 297 & 589 & 518 & 350 & 134 & 76.3 & 6.55 & -51.5 & -185.4 & -11.2 \\ 
    \bottomrule
    \end{tabular}
    \caption{\textbf{Performance of PPO policies on the Ant-wind domain.} A row shows the mean episode return of a single policy on all environments, while a column shows the mean episode return of all policies on a single environment. This table contains performance on the first 10 environments.}
     \label{tab:ppo}
\end{table*}

\begin{table*}[t!]
    \centering
    \small
    \begin{tabular}{c|c|c|c|c|c|c|c|c|c|c}
    \toprule
     & E11 & E12 & E13 & E14 & E15 & E16 & E17 & E18 & E19 & E20 \\ [0.5ex]
    \midrule
     P1  & -336 & -136 & -55.2 & -7.47 & -8.11 & 168 & 262 & 473 & 498 & 603 \\ 
     P2  & -42.2 & -101 & -28.3 & -112 & 14.2 & 132 & 109 & 345 & 510 & 545 \\ 
     P3  & -54.1 & -10.1 & -232 & -6.25 & 59.9 & 120 & 207 & 136 & 563 & 372 \\ 
     P4  & -279 & -278 & -135 & -4.81 & 5.49 & 113 & 276 & 150 & 484 & 196 \\ 
     P5  & -264 & -236 & -69.8 & 14.0 & 77.6 & 248 & 457 & 602 & 813 & 634 \\ 
     P6  & 20.9 & -112 & 98.7 & 147 & 14.9 & 194 & 321 & 524 & 611 & 639 \\ 
     P7  & 0.700 & -5.74 & -61.6 & 9.51 & 89.0 & 267 & 212 & 412 & 579 & 536 \\ 
     P8  & 39.6 & -33.5 & 11.7 & 53.9 & 21.5 & 206 & 218 & 29.9 & 469 & 509 \\ 
     P9  & 7.05 & 216 & 225 & 224 & 248 & 253 & 284 & 266 & 314 & 281 \\ 
    P10  & 176 & -159 & 5.98 & 5.03 & 57.8 & 20.6 & 77.0 & 287 & 249 & 539 \\ 
    P11  & 92.8 & 148 & 206 & 244 & 277 & 351 & 280 & 485 & 556 & 582 \\ 
    P12  & 235 & 235 & 232 & 122 & 139 & 151 & 165 & 299 & 272 & 265 \\ 
    P13  & 84.6 & 171 & 230 & 285 & 312 & 300 & 328 & 523 & 640 & 504 \\ 
    P14  & 45.2 & 159 & 126 & 302 & 369 & 363 & 537 & 457 & 512 & 396 \\ 
    P15  & -4.98 & -70.8 & 41.3 & 360 & 571 & 654 & 687 & 546 & 393 & 537  \\ 
    P16  & -7.77 & 19.7 & 63.4 & 248 & 457 & 600 & 740 & 493 & 731 & 804  \\ 
    P17  & -20.0 & -102 & -8.87 & 69.8 & 254 & 444 & 515 & 658 & 749 & 703 \\ 
    P18  & -93.7 & -59.9 & -11.6 & 125 & 275 & 434 & 577 & 679 & 781 & 781 \\ 
    P19  & -10.1 & -16.9 & 1.38 & 18.6 & 86.5 & 126 & 372 & 508 & 489 & 652 \\ 
    P20  & -118 & -62.4 & -125 & -81.7 & 20.1 & 82.9 & 131 & 195 & 341 & 392 \\ 
    \bottomrule
    \end{tabular}
    \caption{\textbf{Performance of PPO policies on the Ant-wind domain.} A row shows the mean episode return of a single policy on all environments, while a column shows the mean episode return of all policies on a single environment. This table contains performance on the last 10 environments.}
     \label{tab:ppo}
\end{table*}

\begin{table*}[t!]
    \centering
    \small
    \begin{tabular}{c|c|c|c|c|c|c|c|c|c|c}
    \toprule
     & E1 & E2 & E3 & E4 & E5 & E6 & E7 & E8 & E9 & E10 \\ [0.5ex]
    \midrule
    P1 & 139.17 & 133.61 & 119.29 & 96.78 & 69.8 & 38.3 & 7.88 & -19.59 & -41.18 & -54.88 \\
    P2 & 197.01 & 191.23 & 176.68 & 154.75 & 125.97 & 95.2 & 64.06 & 36.33 & 13.79 & -1.1 \\
    P3 & 140.38 & 135.81 & 122.1 & 100.15 & 72.56 & 41.97 & 11.69 & -15.47 & -37.79 & -51.59 \\ 
    P4 & 139.01 & 134.82 & 121.15 & 98.86 & 71.07 & 41.13 & 10.26 & -17.24 & -39.25 & -54.68 \\
    P5 & 215.38 & 209.66 & 196.33 & 174.77 & 147.62 & 115.54 & 84.68 & 59.61 & 36.2 & 22.31 \\
    P6 & 207.08 & 201.55 & 186.96 & 165.49 & 138.86 & 107.64 & 77.12 & 49.94 & 28.6 & 14.55 \\
    P7 & 210.29 & 205.73 & 191.61 & 169.42 & 141.29 & 112.15 & 81.69 & 53.65 & 32.87 & 18.08 \\
    P8 & 213.98 & 209.7 & 198.61 & 179.11 & 152.56 & 124.49 & 95.18 & 67.78 & 45.47 & 32.82 \\
    P9 & 206.51 & 201.71 & 190.56 & 170.55 & 142.16 & 114.04 & 84.64 & 56.6 & 33.8 & 19.84 \\
    P10 & 204.85 & 201.04 & 186.64 & 168.29 & 141.0 & 113.45 & 80.69 & 53.39 & 33.46 & 19.9 \\
    P11 & 197.34 & 193.67 & 182.94 & 164.05 & 137.93 & 109.44 & 80.39 & 52.76 & 32.46 & 18.77 \\
    P12 & 204.92 & 201.11 & 186.85 & 166.49 & 138.03 & 108.22 & 79.32 & 50.95 & 29.97 & 16.23 \\
    P13 & 202.86 & 204.16 & 174.7 & 174.07 & 139.94 & 130.12 & 88.65 & 45.72 & 24.26 & 15.94 \\
    P14 & 205.89 & 201.64 & 187.07 & 166.63 & 138.54 & 108.87 & 77.48 & 51.66 & 29.76 & 15.68 \\
    P15 & 209.19 & 186.31 & 190.71 & 168.87 & 142.23 & 114.88 & 82.87 & 56.37 & 35.82 & 20.97 \\
    P16 & 214.29 & 204.26 & 188.0 & 165.51 & 140.3 & 109.66 & 80.35 & 56.25 & 37.05 & 23.41 \\
    P17 & 202.78 & 197.79 & 183.82 & 160.93 & 133.64 & 103.02 & 71.89 & 44.94 & 22.01 & 9.32 \\
    P18 & 202.66 & 204.15 & 190.16 & 167.73 & 139.81 & 109.03 & 77.82 & 51.18 & 29.4 & 14.83 \\
    P19 & 208.89 & 204.35 & 191.05 & 168.31 & 139.9 & 109.16 & 79.06 & 51.99 & 28.93 & 14.33 \\
    P20 & 141.37 & 136.28 & 122.22 & 100.69 & 73.13 & 42.51 & 12.4 & -15.19 & -37.6 & -51.74 \\
    \bottomrule
    \end{tabular}
    \caption{\textbf{Performance of PPO policies on the Swimmer domain.} A row shows the mean episode return of a single policy on all environments, while a column shows the mean episode return of all policies on a single environment. This table contains performance on the first 10 environments.}
     \label{tab:ppo}
\end{table*}

\begin{table*}[t!]
    \centering
    \small
    \begin{tabular}{c|c|c|c|c|c|c|c|c|c|c}
    \toprule
     & E11 & E12 & E13 & E14 & E15 & E16 & E17 & E18 & E19 & E20 \\ [0.5ex]
    \midrule
     P1 & -59.39 & -54.24 & -40.67 & -19.18 & 9.2 & 39.2 & 69.81 & 97.3 & 119.73 & 134.49 \\
    P2 & -6.11 & -0.47 & 13.55 & 36.02 & 65.06 & 96.16 & 126.34 & 154.61 & 177.55 & 191.34 \\
    P3 & -57.27 & -52.51 & -39.14 & -16.39 & 10.56 & 40.38 & 71.77 & 98.78 & 121.09 & 135.31 \\
    P4 & -58.16 & -53.38 & -39.35 & -18.19 & 9.62 & 40.21 & 70.66 & 98.47 & 120.61 & 134.47 \\
    P5 & 16.13 & 20.32 & 32.89 & 56.59 & 85.07 & 114.87 & 144.05 & 173.91 & 196.76 & 210.27 \\
    P6 & 9.8 & 13.8 & 27.56 & 49.58 & 76.67 & 107.47 & 137.03 & 164.94 & 188.26 & 201.71 \\
    P7  & 14.63 & 19.66 & 34.54 & 56.04 & 82.39 & 114.16 & 143.69 & 171.16 & 190.65 & 205.65 \\
    P8 & 25.48 & 33.05 & 45.05 & 68.32 & 94.65 & 124.24 & 153.37 & 179.63 & 199.71 & 210.72 \\
    P9 & 15.39 & 19.33 & 31.88 & 54.41 & 79.43 & 110.51 & 139.44 & 165.05 & 186.49 & 200.29 \\
    P10 & 12.91 & 16.72 & 32.59 & 52.4 & 79.08 & 107.42 & 138.15 & 161.44 & 184.87 & 199.23 \\
    P11 & 12.84 & 17.28 & 30.41 & 50.77 & 76.78 & 104.17 & 132.33 & 156.83 & 178.78 & 191.66 \\
    P12 & 11.56 & 16.06 & 29.37 & 51.48 & 78.29 & 108.94 & 136.79 & 164.79 & 187.09 & 201.86 \\
    P13 & 3.07 & -8.63 & 30.02 & 59.98 & 72.13 & 110.56 & 128.87 & 148.73 & 173.19 & 192.25 \\
    P14 & 10.65 & 16.23 & 30.17 & 52.24 & 79.07 & 109.76 & 139.88 & 166.59 & 187.43 & 202.78 \\
    P15 & 14.82 & 21.48 & 35.28 & 57.38 & 82.5 & 114.19 & 144.17 & 169.13 & 190.07 & 203.95 \\
    P16 & 18.92 & 23.92 & 36.31 & 57.05 & 88.57 & 116.78 & 147.6 & 172.97 & 196.18 & 209.08 \\
    P17 & 4.5 & 9.35 & 23.38 & 45.9 & 72.68 & 103.65 & 133.52 & 161.88 & 183.18 & 197.69 \\
    P18 & 10.91 & 14.36 & 29.04 & 49.94 & 78.39 & 109.27 & 140.76 & 167.58 & 189.9 & 202.89 \\
    P19 & 9.76 & 14.41 & 27.79 & 49.65 & 76.61 & 109.48 & 140.12 & 167.97 & 188.71 & 203.91 \\
    P20 & -56.61 & -52.49 & -38.36 & -16.25 & 11.95 & 42.14 & 72.88 & 100.43 & 122.74 & 136.87 \\
    \bottomrule
    \end{tabular}
    \caption{\textbf{Performance of PPO policies on the Swimmer domain.} A row shows the mean episode return of a single policy on all environments, while a column shows the mean episode return of all policies on a single environment. This table contains performance on the last 10 environments.}
     \label{tab:ppo}
\end{table*}

\begin{table*}[t!]
    \centering
    \small
    \begin{tabular}{c|c|c|c|c|c|c|c|c|c|c}
    \toprule
     & E1 & E2 & E3 & E4 & E5 & E6 & E7 & E8 & E9 & E10 \\ [0.5ex]
    \midrule
    P1  & 0.00 & 0.00 & 0.00 & 0.00 & 0.00 & 0.00 & 0.00 & 0.00 & 0.00 & 0.00 \\
    P2  & 0.97 & 0.97 & 0.92 & 0.92 & 0.90 & 0.91 & 0.93 & 0.95 & 0.99 & 0.95 \\
    P3  & 0.97 & 0.92 & 0.85 & 0.85 & 0.86 & 0.86 & 0.86 & 0.91 & 0.94 & 0.94 \\
    P4  & 0.00 & 0.00 & 0.00 & 0.00 & 0.00 & 0.00 & 0.00 & 0.00 & 0.00 & 0.00 \\
    P5  & 0.61 & 0.61 & 0.60 & 0.58 & 0.57 & 0.55 & 0.53 & 0.51 & 0.50 & 0.49 \\
    P6  & 0.96 & 0.90 & 0.87 & 0.87 & 0.88 & 0.90 & 0.91 & 0.92 & 0.96 & 0.95 \\
    P7  & 0.83 & 0.89 & 0.93 & 0.96 & 0.97 & 0.96 & 0.94 & 0.89 & 0.84 & 0.80 \\
    P8  & 0.84 & 0.84 & 0.84 & 0.82 & 0.81 & 0.79 & 0.78 & 0.77 & 0.75 & 0.74 \\
    P9  & 0.97 & 0.93 & 0.90 & 0.87 & 0.86 & 0.86 & 0.87 & 0.87 & 0.87 & 0.86 \\
    P10 & 0.91 & 0.90 & 0.88 & 0.87 & 0.87 & 0.88 & 0.89 & 0.92 & 0.96 & 0.98 \\
    P11 & 0.77 & 0.78 & 0.79 & 0.80 & 0.82 & 0.78 & 0.72 & 0.76 & 0.79 & 0.80 \\
    P12 & 0.67 & 0.56 & 0.39 & 0.46 & 0.58 & 0.41 & 0.39 & 0.46 & 0.49 & 0.47 \\
    P13 & 0.38 & 0.36 & 0.32 & 0.26 & 0.30 & 0.25 & 0.32 & 0.32 & 0.33 & 0.40 \\
    P14 & 0.82 & 0.79 & 0.76 & 0.74 & 0.73 & 0.72 & 0.72 & 0.73 & 0.75 & 0.76 \\
    P15 & 0.67 & 0.66 & 0.65 & 0.63 & 0.62 & 0.62 & 0.61 & 0.61 & 0.61 & 0.61 \\
    P16 & 0.00 & 0.00 & 0.00 & 0.00 & 0.00 & 0.00 & 0.00 & 0.00 & 0.00 & 0.00 \\
    P17 & 0.86 & 0.82 & 0.80 & 0.78 & 0.77 & 0.78 & 0.79 & 0.82 & 0.85 & 0.87 \\
    P18 & 0.69 & 0.68 & 0.66 & 0.65 & 0.64 & 0.63 & 0.63 & 0.63 & 0.63 & 0.63 \\
    P19 & 0.80 & 0.81 & 0.81 & 0.73 & 0.75 & 0.78 & 0.81 & 0.79 & 0.78 & 0.90 \\
    P20 & 0.96 & 0.94 & 0.90 & 0.88 & 0.87 & 0.86 & 0.86 & 0.86 & 0.85 & 0.83 \\
    \bottomrule
    \end{tabular}
    \caption{\textbf{Performance of PPO policies on the Spaceship domain.} A row shows the mean episode return of a single policy on all environments, while a column shows the mean episode return of all policies on a single environment. This table contains performance on the first 10 environments.}
     \label{tab:ppo}
\end{table*}

\begin{table*}[t!]
    \centering
    \small
    \begin{tabular}{c|c|c|c|c|c|c|c|c|c|c}
    \toprule
     & E11 & E12 & E13 & E14 & E15 & E16 & E17 & E18 & E19 & E20 \\ [0.5ex]
    \midrule
    P1  & 0.00 & 0.00 & 0.00 & 0.00 & 0.00 & 0.00 & 0.00 & 0.00 & 0.00 & 0.00 \\
    P2  & 0.89 & 0.85 & 0.82 & 0.79 & 0.79 & 0.79 & 0.81 & 0.84 & 0.87 & 0.92 \\
    P3  & 0.93 & 0.91 & 0.86 & 0.83 & 0.82 & 0.82 & 0.84 & 0.86 & 0.91 & 0.95 \\
    P4  & 0.00 & 0.00 & 0.00 & 0.00 & 0.00 & 0.00 & 0.00 & 0.00 & 0.00 & 0.00 \\
    P5  & 0.49 & 0.49 & 0.49 & 0.50 & 0.51 & 0.53 & 0.55 & 0.57 & 0.59 & 0.61 \\
    P6  & 0.91 & 0.85 & 0.82 & 0.79 & 0.78 & 0.78 & 0.80 & 0.82 & 0.86 & 0.92 \\
    P7  & 0.74 & 0.70 & 0.67 & 0.65 & 0.64 & 0.65 & 0.66 & 0.70 & 0.75 & 0.79 \\
    P8  & 0.73 & 0.71 & 0.70 & 0.70 & 0.70 & 0.71 & 0.73 & 0.76 & 0.79 & 0.82 \\
    P9  & 0.84 & 0.82 & 0.80 & 0.79 & 0.80 & 0.81 & 0.84 & 0.87 & 0.93 & 0.97 \\
    P10 & 0.94 & 0.90 & 0.86 & 0.84 & 0.83 & 0.83 & 0.85 & 0.88 & 0.90 & 0.92 \\
    P11 & 0.81 & 0.93 & 0.90 & 0.87 & 0.84 & 0.82 & 0.80 & 0.78 & 0.77 & 0.77 \\
    P12 & 0.46 & 0.83 & 0.61 & 0.44 & 0.77 & 0.74 & 0.64 & 0.59 & 0.59 & 0.63 \\
    P13 & 0.43 & 0.73 & 0.64 & 0.81 & 0.67 & 0.69 & 0.80 & 0.63 & 0.51 & 0.41 \\
    P14 & 0.79 & 0.80 & 0.82 & 0.85 & 0.86 & 0.89 & 0.91 & 0.91 & 0.89 & 0.85 \\
    P15 & 0.61 & 0.62 & 0.62 & 0.63 & 0.64 & 0.65 & 0.66 & 0.67 & 0.68 & 0.68 \\
    P16 & 0.00 & 0.00 & 0.00 & 0.00 & 0.00 & 0.00 & 0.00 & 0.00 & 0.00 & 0.00 \\
    P17 & 0.90 & 0.91 & 0.92 & 0.93 & 0.94 & 0.96 & 0.98 & 0.98 & 0.94 & 0.90 \\
    P18 & 0.63 & 0.63 & 0.64 & 0.64 & 0.65 & 0.67 & 0.68 & 0.69 & 0.70 & 0.70 \\
    P19 & 0.91 & 0.89 & 0.87 & 0.93 & 0.89 & 0.86 & 0.83 & 0.81 & 0.80 & 0.80 \\
    P20 & 0.80 & 0.78 & 0.77 & 0.76 & 0.76 & 0.77 & 0.80 & 0.85 & 0.90 & 0.93 \\

    \bottomrule
    \end{tabular}
    \caption{\textbf{Performance of PPO policies on the Spaceship domain.} A row shows the mean episode return of a single policy on all environments, while a column shows the mean episode return of all policies on a single environment. This table contains performance on the last 10 environments.}
     \label{tab:ppo}
\end{table*}





\end{document}
